\documentclass[10pt,journal,compsoc]{IEEEtran}

\ifCLASSOPTIONcompsoc
  \usepackage[nocompress]{cite}
\else
  \usepackage{cite}
\fi

\ifCLASSINFOpdf
\else
\fi

\usepackage{rotating}
\usepackage{graphicx}
\usepackage{amsmath,amssymb} % define this before the line numbering.
\usepackage{url}
\usepackage{tikz}
\usepackage{pgfplots}
\usepackage[pagebackref=false,breaklinks=true,letterpaper=true,hidelinks,bookmarks=false]{hyperref}
\usepackage{color}
\usepackage{pifont}

\pgfplotsset{compat=newest}
\newcommand{\showodsf}[3]{%
\mbox{\input{data/pr/#1_#2_#3_ods_f.txt}\hspace{-2.5pt}}%
}

\definecolor{olivegreen}{RGB}{0,170,0}
\definecolor{darkred}{RGB}{220,100,10}
\definecolor{tealblue}{RGB}{20,100,200}

\newcommand{\deflinewidth}{1.3pt} % Line width

\usepackage{subfiles}

\usepackage{booktabs}
\usepackage{multicol}
\definecolor{rowblue}{RGB}{220,230,240}

\usepackage{capt-of}

\newcommand{\cfbox}[2]{%
    \colorlet{currentcolor}{.}%
    {\color{#1}%
    \fbox{\color{currentcolor}#2}}%
}

\newcommand{\showinonecol}[6]{%
\fbox{\includegraphics[height=#6\linewidth]{img/#5/#1_#4.png}}
\hfill
\fbox{\includegraphics[height=#6\linewidth]{img/#5/#2_#4.png}}
\hfill
\fbox{\includegraphics[height=#6\linewidth]{img/#5/#3_#4.png}}
\vspace{2mm}
}

\newcommand{\showonerow}[8]{%
\fbox{\includegraphics[height=#8\linewidth]{img/#7/#1_#2.png}}
\hfill
\fbox{\includegraphics[height=#8\linewidth]{img/#7/#1_#3.png}}
\hfill
\fbox{\includegraphics[height=#8\linewidth]{img/#7/#1_#4.png}}
\hfill
\fbox{\includegraphics[height=#8\linewidth]{img/#7/#1_#5.png}}
\hfill
\fbox{\includegraphics[height=#8\linewidth]{img/#7/#1_#6.png}}
\vspace{2mm}
}

\begin{document}

\title{Convolutional Oriented Boundaries:\\From Image Segmentation to High-Level Tasks}

\author{Kevis-Kokitsi Maninis,
        Jordi Pont-Tuset,
        Pablo Arbel\'aez,
        and~Luc Van Gool% <-this % stops a space
\IEEEcompsocitemizethanks{\IEEEcompsocthanksitem K.-K. Maninis, J. Pont-Tuset, and L. Van Gool are with the Computer Vision Laboratory, ETHZ, Switzerland. P. Arbel\'aez is with the Department of Biomedical Engineering, Universidad de los Andes, Colombia. Contacts: see \url{www.vision.ee.ethz.ch/\textasciitilde cvlsegmentation/}}}% <-this % stops an unwanted space

%\markboth{IEEE Transactions on Pattern Analysis and Machine Intelligence}%
%{}

\IEEEtitleabstractindextext{%
\begin{abstract}
We present Convolutional Oriented Boundaries (COB), which produces multiscale oriented contours and region hierarchies starting from generic image classification Convolutional Neural Networks (CNNs). COB is computationally efficient, because it requires a single CNN forward pass for multi-scale contour detection and it uses a novel sparse boundary representation for hierarchical segmentation; it gives a significant leap in performance over the state-of-the-art, and it generalizes very well to unseen categories and datasets. Particularly, we show that learning to estimate not only contour strength but also orientation provides more accurate results. 
We perform extensive experiments for low-level applications on BSDS, PASCAL Context, PASCAL Segmentation, and NYUD to evaluate boundary detection performance, showing that COB provides state-of-the-art contours and region hierarchies in all datasets. We also evaluate COB on high-level tasks when coupled with multiple pipelines for object proposals, semantic contours, semantic segmentation, and object detection on MS-COCO, SBD, and PASCAL; showing that COB also improves the results for all tasks.
\end{abstract}

}

\maketitle

\IEEEdisplaynontitleabstractindextext
\IEEEpeerreviewmaketitle

\IEEEraisesectionheading{\section{Introduction}\label{sec:intro}}

\begin{figure*}[t]
\centering
\includegraphics[width=0.75\textwidth]{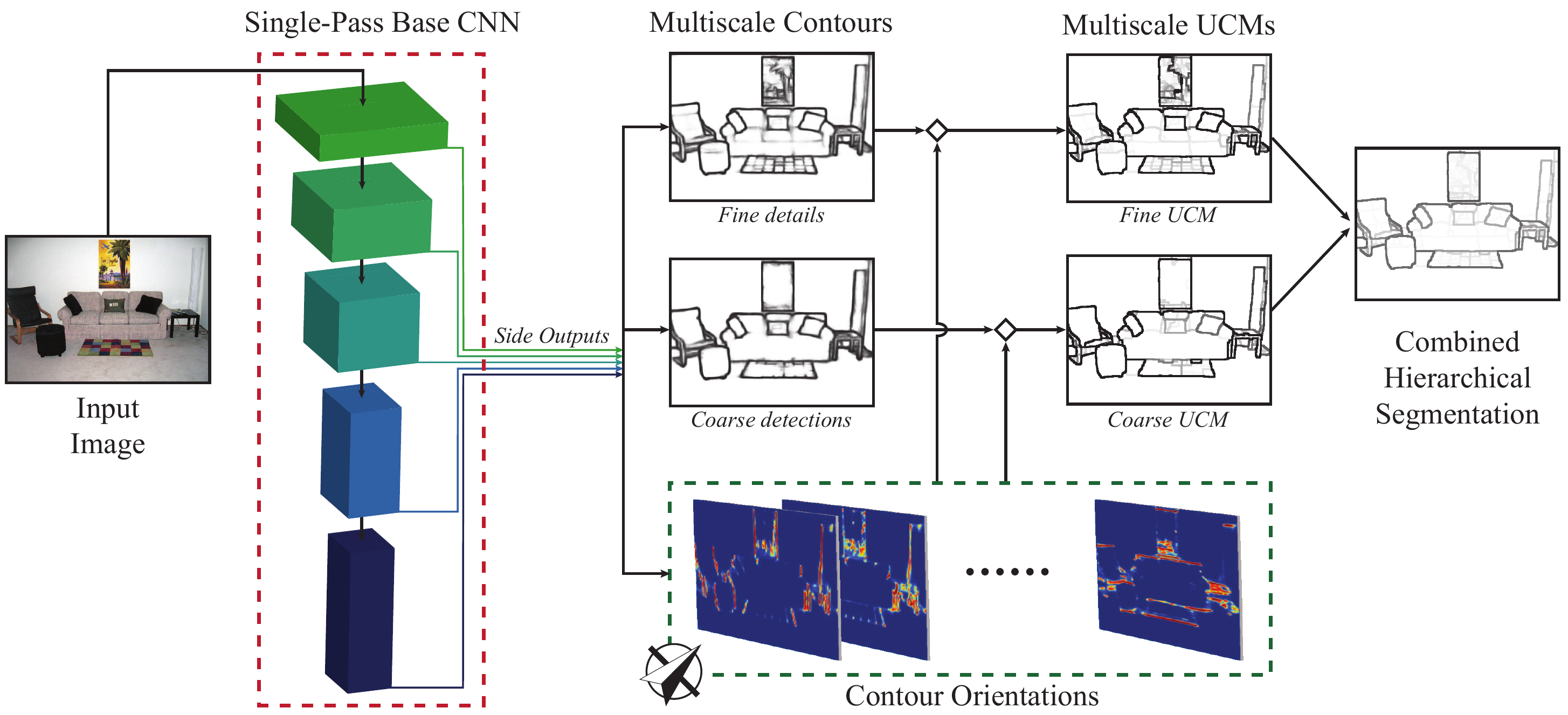}\\[-2mm]
	\caption{\textbf{Overview of COB}: From a single pass of a base CNN, we obtain multiscale oriented contours. We combine them to build Ultrametric Contour Maps (UCMs) at different scales and fuse them into a single hierarchical segmentation structure.}
\label{fig:overview}
\vspace{-4mm}
\end{figure*}

\IEEEPARstart{T}{he} adoption of Convolutional Neural Networks (CNNs) has caused a profound change and a large leap forward in performance throughout the majority of fields in computer vision.
In the case of a traditionally category-agnostic field such as contour detection, it has recently fostered the appearance of systems~\cite{Kokkinos2016,XiTu15,BST15a,BST15b,She+15,GaLe14} that rely on large-scale category-specific information in the form of deep architectures pre-trained on Imagenet~\cite{Russakovsky2015} for image classification~\cite{Krizhevsky2012,Sze+15,SiZi15,He+16}.

This paper proposes Convolutional Oriented Boundaries (COB), a generic CNN architecture that allows end-to-end learning of multiscale oriented contours, and we show how it translates top performing base CNN networks into high-quality contours; allowing to bring future improvements in base CNN architectures into semantic grouping.
We then propose a sparse boundary representation for efficient construction of hierarchical regions from the contour signal. Our overall approach is both efficient (it runs in 0.8 seconds per image) and highly accurate (it produces state-of-the-art contours and regions on PASCAL and on the BSDS). Figure~\ref{fig:overview} shows an overview of our system.

For the last fifteen years, the Berkeley Segmentation Dataset and Benchmark (BSDS)~\cite{Martin2001} has been the experimental testbed of choice for the study of boundary detection and image segmentation.
However, the current large-capacity and very accurate models have underlined the limitations of the BSDS as the primary benchmark for grouping. Its 300 train images are inadequate for training systems with tens of millions of parameters and, critically, current state-of-the-art techniques are reaching human performance for boundary detection on its 200 test images.

In terms of scale and difficulty, the next natural frontier for perceptual grouping is the PASCAL VOC dataset~\cite{Eve+12}, an influential benchmark for image classification, object detection, and semantic segmentation which has a \textit{trainval} set with more than 10\,000 challenging and varied images. A first step in that direction was taken by Hariharan et al.~\cite{Har+11}, who annotated the VOC dataset for category-specific boundary detection on the foreground objects. More recently, the PASCAL Context dataset~\cite{Mot+14} extended this annotation effort to all the background categories, providing thus fully-parsed images which are a direct VOC counterpart to the human ground truth of the BSDS. 
In this direction, this paper investigates the transition from the BSDS to PASCAL Context in the evaluation of image segmentation.

We derive valuable insights from studying perceptual grouping in a larger and more challenging empirical framework. Among them, we observe that COB leverages increasingly deeper state-of-the-art architectures, such as the recent Residual Networks~\cite{He+16}, to produce improved results. This indicates that our approach is generic and can directly benefit from future advances in CNNs. We also observe that, in PASCAL, the globalization strategy of contour strength by spectral graph partitioning proposed in~\cite{Arb+11} and used in state-of-the-art methods~\cite{Pont-Tuset2016,Kokkinos2016} is unnecessary in the presence of the high-level knowledge conveyed by pre-trained CNNs and oriented contours, thus removing a significant computational bottleneck for high-quality contours.

We conduct two types of experiments, the first of which regards low-level vision applications, such as contour detection and generic segmentation on PASCAL Context and the BSDS500. We extend the evaluation to the NYUD RGB-D dataset, showing that the pipeline of COB can benefit from depth embeddings. We also include evaluation of object contour detection on the PASCAL VOC'12 database. In all cases, COB demonstrates state-of-the-art performance on contours and regions while being computationally efficient.

In a second set of experiments, we study the interplay of COB with various downstream recognition applications. We use our hierarchical regions as input to the combinatorial grouping algorithm of~\cite{Pont-Tuset2016} and obtain state-of-the-art segmented object proposals on PASCAL VOC'12 Segmentation by a significant margin. Furthermore, we provide empirical evidence for the generalization power of COB by evaluating our object proposals without any retraining in the even larger and more challenging MS-COCO~\cite{Lin2014a} dataset, where we also report competitive results compared to the state of the art.
We have also studied the effects of COB when coupled with well-known pipelines, showing that injecting COB detections to them lead to improvements on Semantic Segmentation and Object Detection. Finally, we report a new state of the art on Semantic Boundary detection.

Our approach to segmentation has also found application in retinal image segmentation~\cite{Man+16}, obtaining state-of-the-art and super-human performance in vessel and optic disc segmentation, which further highlights its generality.

The COB code, pre-computed results, pre-trained models, and benchmarks are publicly available at
\url{www.vision.ee.ethz.ch/\textasciitilde cvlsegmentation/}.

\section{Related Work}
\label{sec:related}
{\setlength{\parindent}{0cm}\paragraph*{\bf Contour Detection}
Early approaches to contour detection relied on local gradient measurements in an image~\cite{Rob63,Kit83,Pre70}. These simple edge detectors operate by applying local derivative filters on grayscale images. Gradient filtering was followed by detection of zero crossings~\cite{MaHi80}, or by non-maximum suppression~\cite{Cann86}.}

Such simple gradient techniques are unable to handle information captured by richer features such as color and texture~\cite{Martin2004}, or Statistical Edges~\cite{Kon+03}. Martin et al.~\cite{Martin2004} define rich gradient operators out of color, brightness and texture, and use them as input to a logistic regression classifier. Their approach is extended by Arbel\'aez et al.~\cite{Arb+11}, to combine contours at multiple scales.

Machine Learning techniques contributed to learnable features and classifiers that boosted contour detection performance, especially after the manual annotation of the BSDS database~\cite{Martin2004, Arb+11}.
The BEL algorithm~\cite{DTB06} attempts to learn an edge classifier in the form of a probabilistic boosting tree.
Kokkinos~\cite{Kok10a} trains an orientation-sensitive boundary detector using Multiple-Instance Learning.
Ren and Bo~\cite{ReBo12} use patch representations automatically learned through sparse coding.
Sketch Tokens~\cite{LZD13} and  Structured Edges~\cite{DoZi15} tackle both accuracy and speed, by using random forests to classify patches.

\begin{figure*}
\centering
\includegraphics[width=0.85\linewidth]{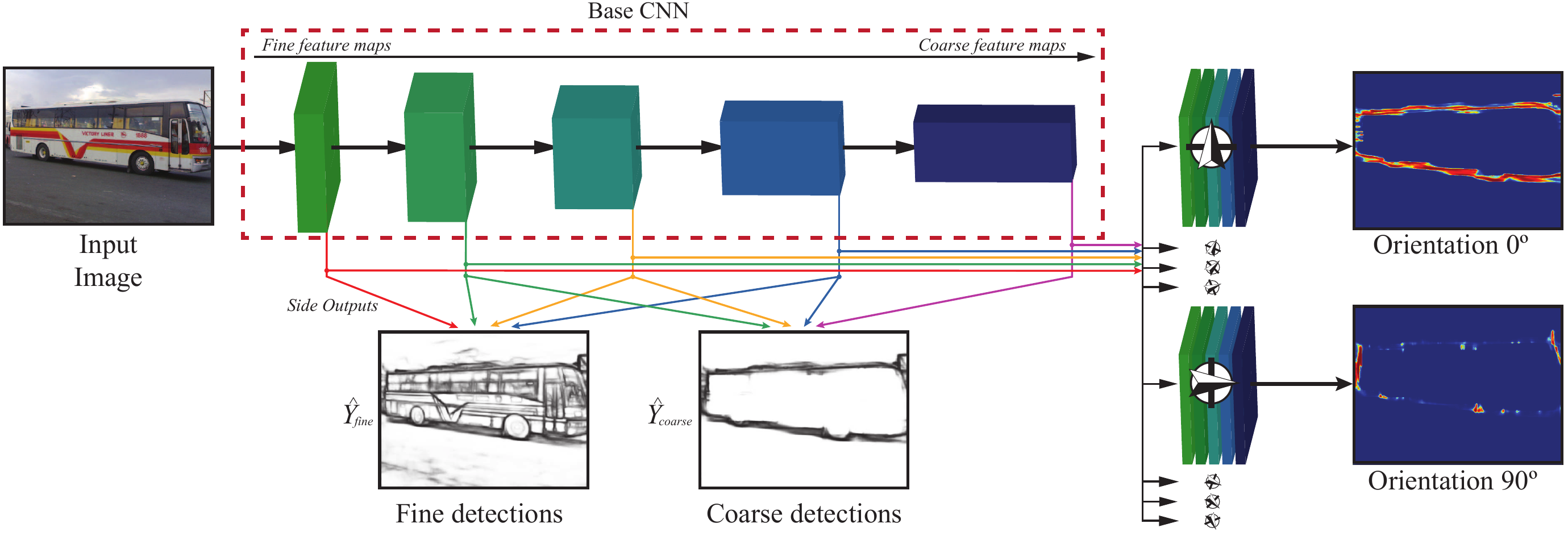}\\[-4mm]
\caption{\textbf{Our deep learning architecture} (best viewed in color). The connections show the different stages that are used to generate the multiscale contours. Orientations further require additional convolutional layers in multiple stages of the network.}
\label{fig:CNN}
\vspace{-5mm}
\end{figure*}

The latest wave of contour detectors takes advantage of deep learning to obtain state-of-the-art results~\cite{Kokkinos2016,XiTu15,BST15a,BST15b,She+15,GaLe14,BST15c}.
Ganin and Lempitsky~\cite{GaLe14} use a deep architecture to extract features of image patches.
They approach contour detection as a multi-class classification task, by matching the extracted features to predefined ground-truth features. The authors of~\cite{BST15a,BST15b} make use of features generated by pre-trained CNNs to regress contours. They prove that object-level information provides powerful cues for the prediction of contours. Shen et al.~\cite{She+15} learn deep features using shape information. Xie and Tu~\cite{XiTu15} provide an end-to-end deep framework to boost the efficiency and accuracy of contour detection, using convolutional feature maps and a novel loss function. An extended version of their work, with many additional experiments can be found in~\cite{XiTu17}.
Kokkinos~\cite{Kokkinos2016} builds upon \cite{XiTu15} and improves the results by tuning the loss function, running the detector at multiple scales, and adding globalization.

What many of the aforementioned methods have in common is that several simple components contribute to increased performance: (i) information at multiple scales~\cite{Ren08, Arb+11, Pont-Tuset2016, Kokkinos2016}, (ii) contour orientation~\cite{HaFo15, Kok10a, Arb+11, LZD13}, and (iii) end-to-end deep learning~\cite{XiTu15, Kokkinos2016}.
COB is able to combine all of the above in a single pass of a CNN, producing an output that is richer than a linear combination of cues at different scales.

At the core of all these deep learning approaches lies a \textit{base CNN}, starting from the seminal AlexNet~\cite{Krizhevsky2012} (8 layers), through the more complex VGGNet~\cite{SiZi15} (16 layers) and inception architecture of GoogLeNet~\cite{Sze+15} (22 layers), to the very recent and very deep ResNets~\cite{He+16} (up to 1001 layers).
Image classification results, which originally motivated these architectures, have been continuously improved by exploring deeper and more complex networks. In this work, we present results both using VGGNet and ResNet, showing that COB is modular and can incorporate and benefit from future improvements in the base CNN.

Recent work has also explored weakly supervised or unsupervised deep learning of contours: Khoreva et al.~\cite{Kho+16} learn from the results of generic contour detectors coupled with object detectors; and Li et al.~\cite{Li+16} train contour detectors from motion boundaries acquired from video sequences.
Yang et al.~\cite{Yan+16} use Conditional Random Fields (CRFs) to refine the inaccurately localized boundary annotations of PASCAL. 
Some works shift the domain of contours detection from abstract perceptual grouping to better defined tasks such as semantic or object contour detection~\cite{Har+11, Kho+16, Yan+16}. Some methods also combine RGB-D cues for contour detection~\cite{GAM13,Gup+14,DoZi15}. Extensive experiments on such benchmarks show that COB has an excellent performance even when shifting domains, showing state-of-the-art performance also in these new situations.

\vspace{2mm}
{\setlength{\parindent}{0cm}\paragraph*{\bf Hierarchical Image Segmentation and Grouping}
One of the most studied category of methods for image segmentation are spectral methods, that rely on the generalized eigenvalue problem to solve a low-level pixel grouping problem. Notable approaches that fall into this category are Normalized Cuts~\cite{Shi2000}, PMI~\cite{Iso+14}, gPb~\cite{Arb+11}, MCG~\cite{Pont-Tuset2016}. Arbel\'aez et al.~\cite{Arb+11} showed the usefulness for jointly optimizing contours and regions (The duality between contours and regions was first studied by Najman and Schmitt~\cite{Najman1996}). Pont-Tuset et al.~\cite{Pont-Tuset2016} leveraged multi-resolution contour detection and proved its interest for generating object proposals. 
COB also exploits the duality between contour detection and segmentation hierarchies. We differentiate from previous approaches mainly in two aspects. First, our sparse boundary representation translates into a clean and highly efficient implementation of hierarchical segmentation. Second, by leveraging high-level knowledge from the CNNs in the estimation of contour strength and orientation, our method benefits naturally from global information, which allows bypassing the globalization step (output of normalized cuts), a bottleneck in terms of computational cost, but a cornerstone of previous approaches.}
\vspace{-2mm}

\section{Deep Multiscale Oriented Contours}
\label{sec:hier_cont}
CNNs are by construction multi-scale feature extractors.  If one examines the standard architecture of a CNN consisting of convolutional and spatial pooling layers, it becomes clear that as we move deeper, feature maps capture more global information due to the decrease in resolution. For contour detection, this architecture implies local and fine-scale contours at shallow levels, coarser spatial resolution and larger receptive fields for the units when going deeper and, consequently, more global information for predicting boundary strength and orientation. CNNs have therefore a built-in globalization strategy for contour detection, analogous to the hand-engineered globalization of contour strength through spectral graph partitioning in~\cite{Arb+11,Pont-Tuset2016}.

Figure~\ref{fig:CNN} depicts how we make use of information provided by the intermediate layers of a CNN to detect contours and their orientations at multiple scales. Different groups of feature maps contain different, scale-specific information, which we combine to build a multiscale oriented contour detector.
The remainder of this section is devoted to introducing the recent approaches to contour detection using deep learning, to presenting our CNN architecture to produce contour detection at different scales, and to explaining how we estimate the orientation of the edges; all in a single CNN forward pass at the image level.

\begin{figure*}
\setlength{\fboxsep}{0pt}
%====\begin{minipage}[b]{0.48\linewidth}
\centering
\cfbox{white}{\resizebox{0.2\textwidth}{!}{%
\begin{tikzpicture}
    \node[anchor=north west,inner sep=0,use as bounding box] (image) at (0,0) {\includegraphics[width=\textwidth]{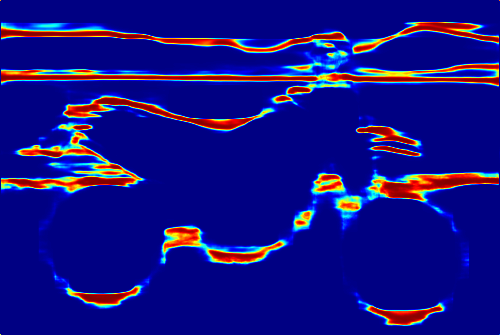}}; 
        \clip(image.north west) rectangle (image.south east);
     \node[rotate around={0:(0,0)}] at (1.2,-1.2) {\includegraphics[width=.14\textwidth]{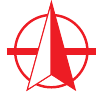}};
\end{tikzpicture}}}
\cfbox{white}{\resizebox{0.2\textwidth}{!}{%
\begin{tikzpicture}
    \node[anchor=north west,inner sep=0,use as bounding box] (image) at (0,0) {\includegraphics[width=\textwidth]{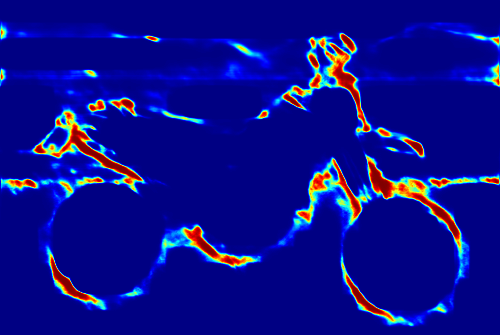}}; 
        \clip(image.north west) rectangle (image.south east);
     \node[rotate around={-45:(0,0)}] at (1.2,-1.2) {\includegraphics[width=.14\textwidth]{img/arrow.pdf}};
\end{tikzpicture}}}
\cfbox{white}{\resizebox{0.2\textwidth}{!}{%
\begin{tikzpicture}
    \node[anchor=north west,inner sep=0,use as bounding box] (image) at (0,0) {\includegraphics[width=\textwidth]{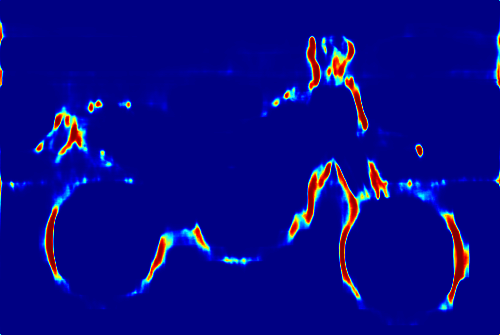}};
        \clip(image.north west) rectangle (image.south east);
     \node[rotate around={-90:(0,0)}] at (1.2,-1.2) {\includegraphics[width=.14\textwidth]{img/arrow.pdf}};
\end{tikzpicture}}}
\cfbox{white}{\resizebox{0.2\textwidth}{!}{%
\begin{tikzpicture}
    \node[anchor=north west,inner sep=0,use as bounding box] (image) at (0,0) {\includegraphics[width=\textwidth]{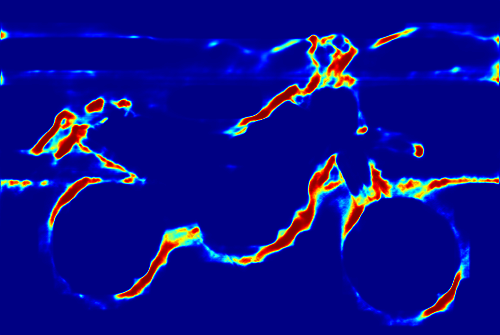}}; 
    \clip(image.north west) rectangle (image.south east);
     \node[rotate around={-135:(0,0)}] at (1.2,-1.2) {\includegraphics[width=.14\textwidth]{img/arrow.pdf}};
\end{tikzpicture}}}
\\[1mm]
\fbox{\resizebox{0.2\textwidth}{!}{%
\begin{tikzpicture}
    \node[anchor=south west,inner sep=0] (image) at (0,0) {\includegraphics[width=\textwidth]{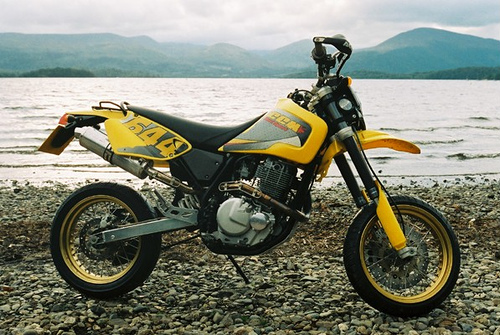}};    
\end{tikzpicture}}}
\fbox{\resizebox{0.2\textwidth}{!}{%
\begin{tikzpicture}
    \node[anchor=south west,inner sep=0] (image) at (0,0) {\includegraphics[width=\textwidth]{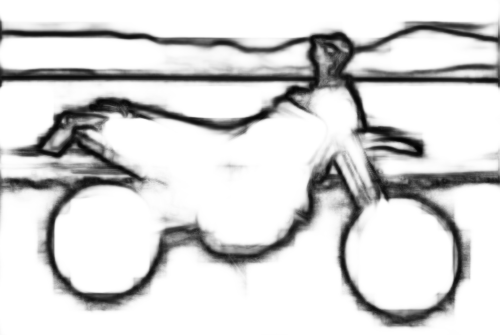}};  
\end{tikzpicture}}}
\fbox{\resizebox{0.2\textwidth}{!}{%
\begin{tikzpicture}
    \node[anchor=south west,inner sep=0] (image) at (0,0) {\includegraphics[width=0.95\textwidth]{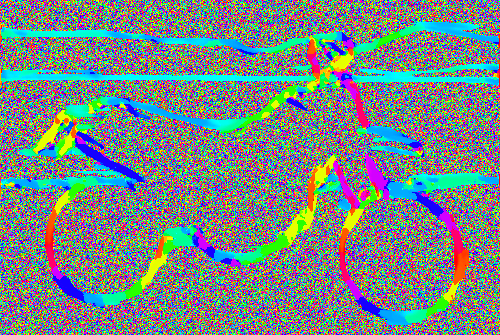}};  
\end{tikzpicture}}}
\fbox{\resizebox{0.2\textwidth}{!}{%
\begin{tikzpicture}
    \node[anchor=south west,inner sep=0] (image) at (0,0) {\includegraphics[width=\textwidth]{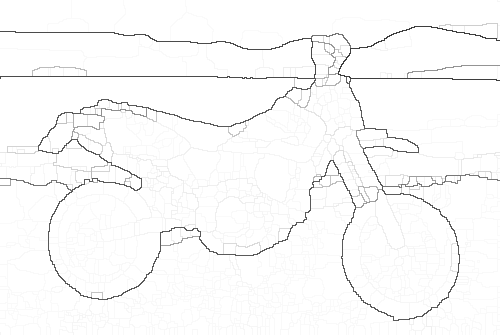}}; 
\end{tikzpicture}}} \\[-3mm]
	\caption{Illustration of contour orientation learning. Row 1 shows the responses $B_k$ for 4 out of the 8 orientation bins. Row 2, from left to right: original image, contour strength, learned orientation map into 8 orientations, and hierarchical boundaries.}
\label{fig:orientations}
%\end{minipage}
\vspace{-3mm}
\end{figure*}

{\vspace{2mm}\setlength{\parindent}{0cm}\paragraph*{\textbf{Training deep contour detectors}}
The recent success of~\cite{XiTu15} is based on a CNN to accurately regress the contours of an image. Within this framework, the idea of employing a CNN in an image-to-image fashion without any post-processing has proven successful, and lead to a big leap in performance for the task of contour detection. Their network, HED, produces scale-specific contour images (side outputs) for different scales of a network, and combines their activations linearly to produce a contour probability map. Using the notation of the authors, we denote the training dataset by $S=\lbrace \left( X_n, Y_n \right),n=1,\dots,N \rbrace$, with $X_n$ being the input image and $Y_n=\lbrace y_j^{(n)},j=1,\dots,|X_{n}| \rbrace ,  y_j^{(n)} \in \lbrace 0,1 \rbrace$  the predicted pixelwise labels. For simplicity, we drop the subscript $n$. Each of the $M$ side outputs minimizes the objective function:}
{\small
\begin{align}
\ell_{side}^{(m)}\!\left(\!\mathbf{W}\!,\!\mathbf{w}^{(m)}\!\right)=&-\beta\sum_{j\in Y_+}{\log{P\left(y_j=1 |X;\mathbf{W},\mathbf{w}^{(m)}\right)}} \nonumber \\
&-\!(1\!-\!\beta)\!\sum_{j\in Y_-}{\!\!\log{P\left(y_j\!=\!0 |X;\mathbf{W},\!\mathbf{w}^{(m)}\!\right)}} \label{eq:hed_cost_side}
\end{align}}where $\ell_{side}^{(m)}$ is the loss function for scale $m \in \lbrace 1,\dots,M \rbrace$, $\mathbf{W}$ denotes the standard set of parameters of the CNN, and $\lbrace \mathbf{w}^{(m)}, m=1,\dots,M \rbrace$ the corresponding weights of the the $m$-th side output. The multiplier $\beta$ is used to handle the imbalance of the substantially greater number of background compared to contour pixels. $Y_+$ and $Y_-$ denote the contour and background sets of the ground-truth $Y$, respectively.  The probability $P \left( \cdot \right)$ is obtained by applying a sigmoid $\sigma \left( \cdot \right)$ to the activations of the side outputs $\hat{A}_{side}^{(m)}=\lbrace a_j^{(m)}, j=1,\dots,|Y| \rbrace$. In HED, the activations are finally fused linearly, as: $\hat{Y}_{fuse} = \sigma \left( \Sigma_{m=1}^{M}h_m \hat{A}_{side}^{(m)} \right)$ where $\mathbf{h}=\lbrace h_m,m=1,\dots,M \rbrace$ are the fusion weights. The fusion output is also trained to resemble the ground-truth applying the same loss function of Equation~\ref{eq:hed_cost_side}, by optimizing the complete set of parameters, including the fusion weights $\mathbf{h_{m}}$. We instead take advantage of the common CNN architectures to regress both the strength of the coarse and detailed (fine) contours, as well as the contour orientations. COB combines these output channels non-linearly to a single hierarchical segmentation. Inside this segmentation, the placement of each region in the hierarchy is determined by the strength of the boundaries to the neighbouring regions. All in all, COB efficiently combines contour strengths and orientations into a segmentation hierarchy which can further facilitate high-level vision tasks related to segmented object proposals. In the rest of the paper we use the class-balancing cross-entropy loss function of Equation~\ref{eq:hed_cost_side}.

{\vspace{2mm}\setlength{\parindent}{0cm}\paragraph*{\textbf{Multiscale contours}}
We start from a deep network pre-trained on ImageNet~\cite{Russakovsky2015}, such as VGG~\cite{SiZi15} or ResNet~\cite{He+16}. The fully connected layers used for classification are removed, and so are the batch normalization layers, since we operate on one image per iteration. Therefore, the network consists mainly of convolutional layers coupled with ReLU activations, divided into 5 stages. We will refer to this architecture as the \textit{base CNN} of our implementation. Each stage is handled as a different scale, since it contains feature maps of a similar size. At the end of a stage, there is a max pooling layer, which reduces the spatial dimensions of the produced feature maps to a half. As discussed before, the CNN naturally contains multiscale information, which we exploit to build a multiscale contour regressor.}

We separately supervise the output of the last layer of each stage (side activation), comparing it to the ground truth using the loss function of Equation~\ref{eq:hed_cost_side}.
This way, we enforce each side activation to produce an intermediate contour map at different resolution.
The idea of supervising intermediate parts of a CNN has successfully been used in previous approaches, for a variety of tasks~\cite{Sze+15,Lee+14,XiTu15}.
In the 5-scale base CNN illustrated in Figure~\ref{fig:CNN}, we linearly combine the side activations 
of the 4 finest and 4 coarsest scales to a fine-scale and a coarse-scale output ($\hat{Y}_{fine}$ and $\hat{Y}_{coarse}$, respectively) with trainable weights.
The finer scale contains better localized contours, whereas the coarse scale leads to less noisy detections.
To train the two sets of weights of the linear combinations, we freeze the pre-trained weights of the base CNN. 

{\vspace{2mm}\setlength{\parindent}{0cm}\paragraph*{\textbf{Estimation of Contour Orientations}}
In order to predict accurate contour orientations, we propose an extension of the CNN that we use to predict contour strength. We define the task as pixel-wise image-to-image multiscale classification into $K$ bins. We connect $K$ different branches (sub-networks) to the base network, each of which is associated with one orientation bin, and has access to feature maps that are generated from the intermediate convolutional layers at $M$ different scales. We assign the parts of the CNN associated with each orientation a different task from the base network: 
classify the pixels of the contours that match a specific orientation. In order to design these orientation-specific subtasks, we classify each pixel of the human contour annotations into $K$ different orientations. The orientation of each contour pixel is obtained by approximating the ground-truth boundaries with polygons, and assigning each pixel the orientation of the closest polygonal segment, as shown in Figure~\ref{fig:pol_simpl}. As in the case of multiscale contours, the weights of the base network remain frozen when training these sub-networks.}

Each sub-network consists of $M$ convolutional layers, each of them appended on different scales of the base network. Thus we need $M*K$ additional layers. In our setup, we use $K=8$ and $M=5$. All $K$ orientations are regressed in parallel, and since they are associated with a certain angle, we post-process them to obtain the orientation map. Specifically, the orientation map is obtained as:
\begin{equation}
O(x,y) = \mathcal{T}\left(\arg\max_k{B_k \left(x,y \right)}\right),  k=1,\dots,K
\end{equation}
where $B_k(x,y)$ denotes the response of the $k$-th orientation bin of the CNN at the pixels with coordinates  $(x,y)$  and $\mathcal{T}\left(\cdot\right)$ is the transformation function which associates each bin with its central angle. For the cases where two neighboring bins lead to strong responses, we compute the angle as their weighted average. At pixels where there is no response for any of the orientations, we assign random values between $0$ and $\pi$, not to bias the orientations. The different orientations as well as the resulting orientation map (color-coded) are illustrated in Figure~\ref{fig:orientations}. 

In \cite{Arb+11,DoZi15,Pont-Tuset2016} the orientations are computed by means of local gradient filters.  In Section~\ref{sec:experiments} we show that our learned orientations are significantly more accurate and lead to better region segmentations.

\section{Fast Hierarchical Regions}
\label{sec:fast_mcg}
This section is devoted to building an efficient hierarchical image segmentation algorithm from the multiscale contours and the orientations extracted in the previous section.
We build on the concept of Ultrametric Contour Map (UCM)~\cite{Arb+11}, which transforms a contour detection probability map into a hierarchical boundary map, which gets partitions at different granularities when thresholding at various contour strength values.
Despite the success of UCMs, their low speed limits their applicability.
We address this issue by using an alternative representation of an image partition which reduces the computation time of UCMs by an order of magnitude.

\begin{figure}[b]
\centering
\resizebox{\linewidth}{!}{\includegraphics{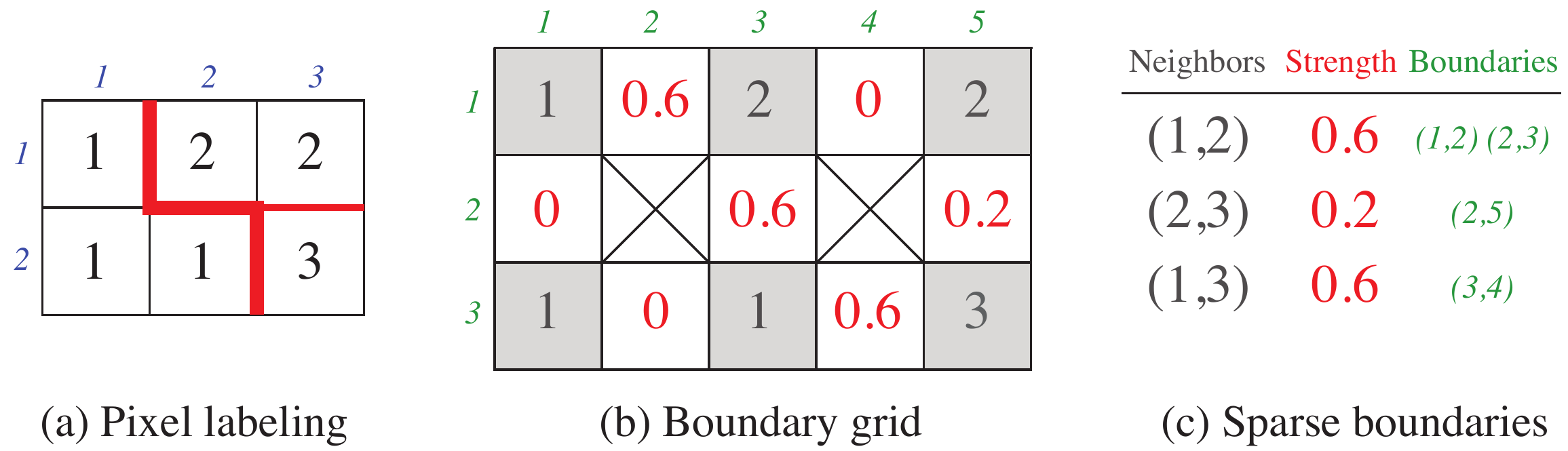}}
\vspace{-5mm}
\caption{\textbf{Image Partition Representation:}\newline(a) Pixel labeling, each pixel gets assigned a region label. (b) Boundary grid, markers of the boundary positions. (c) Sparse boundaries, lists of boundary coordinates between neighboring regions.}
\label{fig:part_rep}
\end{figure}

{\vspace{2mm}\setlength{\parindent}{0cm}\paragraph*{\textbf{Sparse Boundary Representation of Hierarchies of Regions}}
An image partition is a clustering of the set of pixels into different sets,
which we call regions.
The most straightforward way of representing it in a computer is by a matrix of labels, as in the
example in Figure~\ref{fig:part_rep}(a), with three regions on an image of size 2$\times$3.
The boundaries of this partition are the edge elements, or \textit{edgels}, between the pixels with different
labels (highlighted in red).
We can assign different \textit{strengths} to these boundaries (thicknesses of the red lines), which indicate the \textit{confidence} of that piece of being a boundary.
By iteratively \textit{erasing} these boundaries in order of increasing strength
we obtain different partitions, which we call \textit{hierarchy of regions}, or Ultrametric Contour Maps.
}

These boundaries are usually stored in the \textit{boundary grid} (Figure~\ref{fig:part_rep}(b)), a matrix
of double the size of the image (minus one), in which the odd coordinates represent pixels (gray areas), and the positions in between represent boundaries (red numbers) and junctions (crossed positions).
UCMs use this representation to store their boundary \textit{strength} values,
that is, each boundary position stores the threshold value beyond which that edgel \textit{disappears} and the two neighboring regions merge.
This way, simply \textit{binarizing} a UCM we have a partition represented as a boundary grid.
Continuing with the example in Figure~\ref{fig:part_rep}, binarizing the UCM at 0.5 the edge between region 2 and 3 would disappear, that is, 2 and 3 would merge and create a new region.

This representation becomes very inefficient at run time, where the percentage of \textit{activated} boundaries is very sparse.
Not only are we wasting memory by storing those \textit{empty} boundaries, but it also makes 
operating on them very inefficient by having to \textit{sweep} over the entire matrix 
to perform a modification on a single boundary piece.

Inspired by how sparse matrices are handled, we designed the \textit{sparse boundaries} representation
(Figure~\ref{fig:part_rep}(c)).
It stores a look-up table for pairs of neigboring regions, their boundary strength, and the list of coordinates the boundary occupies.
Apart from being more compact in terms of memory, this representation enables efficient operations on specific pieces of a boundary, since one only needs to perform a
search in the look-up table and scan the activated coordinates; instead of sweeping the whole boundary grid.

{\vspace{2mm}\setlength{\parindent}{0cm}\paragraph*{\textbf{Fast Hierarchies from Multiscale Oriented Contours}}
We are inspired by the framework proposed in~\cite{Pont-Tuset2016},
in which a UCM is obtained from contours computed at different image scales and then combined into a single hierarchy.
The motivation behind this work is that the UCMs obtained from downscaled images will focus on the coarse structures and ignore textures, so their localization accuracy will decrease.
On the other hand, upscaled images will bring very good localization in the boundaries, but it will be harder to distinguish between the high- and low-level contents.
To bring the best of the two worlds, \cite{Pont-Tuset2016} progressively \textit{projects} the coarse hierarchies into the finer ones by adapting the high-level contours into the better localized ones.
The final hierarchy keeps the high-level information while being snapped to the correctly localized low-level boundaries.}

The deep CNN presented in Section~\ref{sec:hier_cont} provides different levels of detail for the image contours, so instead of processing the image at multiple resolutions we use the different outputs that are computed in a single pass of the CNN to obtain different hierarchies that focus on high- and low-level features.

A drawback of the original framework~\cite{Pont-Tuset2016}, however, is that the manipulation of the hierarchies and their projection to different scales is very slow (in the order of seconds), so the operations on the UCMs had to be performed at a small subset of the contour strengths (from thousands to a few dozens).
By using the fast sparse boundary representation, we can operate on all thousands of contour strengths, yielding better results at a fraction of the original cost.
Moreover, we use the learned contour orientations for the computation of the Oriented Watershed Transform (OWT)~\cite{Arb+11}, further boosting performance.

\section{Experiments on Low-Level Applications}
\label{sec:experiments}
This section presents the empirical evidence that supports our approach for low-level applications (image segmentation and contour detection).
First, Section~\ref{sec:exp:ablat} explores ablated and baseline techniques in order to isolate and quantify the improvements due to different components of our system.
Section~\ref{sec:exp:orient} further analyzes and evaluates the proposed contour orientations. In Section~\ref{sec:exp:generic}, Section~\ref{sec:exp:obj_boundary}, and Section~\ref{sec:exp:rgbd_boundary} we compare our results against the state of the art in generic RGB image segmentation, RGB object boundary detection, and RGB-D image segmentation, respectively. In all three cases, we obtain the best results to date by a significant margin. Finally, Section~\ref{sec:exp:speed} analyzes the effect of the various components in terms of speed on COB.

In terms of datasets, we extend the main BSDS benchmark~\cite{Martin2004} to the PASCAL Context dataset~\cite{Mot+14}, which contains carefully localized pixel-wise semantic annotations for the entire image on the PASCAL VOC 2010 detection train-val set. 
This results in 459 semantic categories across 10\,103 images, which is an order of magnitude (20$\times$) larger than the BSDS. In order to allow training and optimization of large capacity models, we split the data into train, validation, and test sets as follows: \emph{VOC train} corresponds to the official PASCAL Context train with 4\,998 images, \emph{VOC val} corresponds to half the official PASCAL Context validation set with 2\,607 images and \emph{VOC test} corresponds to the second half with 2\,498 images. In the remainder of the paper, we refer to this dataset division. Note that, in contrast to the BSDS, in this setting boundaries are defined between different semantic categories and not between their parts.

In all our experiments for boundary detection and image segmentation, we used the standard evaluation benchmark evaluating boundaries ($F_b$~\cite{Martin2004}) and regions ($F_\mathit{op}$~\cite{Pont-Tuset2016a}). Through the literature, the tolerance in the boundary localization metric $F_b$ is altered
(the \texttt{maxDist} parameter), depending on the database and the quality of the annotations.
To avoid confusion, we list the value of this parameter for all our experiments in Table~\ref{tab:maxDist}. Please also note that methods that produce open contours instead of regions can not be evaluated using the region measure $F_{op}$. In all the produced curves, markers indicate the optimal operating point that maximizes $F_b$ and $F_{op}$.
We used the publicly available \textit{Caffe}~\cite{Jia+14} framework for training and testing CNNs, and all the state-of-the-art results are computed using the publicly-available code provided by the respective authors.

\begin{table}[b]
\setlength{\tabcolsep}{0.2em}
\resizebox{\linewidth}{!}{%
\begin{tabular}{@{\hspace{2mm}}l@{\hspace{2mm}}|@{\hspace{2mm}}l@{\hspace{2mm}}r@{\hspace{2mm}}r@{\hspace{2mm}}r@{\hspace{2mm}}}
\toprule
Database 		  & Task		 			&    train  	 &     test    		  & \texttt{maxDist}    \\ \midrule
BSDS500           & Generic Segmentation   	&    \ \,\, 300    &    \ \, \,\, 200     & \ \, \, 0.0075 		\\
VOC Context       & Generic	Segmentation   	&    \ 7\,605 	 &    \ \, 2\,498 	  & \ \, \, 0.0075 		\\
VOC'12 Segm.      & Object Contours			&    \ 1\,464 	 &    \ \, 1\,449 	  & \ 0.01 				\\
NYUD-v2    		  & RGB-D Segmentation   	&	 \ \,\, 795    &    \ \, \,\, 654 	  & \ \, 0.011 			\\
\midrule
SBD				  & Semantic Contours		&    \ 8\,498 	 &    \ \, 2\,857 	  & \ 0.02 				\\
VOC'12 Segm.      & Semantic Segmentation	&    \ 1\,464 	 &    \ \, 1\,449 	  & \ - 				\\
COCO			  & Object Proposals		&     - 	  	 &     40\,504  		  & \ - 				\\
VOC'07			  & Object Detection		&    \ 5\,011 	 &    \ \, 4\,952 	  & \ - 				\\
 \bottomrule
\end{tabular}}
\vspace{2mm}
\caption{\textbf{Datasets and Parameters}: The list of databases used to evaluate our approach on various low-level and high-level tasks. We report the number of images used for training and testing our algorithm, along with the tolerance for contour localization used in the literature, when applicable. In all our experiments, we keep those numbers unchanged.}
\label{tab:maxDist}
\end{table}

\paragraph*{\textbf{Training details}} In our two-step training approach, we first train the base networks for the task of contour detection (coarse and fine). We use stochastic gradient descent with a momentum of 0.9 and weight decay of 0.0002 for 40k iterations. The base learning rate is set to $10^{-6}$, and is divided by 10 after 30k iterations. After the first step is finished, the weights of the base network are frozen, and the layers of the orientation sub-network are connected and trained for an additional 10k iterations. Depending to the size of dataset we use different data augmentation strategies: flipping and rotation into 4 angles for PASCAL and NYUD-v2; flipping, rotation into 16 angles, and scaling into 3 scales~\cite{XiTu15} for BSDS500.
In all cases, we initialize the network from ImageNet pre-trained weights. The same ground-truth boundaries are used for training both the fine and the coarse contours.

\subsection{Control Experiments/Ablation Analysis}
\label{sec:exp:ablat}
This section presents the control experiments and ablation analysis to assess the performance of all subsystems of our method.
We train on \emph{VOC train}, and evaluate on \emph{VOC val} set.
We report the standard F measure at Optimal Dataset Scale (ODS) and Optimal Image Scale (OIS), as well as the Average Precision (AP), both evaluating boundaries ($F_b$~\cite{Martin2004}) and regions ($F_\mathit{op}$~\cite{Pont-Tuset2016a}).

Table~\ref{table:ablation} shows the evaluation results of the different variants, 
highlighting whether we include globalization and/or trained orientations. 
As a first baseline, we test the performance of MCG~\cite{Pont-Tuset2016}, which uses Structured Edges~\cite{DoZi15} as input contour signal.
We then substitute SE by the newer HED~\cite{XiTu15}, trained on \emph{VOC train} as input contours and denote it MCG-HED.
Note that the aforementioned baselines require multiple passes of the contour detector (3 scales).

In the direction of using the side outputs of the base CNN architecture as multiscale contour detections in one pass, we tested the baseline of naively taking the 5 side outputs 
directly as the contour detections. We trained both VGGNet~\cite{SiZi15} and ResNet50~\cite{He+16} on \emph{VOC train} and combined the 5 side outputs with 
our fast hierarchical regions of Section~\ref{sec:fast_mcg} (VGGNet-Side and ResNet50-Side).

We finally evaluate different variants of our system, as presented in Section~\ref{sec:hier_cont}. 
We first compare our system with two different base architectures: Ours(VGGNet) and Ours(ResNet50). We observe that the deeper architecture of ResNet translates into better boundaries and regions. Using the even deeper counterparts of ResNet lead to negligible gain in accuracy while significantly sacrificing speed.

We then evaluate the influence of our trained orientations and globalization, by testing 
the four possible combinations (the orientations are further evaluated in the next section).
Our method using ResNet50 together with trained orientations leads to the best results both for boundaries and for regions.
The experiments also show that, when coupled with trained orientations, globalization even decreases performance, so we can safely remove it and get a significant speed up. This behaviour arises from the fact that the image-to-image architecture of the base CNN already captures global information, addressing issues that could not be handled by local approaches, e.g., deleting internal contours of objects.
Our technique with trained orientations and without globalization is therefore selected as our final system and will be referred to in the sequel as Convolutional Oriented Boundaries (COB).

\begin{table}[b]
\setlength{\tabcolsep}{4pt} % General space between cols (6pt standard)
\center
\footnotesize
%\rowcolors{3}{rowblue}{white}
\resizebox{\linewidth}{!}{%
\begin{tabular}{lcccccccc}
\toprule
       & & &\multicolumn{3}{c}{Boundaries - $F_b$} & \multicolumn{3}{c}{Regions - $F_\mathit{op}$}\\
Method & Global. & Orient. & ODS & OIS & AP & ODS & OIS & AP \\
\midrule
MCG~\cite{Pont-Tuset2016}   & {\color{gray}\ding{51}} &  {\color{gray}\ding{55}} & %
\mbox{0.548\hspace{-2.5pt}}%
 & %
\mbox{0.594
\hspace{-2.5pt}}%
 & %
\mbox{0.519
\hspace{-2.5pt}}%
 & %
\mbox{0.355\hspace{-2.5pt}}%
 & %
\mbox{0.419
\hspace{-2.5pt}}%
 & %
\mbox{0.263
\hspace{-2.5pt}}%
 \\
MCG-HED  & {\color{gray}\ding{51}} &  {\color{gray}\ding{55}} & %
\mbox{0.691\hspace{-2.5pt}}%
 & %
\mbox{0.727
\hspace{-2.5pt}}%
 & %
\mbox{0.693
\hspace{-2.5pt}}%
 & %
\mbox{0.459\hspace{-2.5pt}}%
 & %
\mbox{0.520
\hspace{-2.5pt}}%
 & %
\mbox{0.374
\hspace{-2.5pt}}%
\\
\midrule
VGGNet-Side & {\color{gray}\ding{51}} & {\color{gray}\ding{55}} & %
\mbox{0.644\hspace{-2.5pt}}%
 & %
\mbox{0.683
\hspace{-2.5pt}}%
 & %
\mbox{0.664
\hspace{-2.5pt}}%
 & %
\mbox{0.439\hspace{-2.5pt}}%
 & %
\mbox{0.505
\hspace{-2.5pt}}%
 & %
\mbox{0.351
\hspace{-2.5pt}}%
\\
ResNet50-Side & {\color{gray}\ding{51}} & {\color{gray}\ding{55}} & %
\mbox{0.676\hspace{-2.5pt}}%
 & %
\mbox{0.711
\hspace{-2.5pt}}%
 & %
\mbox{0.681
\hspace{-2.5pt}}%
 & %
\mbox{0.456\hspace{-2.5pt}}%
 & %
\mbox{0.521
\hspace{-2.5pt}}%
 & %
\mbox{0.374
\hspace{-2.5pt}}%
\\
\midrule
Ours (VGGNet) & \ding{55} & \ding{51} & %
\mbox{0.705\hspace{-2.5pt}}%
 & %
\mbox{0.735
\hspace{-2.5pt}}%
 & %
\mbox{0.741
\hspace{-2.5pt}}%
 & %
\mbox{0.466\hspace{-2.5pt}}%
 & %
\mbox{0.533
\hspace{-2.5pt}}%
 & %
\mbox{0.384
\hspace{-2.5pt}}%
\\
\midrule
Ours (ResNet50) & \ding{55} & \ding{55} & %
\mbox{0.734\hspace{-2.5pt}}%
 & %
\mbox{0.767
\hspace{-2.5pt}}%
 & %
\mbox{0.757
\hspace{-2.5pt}}%
 & %
\mbox{0.475
\hspace{-2.5pt}}%
 & %
\mbox{0.545
\hspace{-2.5pt}}%
 & %
\mbox{0.405
\hspace{-2.5pt}}%
\\
Ours (ResNet50) & \ding{51} & \ding{55} & %
\mbox{0.726\hspace{-2.5pt}}%
 & %
\mbox{0.759
\hspace{-2.5pt}}%
 & %
\mbox{0.725
\hspace{-2.5pt}}%
 & %
\mbox{0.461
\hspace{-2.5pt}}%
 & %
\mbox{0.531
\hspace{-2.5pt}}%
 & %
\mbox{0.395
\hspace{-2.5pt}}%
\\
Ours (ResNet50) & \ding{51} & \ding{51} & %
\mbox{0.732\hspace{-2.5pt}}%
 & %
\mbox{0.763
\hspace{-2.5pt}}%
 & %
\mbox{0.731
\hspace{-2.5pt}}%
 & %
\mbox{0.481
\hspace{-2.5pt}}%
 & \bf%
\mbox{0.554
\hspace{-2.5pt}}%
 & \bf%
\mbox{0.418
\hspace{-2.5pt}}%
\\
Ours (ResNet50) & \ding{55} & \ding{51} & \bf%
\mbox{0.737\hspace{-2.5pt}}%
 & \bf%
\mbox{0.768
\hspace{-2.5pt}}%
 & \bf%
\mbox{0.758
\hspace{-2.5pt}}%
 & \bf%
\mbox{0.483
\hspace{-2.5pt}}%
 & %
\mbox{0.553
\hspace{-2.5pt}}%
 & %
\mbox{0.417
\hspace{-2.5pt}}%
\\
\bottomrule
\end{tabular}}
\vspace{2mm}
\caption{\textbf{Ablation analysis on \emph{VOC val}}: Comparison of different ablated and baseline versions of our system.}
\label{table:ablation}
\end{table}

\subsection{Contour Orientation}
\label{sec:exp:orient}
\begin{figure}[h]
\centering
\resizebox{0.8\linewidth}{!}{\includegraphics{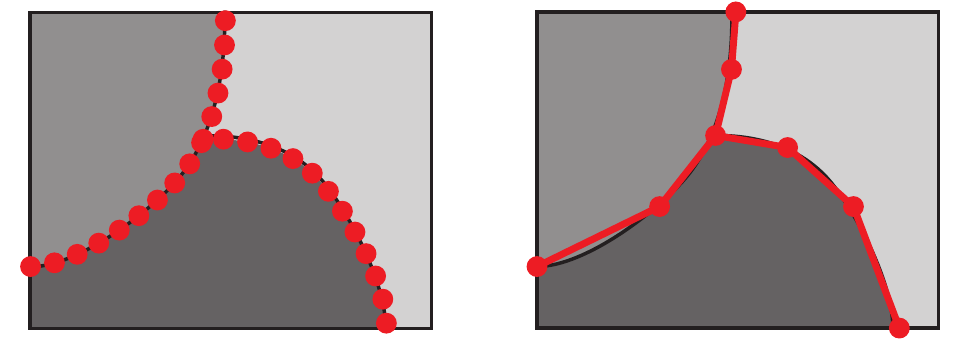}}
\vspace{-3mm}
\caption{\textbf{Polygon simplification:} From all boundary points (left) to simplified polygons (right), which are used to compute the ground-truth orientation robustly.}
\label{fig:pol_simpl}
\end{figure}

We evaluate contour orientation results by the classification accuracy into 8 different orientations, to isolate their performance from the global system.
We compute the ground-truth orientations as depicted in Figure~\ref{fig:pol_simpl} by means of the sparse boundaries representation.
We then sweep all ground-truth boundary pixels and compare the estimated 
orientation with the ground-truth one.
Since the orientations are not well-balanced classes (much more horizontal
and vertical contours), we compute the classification accuracy per each of the 8 classes and then compute the mean.

\begin{figure}[h]
\centering
\scalebox{0.8}{%
\begin{tikzpicture}[/pgfplots/width=0.9\linewidth, /pgfplots/height=0.7\linewidth]
    \begin{axis}[ymin=5,ymax=70,xmin=1,xmax=100,enlargelimits=false,
        xlabel=Confidence percentile (\%),
        ylabel=Classification accuracy (\%),
        font=\scriptsize, grid=both,
        grid style=dotted,
        axis equal image=false,
       	% Legend
		legend cell align=left,
        legend style={at={(0.25,0.16)},anchor=south west},       
        ytick={0,10,...,100},
        xtick={0,10,...,100},
        minor ytick={5,10,...,100},
        minor xtick={5,10,...,100},
		major grid style={white!20!black},
        minor grid style={white!70!black},
        xlabel shift={-3pt},
        ylabel shift={-4pt},
		]
	    
        \addplot+[black,solid,mark=none, ultra thick]                        table[x=Percentile,y expr=100-\thisrow{Trained}] {data/orient/per_class_err.txt};
        \addlegendentry{COB (Ours)}

        \addplot+[red,solid,mark=none, ultra thick]                          table[x=Percentile,y expr=100-\thisrow{Local}] {data/orient/per_class_err.txt};
        \addlegendentry{Local gradients~\cite{Arb+11,DoZi15,Pont-Tuset2016}}
        
        \addplot+[blue,dashed,mark=none, ultra thick]                        table[x=Percentile,y expr=100-\thisrow{Random}] {data/orient/per_class_err.txt};
        \addlegendentry{Random}
	\end{axis}
   \end{tikzpicture}}
   \vspace{-3mm}
\caption{\textbf{Contour orientation}: Classification accuracy into 8 bins.}
\label{fig:orient_eval}
\end{figure}
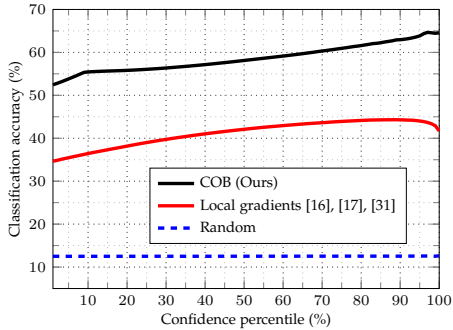

Figure~\ref{fig:orient_eval} shows the classification accuracy with respect to
the confidence of the estimation.
We compare our proposed technique against the local gradient estimation used in previous literature~\cite{Arb+11,DoZi15,Pont-Tuset2016}. As a baseline, we plot the result a random guess of the orientations would get.
We observe that our estimation is significantly better than the previous approach.
As a summary measure, we compute the area under the curve of the accuracy (ours 58.6\%, local gradients 41.2\%, random 12.5\%), which corroborates the superior results from our technique.

\begin{figure*}
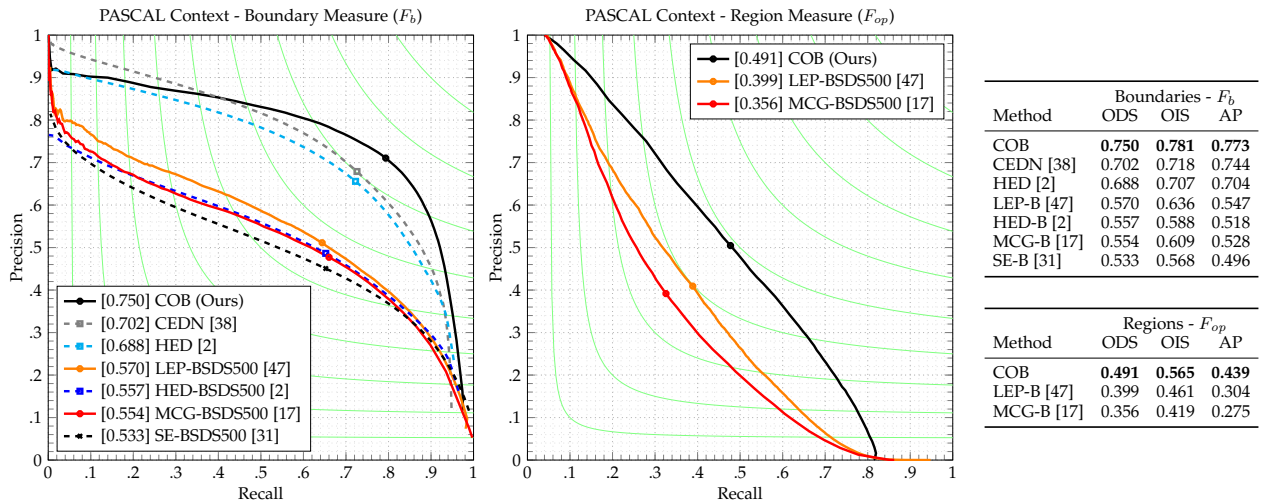

\centering
\begin{minipage}{0.70\linewidth}
\resizebox{0.5\linewidth}{!}{%
\begin{tikzpicture}[/pgfplots/width=0.75\linewidth, /pgfplots/height=0.75\linewidth]
    \begin{axis}[% Axis labels
                 ymin=0,ymax=1,xmin=0,xmax=1,
    			 % Axis labels
        		 xlabel=Recall,
        		 ylabel=Precision,
         		 xlabel shift={-2pt},
        		 ylabel shift={-3pt},
         		 % General appearance
		         font=\small,
		         axis equal image=true,
		         enlargelimits=false,
		         clip=true,
		         % Grids 
        	     grid style=dotted, grid=both,
                 major grid style={white!65!black},
        		 minor grid style={white!85!black},
		 		 xtick={0,0.1,...,1.1},
        		 ytick={0,0.1,...,1.1},
         		 minor xtick={0,0.02,...,1},
		         minor ytick={0,0.02,...,1},
		         xticklabels={0,.1,.2,.3,.4,.5,.6,.7,.8,.9,1},
		         yticklabels={0,.1,.2,.3,.4,.5,.6,.7,.8,.9,1},
        		 % Legend
				 legend cell align=left,
        		 legend style={at={(0.02,0.02)},anchor=south west},
		         % Title
		         title style={yshift=-1ex,},
		         title={PASCAL Context - Boundary Measure ($F_b$)}]
        
    % Iso-f curves
    \foreach \f in {0.1,0.2,...,0.9}{%
       \addplot[white!50!green,line width=0.2pt,domain=(\f/(2-\f)):1,samples=200,forget plot]{(\f*x)/(2*x-\f)};
    }
    
	% COB trainval
    \addplot+[black,solid,mark=none, line width=\deflinewidth,forget plot] table[x=Recall,y=Precision] {data/pr/PASCALContext_test_new_fb_COB_trainval.txt};
    \addplot+[black,solid,mark=o, mark size=1.3, mark options={solid},line width=\deflinewidth] table[x=Recall,y=Precision] {data/pr/PASCALContext_test_new_fb_COB_trainval_ods.txt};
    \addlegendentry{[\showodsf{PASCALContext_test_new}{fb}{COB_trainval}] COB (Ours)}
       
    % CEDN
    \addplot+[gray,dashed,mark=none, line width=\deflinewidth,forget plot] table[x=Recall,y=Precision] {data/pr/PASCALContext_test_new_fb_CEDN.txt};
    \addplot+[gray,dashed,mark=square,mark options={solid}, mark size=1.25, line width=\deflinewidth] table[x=Recall,y=Precision] {data/pr/PASCALContext_test_new_fb_CEDN_ods.txt};
    \addlegendentry{[\showodsf{PASCALContext_test_new}{fb}{CEDN}] CEDN~\cite{Yan+16}}
     
	% HED
    \addplot+[cyan,dashed,mark=none, line width=\deflinewidth,forget plot] table[x=Recall,y=Precision] {data/pr/PASCALContext_test_new_fb_HED_trainval.txt};
    \addplot+[cyan,dashed,mark=square,mark options={solid}, mark size=1.25, line width=\deflinewidth] table[x=Recall,y=Precision] {data/pr/PASCALContext_test_new_fb_HED_trainval_ods.txt};
    \addlegendentry{[\showodsf{PASCALContext_test_new}{fb}{HED_trainval}] HED~\cite{XiTu15}}
    
    % LEP-BSDS500
    \addplot+[orange,solid,mark=none, line width=\deflinewidth,forget plot] table[x=Recall,y=Precision] {data/pr/PASCALContext_test_new_fb_LEP-BSDS500.txt};
    \addplot+[orange,solid,mark=o, mark size=1.3, mark options={solid},line width=\deflinewidth] table[x=Recall,y=Precision] {data/pr/PASCALContext_test_new_fb_LEP-BSDS500_ods.txt};
    \addlegendentry{[\showodsf{PASCALContext_test_new}{fb}{LEP-BSDS500}] LEP-BSDS500~\cite{Zhao2015}}
     
    % HED-BSDS500
    \addplot+[blue,dashed,mark=none, line width=\deflinewidth,forget plot] table[x=Recall,y=Precision] {data/pr/PASCALContext_test_new_fb_HED-BSDS500.txt};
    \addplot+[blue,dashed,mark=square, mark size=1.25,mark options={solid},line width=\deflinewidth] table[x=Recall,y=Precision] {data/pr/PASCALContext_test_new_fb_HED-BSDS500_ods.txt};
    \addlegendentry{[\showodsf{PASCALContext_test_new}{fb}{HED-BSDS500}] HED-BSDS500~\cite{XiTu15}}
    
	% MCG-BSDS500
    \addplot+[red,solid,mark=none, line width=\deflinewidth,forget plot] table[x=Recall,y=Precision] {data/pr/PASCALContext_test_new_fb_MCG-BSDS500.txt};
    \addplot+[red,solid,mark=o, mark size=1.3, mark options={solid},line width=\deflinewidth] table[x=Recall,y=Precision] {data/pr/PASCALContext_test_new_fb_MCG-BSDS500_ods.txt};
    \addlegendentry{[\showodsf{PASCALContext_test_new}{fb}{MCG-BSDS500}] MCG-BSDS500~\cite{Pont-Tuset2016}}
    
    % SE-BSDS500
    \addplot+[black,dashed,mark=none,line width=\deflinewidth,forget plot] table[x=Recall,y=Precision] {data/pr/PASCALContext_test_new_fb_SE-BSDS500.txt};
    \addplot+[black,dashed,mark=x, mark size=1.6, mark options={solid},line width=\deflinewidth] table[x=Recall,y=Precision] {data/pr/PASCALContext_test_new_fb_SE-BSDS500_ods.txt};
    \addlegendentry{[\showodsf{PASCALContext_test_new}{fb}{SE-BSDS500}] SE-BSDS500~\cite{DoZi15}}

    \end{axis}
\end{tikzpicture}}
\hspace{-1.5mm}
\resizebox{0.5\linewidth}{!}{%
\begin{tikzpicture}[/pgfplots/width=0.75\linewidth, /pgfplots/height=0.75\linewidth]
    \begin{axis}[% Axis labels
                 ymin=0,ymax=1,xmin=0,xmax=1,
    			 % Axis labels
        		 xlabel=Recall,
        		 ylabel=Precision,
         		 xlabel shift={-2pt},
        		 ylabel shift={-3pt},
         		 % General appearance
		         font=\small,
		         axis equal image=true,
		         enlargelimits=false,
		         clip=true,
		         % Grids 
        	     grid style=dotted, grid=both,
                 major grid style={white!65!black},
        		 minor grid style={white!85!black},
		 		 xtick={0,0.1,...,1.1},
        		 ytick={0,0.1,...,1.1},
         		 minor xtick={0,0.02,...,1},
		         minor ytick={0,0.02,...,1},
		         xticklabels={0,.1,.2,.3,.4,.5,.6,.7,.8,.9,1},
		         yticklabels={0,.1,.2,.3,.4,.5,.6,.7,.8,.9,1},
		         % Legend
				 legend cell align=left,
        		 legend style={at={(0.98,0.98)}, anchor=north east},
		         % Title
		         title style={yshift=-1ex,},
		         title={PASCAL Context - Region Measure ($F_{op}$)}]
        
    % Iso-f curves
    \foreach \f in {0.1,0.2,...,0.9}{%
       \addplot[white!50!green,line width=0.2pt,domain=(\f/(2-\f)):1,samples=200,forget plot]{(\f*x)/(2*x-\f)};
    }
	
	% COB_trainval
    \addplot+[black,solid,mark=none, line width=\deflinewidth,forget plot] table[x=Recall,y=Precision] {data/pr/PASCALContext_test_new_fop_COB_trainval.txt};
    \addplot+[black,solid,mark=o, mark size=1.3, mark options={solid},line width=\deflinewidth] table[x=Recall,y=Precision] {data/pr/PASCALContext_test_new_fop_COB_trainval_ods.txt};
    \addlegendentry{[\showodsf{PASCALContext_test_new}{fop}{COB_trainval}] COB (Ours)}
    
    % LEP-BSDS500
    \addplot+[orange,solid,mark=none, line width=\deflinewidth,forget plot] table[x=Recall,y=Precision] {data/pr/PASCALContext_test_new_fop_LEP-BSDS500.txt};
    \addplot+[orange,solid,mark=o, mark size=1.3, mark options={solid},line width=\deflinewidth] table[x=Recall,y=Precision] {data/pr/PASCALContext_test_new_fop_LEP-BSDS500_ods.txt};
    \addlegendentry{[\showodsf{PASCALContext_test_new}{fop}{LEP-BSDS500}] LEP-BSDS500~\cite{Zhao2015}}
	
    % MCG-BSDS500
    \addplot+[red,solid,mark=none, line width=\deflinewidth,forget plot] table[x=Recall,y=Precision] {data/pr/PASCALContext_test_new_fop_MCG-BSDS500.txt};
    \addplot+[red,solid,mark=o, mark size=1.3, mark options={solid},line width=\deflinewidth] table[x=Recall,y=Precision] {data/pr/PASCALContext_test_new_fop_MCG-BSDS500_ods.txt};
    \addlegendentry{[\showodsf{PASCALContext_test_new}{fop}{MCG-BSDS500}] MCG-BSDS500~\cite{Pont-Tuset2016}}

    \end{axis}
\end{tikzpicture}}
\end{minipage}
\hspace{1mm}
\begin{minipage}{0.20\linewidth}
\setlength{\tabcolsep}{4pt} % General space between cols (6pt standard)
\center
\footnotesize
%\rowcolors{3}{rowblue}{white}
\resizebox{\textwidth}{!}{%
\begin{tabular}{lccc}
\toprule
       & \multicolumn{3}{c}{Boundaries - $F_b$} \\
Method &  ODS & OIS & AP\\
\midrule
COB   & \bf%
\mbox{\input{data/pr/PASCALContext_test_new_fb_COB_trainval_ods_f.txt}\hspace{-2.5pt}}%
 & \bf%
\mbox{\input{data/pr/PASCALContext_test_new_fb_COB_trainval_ois_f.txt}\hspace{-2.5pt}}%
 & \bf%
\mbox{\input{data/pr/PASCALContext_test_new_fb_COB_trainval_ap.txt}\hspace{-2.5pt}}%
 \\
CEDN~\cite{Yan+16}   & %
\mbox{\input{data/pr/PASCALContext_test_new_fb_CEDN_ods_f.txt}\hspace{-2.5pt}}%
 & %
\mbox{\input{data/pr/PASCALContext_test_new_fb_CEDN_ois_f.txt}\hspace{-2.5pt}}%
 & %
\mbox{\input{data/pr/PASCALContext_test_new_fb_CEDN_ap.txt}\hspace{-2.5pt}}%
 \\
HED~\cite{XiTu15}   & %
\mbox{\input{data/pr/PASCALContext_test_new_fb_HED_trainval_ods_f.txt}\hspace{-2.5pt}}%
 & %
\mbox{\input{data/pr/PASCALContext_test_new_fb_HED_trainval_ois_f.txt}\hspace{-2.5pt}}%
 & %
\mbox{\input{data/pr/PASCALContext_test_new_fb_HED_trainval_ap.txt}\hspace{-2.5pt}}%
 \\
LEP-B~\cite{Zhao2015}   & %
\mbox{\input{data/pr/PASCALContext_test_new_fb_LEP-BSDS500_ods_f.txt}\hspace{-2.5pt}}%
 & %
\mbox{\input{data/pr/PASCALContext_test_new_fb_LEP-BSDS500_ois_f.txt}\hspace{-2.5pt}}%
 & %
\mbox{\input{data/pr/PASCALContext_test_new_fb_LEP-BSDS500_ap.txt}\hspace{-2.5pt}}%
 \\
HED-B~\cite{XiTu15}   & %
\mbox{\input{data/pr/PASCALContext_test_new_fb_HED-BSDS500_ods_f.txt}\hspace{-2.5pt}}%
 & %
\mbox{\input{data/pr/PASCALContext_test_new_fb_HED-BSDS500_ois_f.txt}\hspace{-2.5pt}}%
 & %
\mbox{\input{data/pr/PASCALContext_test_new_fb_HED-BSDS500_ap.txt}\hspace{-2.5pt}}%
 \\
MCG-B~\cite{Pont-Tuset2016}   & %
\mbox{\input{data/pr/PASCALContext_test_new_fb_MCG-BSDS500_ods_f.txt}\hspace{-2.5pt}}%
 & %
\mbox{\input{data/pr/PASCALContext_test_new_fb_MCG-BSDS500_ois_f.txt}\hspace{-2.5pt}}%
 & %
\mbox{\input{data/pr/PASCALContext_test_new_fb_MCG-BSDS500_ap.txt}\hspace{-2.5pt}}%
 \\
SE-B~\cite{DoZi15}   & %
\mbox{\input{data/pr/PASCALContext_test_new_fb_SE-BSDS500_ods_f.txt}\hspace{-2.5pt}}%
 & %
\mbox{\input{data/pr/PASCALContext_test_new_fb_SE-BSDS500_ois_f.txt}\hspace{-2.5pt}}%
 & %
\mbox{\input{data/pr/PASCALContext_test_new_fb_SE-BSDS500_ap.txt}\hspace{-2.5pt}}%
 \\
\bottomrule
\end{tabular}}
\rule{0mm}{4mm}
\resizebox{\textwidth}{!}{%
\begin{tabular}{lccc}
\toprule
       & \multicolumn{3}{c}{Regions - $F_\mathit{op}$} \\
Method &  ODS & OIS & AP\\
\midrule
COB   & \bf%
\mbox{\input{data/pr/PASCALContext_test_new_fop_COB_trainval_ods_f.txt}\hspace{-2.5pt}}%
 & \bf%
\mbox{\input{data/pr/PASCALContext_test_new_fop_COB_trainval_ois_f.txt}\hspace{-2.5pt}}%
 & \bf%
\mbox{\input{data/pr/PASCALContext_test_new_fop_COB_trainval_ap.txt}\hspace{-2.5pt}}%
 \\
LEP-B~\cite{Zhao2015}   & %
\mbox{\input{data/pr/PASCALContext_test_new_fop_LEP-BSDS500_ods_f.txt}\hspace{-2.5pt}}%
 & %
\mbox{\input{data/pr/PASCALContext_test_new_fop_LEP-BSDS500_ois_f.txt}\hspace{-2.5pt}}%
 & %
\mbox{\input{data/pr/PASCALContext_test_new_fop_LEP-BSDS500_ap.txt}\hspace{-2.5pt}}%
 \\
MCG-B~\cite{Pont-Tuset2016}   & %
\mbox{\input{data/pr/PASCALContext_test_new_fop_MCG-BSDS500_ods_f.txt}\hspace{-2.5pt}}%
 & %
\mbox{\input{data/pr/PASCALContext_test_new_fop_MCG-BSDS500_ois_f.txt}\hspace{-2.5pt}}%
 & %
\mbox{\input{data/pr/PASCALContext_test_new_fop_MCG-BSDS500_ap.txt}\hspace{-2.5pt}}%
 \\
\bottomrule
\end{tabular}}
\end{minipage}
\vspace{-1mm}
\caption{\textbf{PASCAL Context \textit{VOC test} Evaluation}: Precision-recall curves for evaluation of boundaries ($F_b$~\cite{Martin2004}), and regions ($F_{op}$~\cite{Pont-Tuset2016a}). Open contour methods in dashed lines and closed boundaries (from segmentation) in solid lines. ODS, OIS, and AP summary measures. Markers indicate the optimal operating point, where $F_b$ and $F_{op}$ are maximized.}
\label{fig:pr_pascal_test}
\end{figure*}

\begin{figure*}
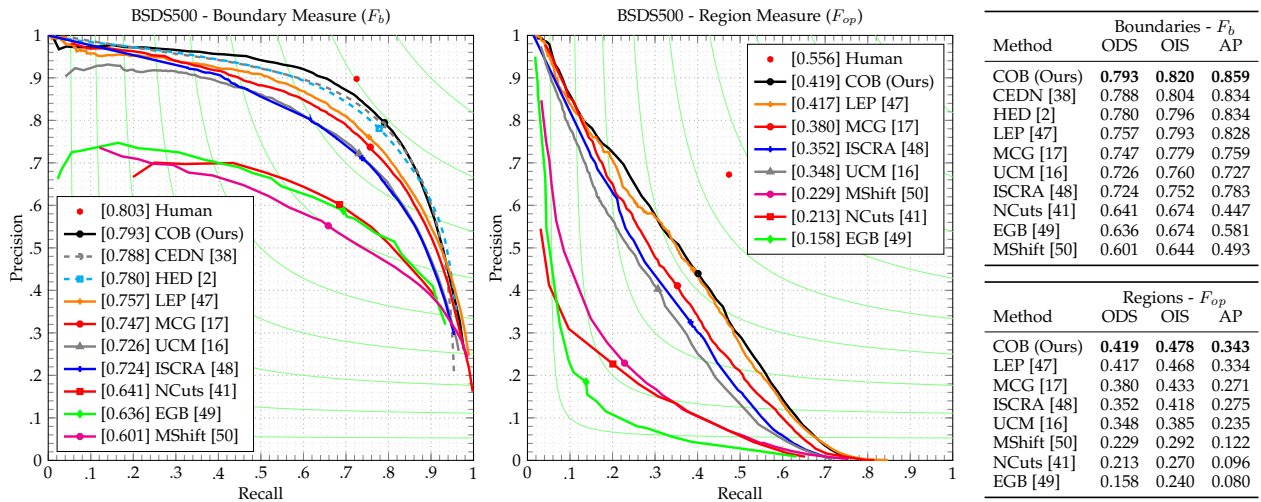

\centering
\begin{minipage}{0.70\linewidth}
\resizebox{0.5\linewidth}{!}{%
\begin{tikzpicture}[/pgfplots/width=0.75\linewidth, /pgfplots/height=0.75\linewidth]
    \begin{axis}[% Axis labels
                 ymin=0,ymax=1,xmin=0,xmax=1,
    			 % Axis labels
        		 xlabel=Recall,
        		 ylabel=Precision,
         		 xlabel shift={-2pt},
        		 ylabel shift={-3pt},
         		 % General appearance
		         font=\small,
		         axis equal image=true,
		         enlargelimits=false,
		         clip=true,
		         % Grids 
        	     grid style=dotted, grid=both,
                 major grid style={white!65!black},
        		 minor grid style={white!85!black},
		 		 xtick={0,0.1,...,1.1},
        		 ytick={0,0.1,...,1.1},
         		 minor xtick={0,0.02,...,1},
		         minor ytick={0,0.02,...,1},
		         xticklabels={0,.1,.2,.3,.4,.5,.6,.7,.8,.9,1},
		         yticklabels={0,.1,.2,.3,.4,.5,.6,.7,.8,.9,1},
        		 % Legend
				 legend cell align=left,
        		 legend style={at={(0.02,0.02)},anchor=south west},
		         % Title
		         title style={yshift=-1ex,},
		         title={BSDS500 - Boundary Measure ($F_b$)}]
        
    % Iso-f curves
    \foreach \f in {0.1,0.2,...,0.9}{%
       \addplot[white!50!green,line width=0.2pt,domain=(\f/(2-\f)):1,samples=200,forget plot]{(\f*x)/(2*x-\f)};
    }

	% Human partitions leave-one-out evaluation
    \addplot+[only marks,red,mark=asterisk,mark size=1.7,line width=\deflinewidth] table[x=Recall,y=Precision] {data/pr/BSDS500_test_fb_human.txt};
    \addlegendentry{[\showodsf{BSDS500_test}{fb}{human}] Human}
	
    % COB
    \addplot+[black,solid,mark=none, line width=\deflinewidth,forget plot] table[x=Recall,y=Precision] {data/pr/BSDS500_test_fb_COB.txt};
    \addplot+[black,solid,mark=o, mark size=1.3, line width=\deflinewidth] table[x=Recall,y=Precision] {data/pr/BSDS500_test_fb_COB_ods.txt};
    \addlegendentry{[\showodsf{BSDS500_test}{fb}{COB}] COB (Ours)}
    
    % CEDN
    \addplot+[gray,dashed,mark=none, line width=\deflinewidth,forget plot] table[x=Recall,y=Precision] {data/pr/BSDS500_test_fb_CEDN.txt};
    \addplot+[gray,dashed,mark=o, mark size=1.3, line width=\deflinewidth] table[x=Recall,y=Precision] {data/pr/BSDS500_test_fb_CEDN_ods.txt};
    \addlegendentry{[\showodsf{BSDS500_test}{fb}{CEDN}] CEDN~\cite{Yan+16}}

	% HED
    \addplot+[cyan,dashed,mark=none, line width=\deflinewidth,forget plot] table[x=Recall,y=Precision] {data/pr/BSDS500_test_fb_HED.txt};
    \addplot+[cyan,dashed,mark=square,mark options={solid}, mark size=1.25, line width=\deflinewidth] table[x=Recall,y=Precision] {data/pr/BSDS500_test_fb_HED_ods.txt};
    \addlegendentry{[\showodsf{BSDS500_test}{fb}{HED}] HED~\cite{XiTu15}}
    
	% LEP
    \addplot+[orange,solid,mark=none, line width=\deflinewidth,forget plot] table[x=Recall,y=Precision] {data/pr/BSDS500_test_fb_LEP.txt};
    \addplot+[orange,solid,mark=+, mark size=1.6, line width=\deflinewidth] table[x=Recall,y=Precision] {data/pr/BSDS500_test_fb_LEP_ods.txt};
    \addlegendentry{[\showodsf{BSDS500_test}{fb}{LEP}] LEP~\cite{Zhao2015}}
    
	% MCG
    \addplot+[red,solid,mark=none, line width=\deflinewidth,forget plot] table[x=Recall,y=Precision] {data/pr/BSDS500_test_fb_MCG.txt};
    \addplot+[red,solid,mark=o, mark size=1.3, line width=\deflinewidth] table[x=Recall,y=Precision] {data/pr/BSDS500_test_fb_MCG_ods.txt};
    \addlegendentry{[\showodsf{BSDS500_test}{fb}{MCG}] MCG~\cite{Pont-Tuset2016}}

	% gPb-UCM
    \addplot+[gray,solid,mark=none, line width=\deflinewidth,forget plot] table[x=Recall,y=Precision] {data/pr/BSDS500_test_fb_gPb-UCM.txt};
    \addplot+[gray,solid,mark=triangle, mark size=1.6, line width=\deflinewidth] table[x=Recall,y=Precision] {data/pr/BSDS500_test_fb_gPb-UCM_ods.txt};
    \addlegendentry{[\showodsf{BSDS500_test}{fb}{gPb-UCM}] UCM~\cite{Arb+11}}
    
    % ISCRA
    \addplot+[blue,solid,mark=none,  line width=\deflinewidth,forget plot] table[x=Recall,y=Precision] {data/pr/BSDS500_test_fb_ISCRA.txt};
    \addplot+[blue,solid,mark=+, mark size=1.6,  line width=\deflinewidth] table[x=Recall,y=Precision] {data/pr/BSDS500_test_fb_ISCRA_ods.txt};
    \addlegendentry{[\showodsf{BSDS500_test}{fb}{ISCRA}] ISCRA~\cite{Ren2013}}
	
	% NCut
    \addplot+[red,solid,mark=none,  line width=\deflinewidth,forget plot] table[x=Recall,y=Precision] {data/pr/BSDS500_test_fb_NCut.txt};
    \addplot+[red,solid,mark=square, mark size=1.25,  line width=\deflinewidth] table[x=Recall,y=Precision] {data/pr/BSDS500_test_fb_NCut_ods.txt};
    \addlegendentry{[\showodsf{BSDS500_test}{fb}{NCut}] NCuts~\cite{Shi2000}}
	
	% EGB
    \addplot+[green,solid,mark=none,  line width=\deflinewidth,forget plot] table[x=Recall,y=Precision] {data/pr/BSDS500_test_fb_EGB.txt};
    \addplot+[green,solid,mark=diamond, mark size=1.5,  line width=\deflinewidth] table[x=Recall,y=Precision] {data/pr/BSDS500_test_fb_EGB_ods.txt};
    \addlegendentry{[\showodsf{BSDS500_test}{fb}{EGB}] EGB~\cite{Felzenszwalb2004}}
	
	% M-Shift
    \addplot+[magenta,solid,mark=none,line width=\deflinewidth,forget plot] table[x=Recall,y=Precision] {data/pr/BSDS500_test_fb_MShift.txt};
    \addplot+[magenta,solid,mark=o, mark size=1.3,  line width=\deflinewidth] table[x=Recall,y=Precision] {data/pr/BSDS500_test_fb_MShift_ods.txt};
    \addlegendentry{[\showodsf{BSDS500_test}{fb}{MShift}] MShift~\cite{Comaniciu2002}}

    \end{axis}
\end{tikzpicture}}
\hspace{-1.5mm}
\resizebox{0.5\linewidth}{!}{%
\begin{tikzpicture}[/pgfplots/width=0.75\linewidth, /pgfplots/height=0.75\linewidth]
    \begin{axis}[% Axis labels
                 ymin=0,ymax=1,xmin=0,xmax=1,
    			 % Axis labels
        		 xlabel=Recall,
        		 ylabel=Precision,
         		 xlabel shift={-2pt},
        		 ylabel shift={-3pt},
         		 % General appearance
		         font=\small,
		         axis equal image=true,
		         enlargelimits=false,
		         clip=true,
		         % Grids 
        	     grid style=dotted, grid=both,
                 major grid style={white!65!black},
        		 minor grid style={white!85!black},
		 		 xtick={0,0.1,...,1.1},
        		 ytick={0,0.1,...,1.1},
         		 minor xtick={0,0.02,...,1},
		         minor ytick={0,0.02,...,1},
		         xticklabels={0,.1,.2,.3,.4,.5,.6,.7,.8,.9,1},
		         yticklabels={0,.1,.2,.3,.4,.5,.6,.7,.8,.9,1},
        		 % Legend
				 legend cell align=left,
        		 legend style={at={(0.98,0.98)},anchor=north east},
		         % Title
		         title style={yshift=-1ex,},
		         title={BSDS500 - Region Measure ($F_{op}$)}]
        
    % Iso-f curves
    \foreach \f in {0.1,0.2,...,0.9}{%
       \addplot[white!50!green,line width=0.2pt,domain=(\f/(2-\f)):1,samples=200,forget plot]{(\f*x)/(2*x-\f)};
    }
	
	% Human partitions leave-one-out evaluation
    \addplot+[only marks,red,mark=asterisk,mark size=1.7,line width=\deflinewidth] table[x=Recall,y=Precision] {data/pr/BSDS500_test_fop_human.txt};
    \addlegendentry{[\showodsf{BSDS500_test}{fop}{human}] Human}
	
    % COB
    \addplot+[black,solid,mark=none, line width=\deflinewidth,forget plot] table[x=Recall,y=Precision] {data/pr/BSDS500_test_fop_COB.txt};
    \addplot+[black,solid,mark=o, mark size=1.3, line width=\deflinewidth] table[x=Recall,y=Precision] {data/pr/BSDS500_test_fop_COB_ods.txt};
    \addlegendentry{[\showodsf{BSDS500_test}{fop}{COB}] COB (Ours)}
    
	% LEP
    \addplot+[orange,solid,mark=none, line width=\deflinewidth,forget plot] table[x=Recall,y=Precision] {data/pr/BSDS500_test_fop_LEP.txt};
    \addplot+[orange,solid,mark=+, mark size=1.6, line width=\deflinewidth] table[x=Recall,y=Precision] {data/pr/BSDS500_test_fop_LEP_ods.txt};
    \addlegendentry{[\showodsf{BSDS500_test}{fop}{LEP}] LEP~\cite{Zhao2015}}

	% MCG
    \addplot+[red,solid,mark=none, line width=\deflinewidth,forget plot] table[x=Recall,y=Precision] {data/pr/BSDS500_test_fop_MCG.txt};
    \addplot+[red,solid,mark=o, mark size=1.3, line width=\deflinewidth] table[x=Recall,y=Precision] {data/pr/BSDS500_test_fop_MCG_ods.txt};
    \addlegendentry{[\showodsf{BSDS500_test}{fop}{MCG}] MCG~\cite{Pont-Tuset2016}}
    
    % ISCRA
    \addplot+[blue,solid,mark=none,  line width=\deflinewidth,forget plot] table[x=Recall,y=Precision] {data/pr/BSDS500_test_fop_ISCRA.txt};
    \addplot+[blue,solid,mark=+, mark size=1.6,  line width=\deflinewidth] table[x=Recall,y=Precision] {data/pr/BSDS500_test_fop_ISCRA_ods.txt};
    \addlegendentry{[\showodsf{BSDS500_test}{fop}{ISCRA}] ISCRA~\cite{Ren2013}}
    
    % gPb-UCM
    \addplot+[gray,solid,mark=none, line width=\deflinewidth,forget plot] table[x=Recall,y=Precision] {data/pr/BSDS500_test_fop_gPb-UCM.txt};
    \addplot+[gray,solid,mark=triangle, mark size=1.6, line width=\deflinewidth] table[x=Recall,y=Precision] {data/pr/BSDS500_test_fop_gPb-UCM_ods.txt};
    \addlegendentry{[\showodsf{BSDS500_test}{fop}{gPb-UCM}] UCM~\cite{Arb+11}}
   
    % M-Shift
    \addplot+[magenta,solid,mark=none,line width=\deflinewidth,forget plot] table[x=Recall,y=Precision] {data/pr/BSDS500_test_fop_MShift.txt};
    \addplot+[magenta,solid,mark=o, mark size=1.3,  line width=\deflinewidth] table[x=Recall,y=Precision] {data/pr/BSDS500_test_fop_MShift_ods.txt};
    \addlegendentry{[\showodsf{BSDS500_test}{fop}{MShift}] MShift~\cite{Comaniciu2002}}
	
	% NCut
    \addplot+[red,solid,mark=none,  line width=\deflinewidth,forget plot] table[x=Recall,y=Precision] {data/pr/BSDS500_test_fop_NCut.txt};
    \addplot+[red,solid,mark=square, mark size=1.25,  line width=\deflinewidth] table[x=Recall,y=Precision] {data/pr/BSDS500_test_fop_NCut_ods.txt};
    \addlegendentry{[\showodsf{BSDS500_test}{fop}{NCut}] NCuts~\cite{Shi2000}}
	
	% EGB
    \addplot+[green,solid,mark=none,  line width=\deflinewidth,forget plot] table[x=Recall,y=Precision] {data/pr/BSDS500_test_fop_EGB.txt};
    \addplot+[green,solid,mark=diamond, mark size=1.5,  line width=\deflinewidth] table[x=Recall,y=Precision] {data/pr/BSDS500_test_fop_EGB_ods.txt};
    \addlegendentry{[\showodsf{BSDS500_test}{fop}{EGB}] EGB~\cite{Felzenszwalb2004}}

    \end{axis}
\end{tikzpicture}}
\end{minipage}
\hspace{1mm}
\begin{minipage}{0.20\linewidth}
\setlength{\tabcolsep}{4pt} % General space between cols (6pt standard)
\center
\footnotesize
%\rowcolors{3}{rowblue}{white}
\resizebox{\textwidth}{!}{%
\begin{tabular}{lccc}
\toprule
       & \multicolumn{3}{c}{Boundaries - $F_b$} \\
Method &  ODS & OIS & AP\\
\midrule
COB (Ours)  & \bf%
\mbox{\input{data/pr/BSDS500_test_fb_COB_ods_f.txt}\hspace{-2.5pt}}%
 & \bf%
\mbox{\input{data/pr/BSDS500_test_fb_COB_ois_f.txt}\hspace{-2.5pt}}%
 & \bf%
\mbox{\input{data/pr/BSDS500_test_fb_COB_ap.txt}\hspace{-2.5pt}}%
 \\
CEDN~\cite{Yan+16}  & %
\mbox{\input{data/pr/BSDS500_test_fb_CEDN_ods_f.txt}\hspace{-2.5pt}}%
 & %
\mbox{\input{data/pr/BSDS500_test_fb_CEDN_ois_f.txt}\hspace{-2.5pt}}%
 & %
\mbox{\input{data/pr/BSDS500_test_fb_CEDN_ap.txt}\hspace{-2.5pt}}%
 \\
HED~\cite{XiTu15}   & %
\mbox{\input{data/pr/BSDS500_test_fb_HED_ods_f.txt}\hspace{-2.5pt}}%
 & %
\mbox{\input{data/pr/BSDS500_test_fb_HED_ois_f.txt}\hspace{-2.5pt}}%
 & %
\mbox{\input{data/pr/BSDS500_test_fb_HED_ap.txt}\hspace{-2.5pt}}%
 \\
LEP~\cite{Zhao2015}   & %
\mbox{\input{data/pr/BSDS500_test_fb_LEP_ods_f.txt}\hspace{-2.5pt}}%
 & %
\mbox{\input{data/pr/BSDS500_test_fb_LEP_ois_f.txt}\hspace{-2.5pt}}%
 & %
\mbox{\input{data/pr/BSDS500_test_fb_LEP_ap.txt}\hspace{-2.5pt}}%
 \\
MCG~\cite{Pont-Tuset2016}   & %
\mbox{\input{data/pr/BSDS500_test_fb_MCG_ods_f.txt}\hspace{-2.5pt}}%
 &  %
\mbox{\input{data/pr/BSDS500_test_fb_MCG_ois_f.txt}\hspace{-2.5pt}}%
 & %
\mbox{\input{data/pr/BSDS500_test_fb_MCG_ap.txt}\hspace{-2.5pt}}%
 \\
UCM~\cite{Arb+11}   & %
\mbox{\input{data/pr/BSDS500_test_fb_gPb-UCM_ods_f.txt}\hspace{-2.5pt}}%
 &  %
\mbox{\input{data/pr/BSDS500_test_fb_gPb-UCM_ois_f.txt}\hspace{-2.5pt}}%
 & %
\mbox{\input{data/pr/BSDS500_test_fb_gPb-UCM_ap.txt}\hspace{-2.5pt}}%
 \\
ISCRA~\cite{Ren2013}   & %
\mbox{\input{data/pr/BSDS500_test_fb_ISCRA_ods_f.txt}\hspace{-2.5pt}}%
 & %
\mbox{\input{data/pr/BSDS500_test_fb_ISCRA_ois_f.txt}\hspace{-2.5pt}}%
 & %
\mbox{\input{data/pr/BSDS500_test_fb_ISCRA_ap.txt}\hspace{-2.5pt}}%
 \\
NCuts~\cite{Shi2000}   & %
\mbox{\input{data/pr/BSDS500_test_fb_NCut_ods_f.txt}\hspace{-2.5pt}}%
 &  %
\mbox{\input{data/pr/BSDS500_test_fb_NCut_ois_f.txt}\hspace{-2.5pt}}%
 & %
\mbox{\input{data/pr/BSDS500_test_fb_NCut_ap.txt}\hspace{-2.5pt}}%
 \\
EGB~\cite{Felzenszwalb2004}   & %
\mbox{\input{data/pr/BSDS500_test_fb_EGB_ods_f.txt}\hspace{-2.5pt}}%
 &  %
\mbox{\input{data/pr/BSDS500_test_fb_EGB_ois_f.txt}\hspace{-2.5pt}}%
 & %
\mbox{\input{data/pr/BSDS500_test_fb_EGB_ap.txt}\hspace{-2.5pt}}%
 \\
MShift~\cite{Comaniciu2002}   & %
\mbox{\input{data/pr/BSDS500_test_fb_MShift_ods_f.txt}\hspace{-2.5pt}}%
 &  %
\mbox{\input{data/pr/BSDS500_test_fb_MShift_ois_f.txt}\hspace{-2.5pt}}%
 & %
\mbox{\input{data/pr/BSDS500_test_fb_MShift_ap.txt}\hspace{-2.5pt}}%
 \\
\bottomrule
\end{tabular}}
\rule{0mm}{2mm}
\resizebox{\textwidth}{!}{%
\begin{tabular}{lccc}
\toprule
       & \multicolumn{3}{c}{Regions - $F_\mathit{op}$} \\
Method &  ODS & OIS & AP\\
\midrule
COB (Ours)  & \bf%
\mbox{\input{data/pr/BSDS500_test_fop_COB_ods_f.txt}\hspace{-2.5pt}}%
 & \bf%
\mbox{\input{data/pr/BSDS500_test_fop_COB_ois_f.txt}\hspace{-2.5pt}}%
 & \bf%
\mbox{\input{data/pr/BSDS500_test_fop_COB_ap.txt}\hspace{-2.5pt}}%
 \\
LEP~\cite{Zhao2015}   & %
\mbox{\input{data/pr/BSDS500_test_fop_LEP_ods_f.txt}\hspace{-2.5pt}}%
 & %
\mbox{\input{data/pr/BSDS500_test_fop_LEP_ois_f.txt}\hspace{-2.5pt}}%
 & %
\mbox{\input{data/pr/BSDS500_test_fop_LEP_ap.txt}\hspace{-2.5pt}}%
 \\
MCG~\cite{Pont-Tuset2016}   & %
\mbox{\input{data/pr/BSDS500_test_fop_MCG_ods_f.txt}\hspace{-2.5pt}}%
 & %
\mbox{\input{data/pr/BSDS500_test_fop_MCG_ois_f.txt}\hspace{-2.5pt}}%
 & %
\mbox{\input{data/pr/BSDS500_test_fop_MCG_ap.txt}\hspace{-2.5pt}}%
 \\
ISCRA~\cite{Ren2013}   & %
\mbox{\input{data/pr/BSDS500_test_fop_ISCRA_ods_f.txt}\hspace{-2.5pt}}%
 &  %
\mbox{\input{data/pr/BSDS500_test_fop_ISCRA_ois_f.txt}\hspace{-2.5pt}}%
 & %
\mbox{\input{data/pr/BSDS500_test_fop_ISCRA_ap.txt}\hspace{-2.5pt}}%
 \\
UCM~\cite{Arb+11}   & %
\mbox{\input{data/pr/BSDS500_test_fop_gPb-UCM_ods_f.txt}\hspace{-2.5pt}}%
 &  %
\mbox{\input{data/pr/BSDS500_test_fop_gPb-UCM_ois_f.txt}\hspace{-2.5pt}}%
 & %
\mbox{\input{data/pr/BSDS500_test_fop_gPb-UCM_ap.txt}\hspace{-2.5pt}}%
 \\
MShift~\cite{Comaniciu2002}   & %
\mbox{\input{data/pr/BSDS500_test_fop_MShift_ods_f.txt}\hspace{-2.5pt}}%
 &  %
\mbox{\input{data/pr/BSDS500_test_fop_MShift_ois_f.txt}\hspace{-2.5pt}}%
 & %
\mbox{\input{data/pr/BSDS500_test_fop_MShift_ap.txt}\hspace{-2.5pt}}%
 \\
NCuts~\cite{Shi2000}   & %
\mbox{\input{data/pr/BSDS500_test_fop_NCut_ods_f.txt}\hspace{-2.5pt}}%
 &  %
\mbox{\input{data/pr/BSDS500_test_fop_NCut_ois_f.txt}\hspace{-2.5pt}}%
 & %
\mbox{\input{data/pr/BSDS500_test_fop_NCut_ap.txt}\hspace{-2.5pt}}%
 \\
EGB~\cite{Felzenszwalb2004}   & %
\mbox{\input{data/pr/BSDS500_test_fop_EGB_ods_f.txt}\hspace{-2.5pt}}%
 &  %
\mbox{\input{data/pr/BSDS500_test_fop_EGB_ois_f.txt}\hspace{-2.5pt}}%
 & %
\mbox{\input{data/pr/BSDS500_test_fop_EGB_ap.txt}\hspace{-2.5pt}}%
 \\
\bottomrule
\end{tabular}}
\end{minipage}
\vspace{-1mm}
\caption{\textbf{BSDS500 Test Evaluation}: Precision-recall curves for evaluation of boundaries ($F_b$~\cite{Martin2004}), and regions ($F_{op}$~\cite{Pont-Tuset2016a}).}
\label{fig:pr_bsds_test}
\vspace{-2mm}
\end{figure*}

\subsection{Generic Image Segmentation}
\label{sec:exp:generic}
We present our results for contour detection and generic image segmentation on PASCAL Context~\cite{Mot+14} as well as on the BSDS500~\cite{Martin2001}, which is the most established benchmark for perceptual grouping.

\paragraph*{\textbf{PASCAL Context}}
We train COB in the \emph{VOC train}, and perform hyper-parameter selection on \emph{VOC val}. We report the final results on the unseen \emph{VOC test} when trained on \emph{VOC trainval}, using the previously tuned hyper-parameters. We compare our approach to several methods trained on the BSDS~\cite{DoZi15,Pont-Tuset2016,Zhao2015,XiTu15} and we also retrain the current state-of-the-art contour detection methods HED~\cite{XiTu15} and the recent CEDN~\cite{Yan+16} on \emph{VOC trainval} using the code provided by the respective authors.

Figure~\ref{fig:pr_pascal_test} presents the evaluation results of COB compared to the state of the art, showing that it outperforms all others by a considerable margin both in terms of boundaries and in terms of regions. The lower performance of the methods trained on the BSDS quantifies the difficulty of the task when moving to a larger and more challenging dataset.

\paragraph*{\textbf{BSDS500}}
We retrain COB using only the 300 \textit{trainval} images of the BSDS, after data augmentation as suggested in~\cite{XiTu15}, keeping the architecture decided in Section~\ref{sec:exp:ablat}. For comparison to HED~\cite{XiTu15}, we used the model that the authors provide online. We also compare with CEDN~\cite{Yan+16}, by evaluating the results provided by the authors. 

Figure~\ref{fig:pr_bsds_test} presents the evaluation results, which show that we also obtain state-of-the-art results in this dataset. The smaller margins are in all likelihood due to the fact that we almost reach human performance for the task of contour detection on the BSDS, which motivates the shift to PASCAL Context to achieve further progress in the field.

\begin{figure}
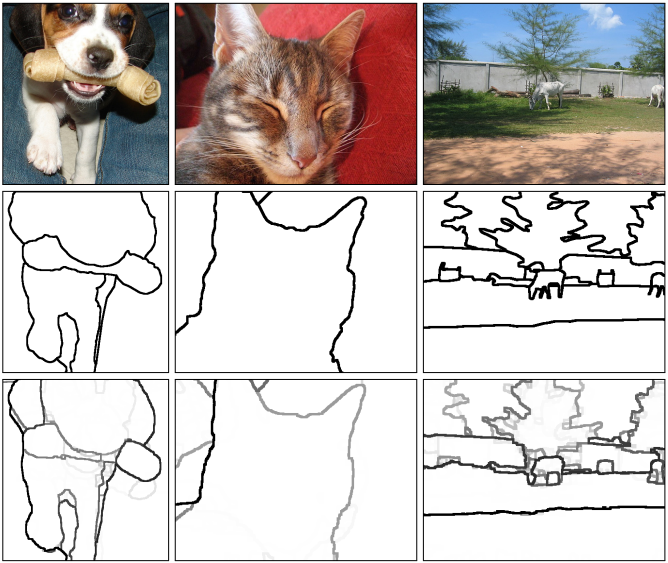

\centering
\resizebox{\linewidth}{!}{\setlength{\fboxsep}{0pt}\noindent%
\showinonecol{2008_000078}{2008_000056}{2008_000009}{img}{gen_cont}{0.33}}\\[0.8mm]
\resizebox{\linewidth}{!}{\setlength{\fboxsep}{0pt}\noindent%
\showinonecol{2008_000078}{2008_000056}{2008_000009}{gt}{gen_cont}{0.33}}\\[0.8mm]
\resizebox{\linewidth}{!}{\setlength{\fboxsep}{0pt}\noindent%
\showinonecol{2008_000078}{2008_000056}{2008_000009}{seg}{gen_cont}{0.33}}\\
\caption{\textbf{Qualitative results on PASCAL - Hierarchical Regions}. Row 1: original images, Row 2: ground-truth boundaries, Row 3: hierarchical regions with COB.}
\label{fig:qual_cont1}
\end{figure}

\paragraph*{\textbf{Qualitative Results}}
Figure~\ref{fig:qual_cont1} shows some qualitative results of our hierarchical contours.
Please note that COB is capable of correctly distinguishing between internal contours and external, semantically meaningful boundaries.

\subsection{Object boundary detection}
\label{sec:exp:obj_boundary}
Concurrent works with the conference version of our paper~\cite{Maninis2016a} presented results on object boundary detection~\cite{Yan+16, Kho+16} on the PASCAL VOC'12 Segmentation database. The database consists of 1464 training and 1449 validation images, including pixel-wise annotations of the instances and the semantic classes of the objects. The goal is to detect the boundaries of the objects that belong to the 20 classes of PASCAL, without distinguishing the semantics. Different from generic image segmentation, boundaries that do not belong to an object are treated as background.

\begin{figure}[b]
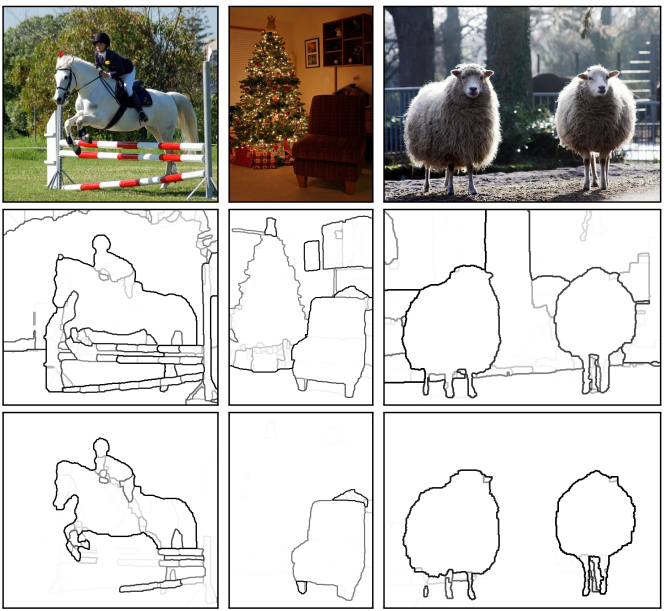

\centering
\resizebox{\linewidth}{!}{\setlength{\fboxsep}{0pt}\noindent%
\showinonecol{2007_000783}{2007_006553}{2007_000925}{img}{obj_cont}{0.21}}\\[0.8mm]
\resizebox{\linewidth}{!}{\setlength{\fboxsep}{0pt}\noindent%
\showinonecol{2007_000783}{2007_006553}{2007_000925}{context}{obj_cont}{0.21}}\\[0.8mm]
\resizebox{\linewidth}{!}{\setlength{\fboxsep}{0pt}\noindent%
\showinonecol{2007_000783}{2007_006553}{2007_000925}{obj}{obj_cont}{0.21}}\\
\caption{\textbf{Qualitative results for Object Boundaries}. Row~1: original images, Row 2: Generic Image Segmentation results, Row 3: Object Boundary results.}
\label{fig:qual_obj_cont}
\end{figure}

\begin{figure*}
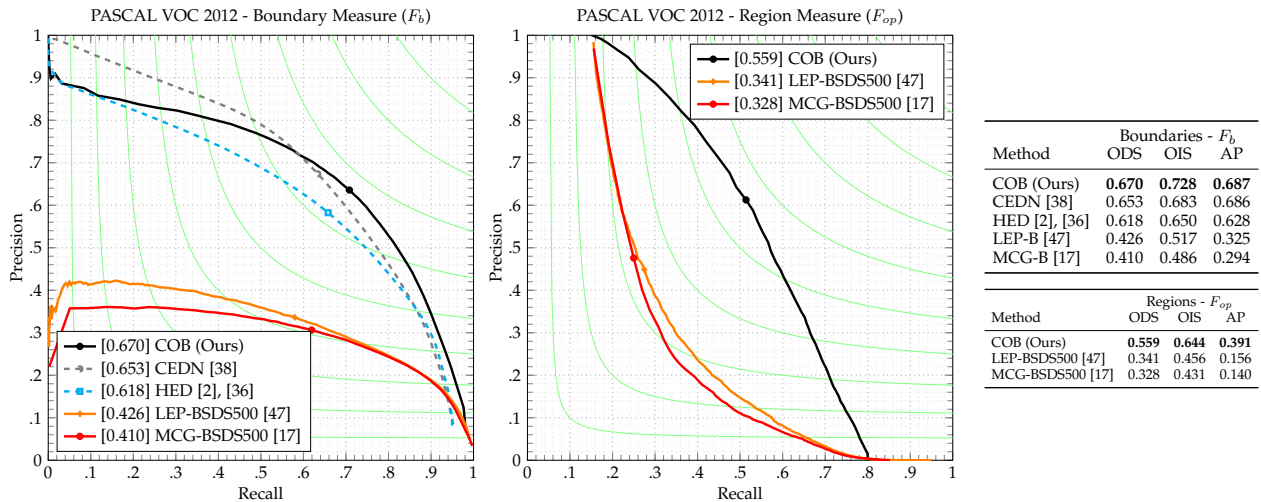

\centering
\begin{minipage}{0.70\linewidth}
\resizebox{0.5\linewidth}{!}{%
\begin{tikzpicture}[/pgfplots/width=0.75\linewidth, /pgfplots/height=0.75\linewidth]
    \begin{axis}[% Axis labels
                 ymin=0,ymax=1,xmin=0,xmax=1,
    			 % Axis labels
        		 xlabel=Recall,
        		 ylabel=Precision,
         		 xlabel shift={-2pt},
        		 ylabel shift={-3pt},
         		 % General appearance
		         font=\small,
		         axis equal image=true,
		         enlargelimits=false,
		         clip=true,
		         % Grids 
        	     grid style=dotted, grid=both,
                 major grid style={white!65!black},
        		 minor grid style={white!85!black},
		 		 xtick={0,0.1,...,1.1},
        		 ytick={0,0.1,...,1.1},
         		 minor xtick={0,0.02,...,1},
		         minor ytick={0,0.02,...,1},
		         xticklabels={0,.1,.2,.3,.4,.5,.6,.7,.8,.9,1},
		         yticklabels={0,.1,.2,.3,.4,.5,.6,.7,.8,.9,1},
        		 % Legend
				 legend cell align=left,
        		 legend style={at={(0.02,0.02)},anchor=south west},
		         % Title
		         title style={yshift=-1ex,},
		         title={PASCAL VOC 2012 - Boundary Measure ($F_b$)}]
        
    % Iso-f curves
    \foreach \f in {0.1,0.2,...,0.9}{%
       \addplot[white!50!green,line width=0.2pt,domain=(\f/(2-\f)):1,samples=200,forget plot]{(\f*x)/(2*x-\f)};
    }
	
    % COB
    \addplot+[black,solid,mark=none, line width=\deflinewidth,forget plot] table[x=Recall,y=Precision] {data/pr/Pascal_Segmentation_val_2012_fb_COB.txt};
    \addplot+[black,solid,mark=o, mark size=1.3, line width=\deflinewidth] table[x=Recall,y=Precision] {data/pr/Pascal_Segmentation_val_2012_fb_COB_ods.txt};
    \addlegendentry{[\showodsf{Pascal_Segmentation_val_2012}{fb}{COB}] COB (Ours)}
    
    % CEDN
    \addplot+[gray,dashed,mark=none, line width=\deflinewidth,forget plot] table[x=Recall,y=Precision] {data/pr/Pascal_Segmentation_val_2012_fb_CEDN.txt};
    \addplot+[gray,dashed,mark=o, mark size=1.3, line width=\deflinewidth] table[x=Recall,y=Precision] {data/pr/Pascal_Segmentation_val_2012_fb_CEDN_ods.txt};
    \addlegendentry{[\showodsf{Pascal_Segmentation_val_2012}{fb}{CEDN}] CEDN~\cite{Yan+16}}

	% HED
    \addplot+[cyan,dashed,mark=none, line width=\deflinewidth,forget plot] table[x=Recall,y=Precision] {data/pr/Pascal_Segmentation_val_2012_fb_HED.txt};
    \addplot+[cyan,dashed,mark=square,mark options={solid}, mark size=1.25, line width=\deflinewidth] table[x=Recall,y=Precision] {data/pr/Pascal_Segmentation_val_2012_fb_HED_ods.txt};
    \addlegendentry{[\showodsf{Pascal_Segmentation_val_2012}{fb}{HED}] HED~\cite{XiTu15,Kho+16}}
    
	% LEP-BSDS500
    \addplot+[orange,solid,mark=none, line width=\deflinewidth,forget plot] table[x=Recall,y=Precision] {data/pr/Pascal_Segmentation_val_2012_fb_LEP-BSDS500.txt};
    \addplot+[orange,solid,mark=+, mark size=1.6, line width=\deflinewidth] table[x=Recall,y=Precision] {data/pr/Pascal_Segmentation_val_2012_fb_LEP-BSDS500_ods.txt};
    \addlegendentry{[\showodsf{Pascal_Segmentation_val_2012}{fb}{LEP-BSDS500}] LEP-BSDS500~\cite{Zhao2015}}
    
	% MCG-BSDS500
    \addplot+[red,solid,mark=none, line width=\deflinewidth,forget plot] table[x=Recall,y=Precision] {data/pr/Pascal_Segmentation_val_2012_fb_MCG-BSDS500.txt};
    \addplot+[red,solid,mark=o, mark size=1.3, line width=\deflinewidth] table[x=Recall,y=Precision] {data/pr/Pascal_Segmentation_val_2012_fb_MCG-BSDS500_ods.txt};
    \addlegendentry{[\showodsf{Pascal_Segmentation_val_2012}{fb}{MCG-BSDS500}] MCG-BSDS500~\cite{Pont-Tuset2016}}

    \end{axis}
\end{tikzpicture}}
\hspace{-1.5mm}
\resizebox{0.5\linewidth}{!}{%
\begin{tikzpicture}[/pgfplots/width=0.75\linewidth, /pgfplots/height=0.75\linewidth]
    \begin{axis}[% Axis labels
                 ymin=0,ymax=1,xmin=0,xmax=1,
    			 % Axis labels
        		 xlabel=Recall,
        		 ylabel=Precision,
         		 xlabel shift={-2pt},
        		 ylabel shift={-3pt},
         		 % General appearance
		         font=\small,
		         axis equal image=true,
		         enlargelimits=false,
		         clip=true,
		         % Grids 
        	     grid style=dotted, grid=both,
                 major grid style={white!65!black},
        		 minor grid style={white!85!black},
		 		 xtick={0,0.1,...,1.1},
        		 ytick={0,0.1,...,1.1},
         		 minor xtick={0,0.02,...,1},
		         minor ytick={0,0.02,...,1},
		         xticklabels={0,.1,.2,.3,.4,.5,.6,.7,.8,.9,1},
		         yticklabels={0,.1,.2,.3,.4,.5,.6,.7,.8,.9,1},
        		 % Legend
				 legend cell align=left,
        		 legend style={at={(0.98,0.98)},anchor=north east},
		         % Title
		         title style={yshift=-1ex,},
		         title={PASCAL VOC 2012 - Region Measure ($F_{op}$)}]
        
    % Iso-f curves
    \foreach \f in {0.1,0.2,...,0.9}{%
       \addplot[white!50!green,line width=0.2pt,domain=(\f/(2-\f)):1,samples=200,forget plot]{(\f*x)/(2*x-\f)};
    }

    % COB
    \addplot+[black,solid,mark=none, line width=\deflinewidth,forget plot] table[x=Recall,y=Precision] {data/pr/Pascal_Segmentation_val_2012_fop_COB.txt};
    \addplot+[black,solid,mark=o, mark size=1.3, line width=\deflinewidth] table[x=Recall,y=Precision] {data/pr/Pascal_Segmentation_val_2012_fop_COB_ods.txt};
    \addlegendentry{[\showodsf{Pascal_Segmentation_val_2012}{fop}{COB}] COB (Ours)}
    
	% LEP
    \addplot+[orange,solid,mark=none, line width=\deflinewidth,forget plot] table[x=Recall,y=Precision] {data/pr/Pascal_Segmentation_val_2012_fop_LEP-BSDS500.txt};
    \addplot+[orange,solid,mark=+, mark size=1.6, line width=\deflinewidth] table[x=Recall,y=Precision] {data/pr/Pascal_Segmentation_val_2012_fop_LEP-BSDS500_ods.txt};
    \addlegendentry{[\showodsf{Pascal_Segmentation_val_2012}{fop}{LEP-BSDS500}] LEP-BSDS500~\cite{Zhao2015}}

	% MCG
    \addplot+[red,solid,mark=none, line width=\deflinewidth,forget plot] table[x=Recall,y=Precision] {data/pr/Pascal_Segmentation_val_2012_fop_MCG-BSDS500.txt};
    \addplot+[red,solid,mark=o, mark size=1.3, line width=\deflinewidth] table[x=Recall,y=Precision] {data/pr/Pascal_Segmentation_val_2012_fop_MCG-BSDS500_ods.txt};
    \addlegendentry{[\showodsf{Pascal_Segmentation_val_2012}{fop}{MCG-BSDS500}] MCG-BSDS500~\cite{Pont-Tuset2016}}

    \end{axis}
\end{tikzpicture}}
\end{minipage}
\hspace{1mm}
\begin{minipage}{0.20\linewidth}
\setlength{\tabcolsep}{4pt} % General space between cols (6pt standard)
\center
\footnotesize
%\rowcolors{3}{rowblue}{white}
\resizebox{\textwidth}{!}{%
\begin{tabular}{lccc}
\toprule
       & \multicolumn{3}{c}{Boundaries - $F_b$} \\
Method &  ODS & OIS & AP\\
\midrule
COB (Ours)  & \bf%
\mbox{\input{data/pr/Pascal_Segmentation_val_2012_fb_COB_ods_f.txt}\hspace{-2.5pt}}%
 & \bf%
\mbox{\input{data/pr/Pascal_Segmentation_val_2012_fb_COB_ois_f.txt}\hspace{-2.5pt}}%
 & \bf%
\mbox{\input{data/pr/Pascal_Segmentation_val_2012_fb_COB_ap.txt}\hspace{-2.5pt}}%
 \\
CEDN~\cite{Yan+16}  & %
\mbox{\input{data/pr/Pascal_Segmentation_val_2012_fb_CEDN_ods_f.txt}\hspace{-2.5pt}}%
 & %
\mbox{\input{data/pr/Pascal_Segmentation_val_2012_fb_CEDN_ois_f.txt}\hspace{-2.5pt}}%
 & %
\mbox{\input{data/pr/Pascal_Segmentation_val_2012_fb_CEDN_ap.txt}\hspace{-2.5pt}}%
 \\
HED~\cite{XiTu15,Kho+16}   & %
\mbox{\input{data/pr/Pascal_Segmentation_val_2012_fb_HED_ods_f.txt}\hspace{-2.5pt}}%
 & %
\mbox{\input{data/pr/Pascal_Segmentation_val_2012_fb_HED_ois_f.txt}\hspace{-2.5pt}}%
 & %
\mbox{\input{data/pr/Pascal_Segmentation_val_2012_fb_HED_ap.txt}\hspace{-2.5pt}}%
 \\
LEP-B~\cite{Zhao2015}   & %
\mbox{\input{data/pr/Pascal_Segmentation_val_2012_fb_LEP-BSDS500_ods_f.txt}\hspace{-2.5pt}}%
 & %
\mbox{\input{data/pr/Pascal_Segmentation_val_2012_fb_LEP-BSDS500_ois_f.txt}\hspace{-2.5pt}}%
 & %
\mbox{\input{data/pr/Pascal_Segmentation_val_2012_fb_LEP-BSDS500_ap.txt}\hspace{-2.5pt}}%
 \\
MCG-B~\cite{Pont-Tuset2016}   & %
\mbox{\input{data/pr/Pascal_Segmentation_val_2012_fb_MCG-BSDS500_ods_f.txt}\hspace{-2.5pt}}%
 &  %
\mbox{\input{data/pr/Pascal_Segmentation_val_2012_fb_MCG-BSDS500_ois_f.txt}\hspace{-2.5pt}}%
 & %
\mbox{\input{data/pr/Pascal_Segmentation_val_2012_fb_MCG-BSDS500_ap.txt}\hspace{-2.5pt}}%
 \\
\bottomrule
\end{tabular}}
\rule{0mm}{2mm}
\resizebox{\textwidth}{!}{%
\begin{tabular}{lccc}
\toprule
       & \multicolumn{3}{c}{Regions - $F_\mathit{op}$} \\
Method &  ODS & OIS & AP\\
\midrule
COB (Ours)  & \bf%
\mbox{\input{data/pr/Pascal_Segmentation_val_2012_fop_COB_ods_f.txt}\hspace{-2.5pt}}%
 & \bf%
\mbox{\input{data/pr/Pascal_Segmentation_val_2012_fop_COB_ois_f.txt}\hspace{-2.5pt}}%
 & \bf%
\mbox{\input{data/pr/Pascal_Segmentation_val_2012_fop_COB_ap.txt}\hspace{-2.5pt}}%
 \\
LEP-BSDS500~\cite{Zhao2015}   & %
\mbox{\input{data/pr/Pascal_Segmentation_val_2012_fop_LEP-BSDS500_ods_f.txt}\hspace{-2.5pt}}%
 & %
\mbox{\input{data/pr/Pascal_Segmentation_val_2012_fop_LEP-BSDS500_ois_f.txt}\hspace{-2.5pt}}%
 & %
\mbox{\input{data/pr/Pascal_Segmentation_val_2012_fop_LEP-BSDS500_ap.txt}\hspace{-2.5pt}}%
 \\
MCG-BSDS500~\cite{Pont-Tuset2016}   & %
\mbox{\input{data/pr/Pascal_Segmentation_val_2012_fop_MCG-BSDS500_ods_f.txt}\hspace{-2.5pt}}%
 & %
\mbox{\input{data/pr/Pascal_Segmentation_val_2012_fop_MCG-BSDS500_ois_f.txt}\hspace{-2.5pt}}%
 & %
\mbox{\input{data/pr/Pascal_Segmentation_val_2012_fop_MCG-BSDS500_ap.txt}\hspace{-2.5pt}}%
 \\
\bottomrule
\end{tabular}}
\end{minipage}
\vspace{-1mm}
\caption{\textbf{PASCAL VOC 2012}: Precision-recall curves for boundaries ($F_b$~\cite{Martin2004}), and regions ($F_{op}$~\cite{Pont-Tuset2016a}). ODS, OIS, and AP summary measures.}
\label{fig:pr_pascal_obj}
\end{figure*}

\begin{figure*}
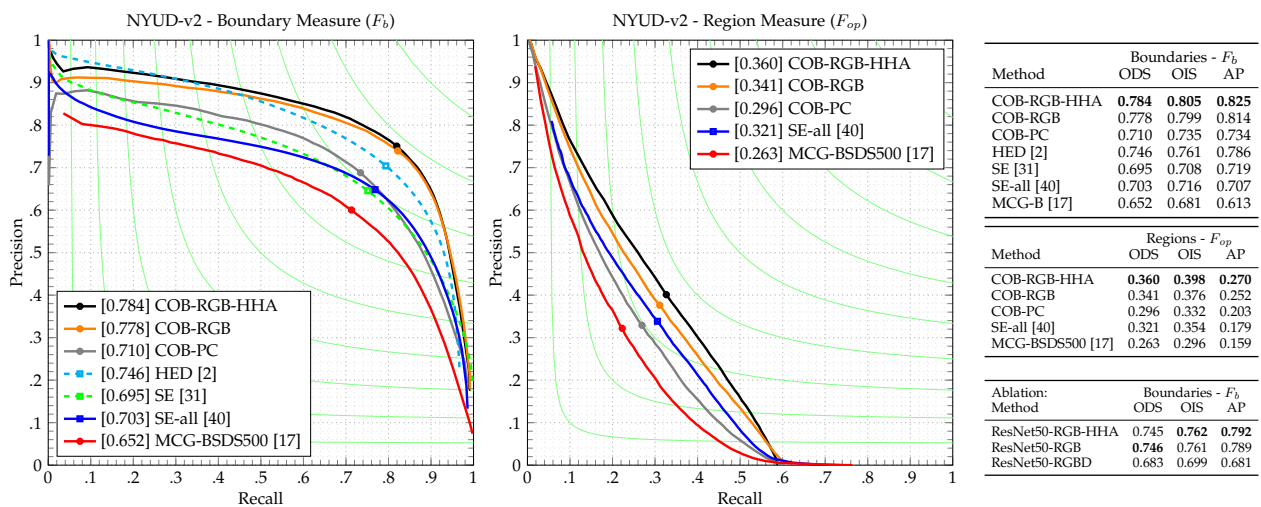

\centering
\begin{minipage}{0.70\linewidth}
\resizebox{0.5\linewidth}{!}{%
\begin{tikzpicture}[/pgfplots/width=0.75\linewidth, /pgfplots/height=0.75\linewidth]
    \begin{axis}[% Axis labels
                 ymin=0,ymax=1,xmin=0,xmax=1,
    			 % Axis labels
        		 xlabel=Recall,
        		 ylabel=Precision,
         		 xlabel shift={-2pt},
        		 ylabel shift={-3pt},
         		 % General appearance
		         font=\small,
		         axis equal image=true,
		         enlargelimits=false,
		         clip=true,
		         % Grids 
        	     grid style=dotted, grid=both,
                 major grid style={white!65!black},
        		 minor grid style={white!85!black},
		 		 xtick={0,0.1,...,1.1},
        		 ytick={0,0.1,...,1.1},
         		 minor xtick={0,0.02,...,1},
		         minor ytick={0,0.02,...,1},
		         xticklabels={0,.1,.2,.3,.4,.5,.6,.7,.8,.9,1},
		         yticklabels={0,.1,.2,.3,.4,.5,.6,.7,.8,.9,1},
        		 % Legend
				 legend cell align=left,
        		 legend style={at={(0.02,0.02)},anchor=south west},
		         % Title
		         title style={yshift=-1ex,},
		         title={NYUD-v2 - Boundary Measure ($F_b$)}]
        
    % Iso-f curves
    \foreach \f in {0.1,0.2,...,0.9}{%
       \addplot[white!50!green,line width=0.2pt,domain=(\f/(2-\f)):1,samples=200,forget plot]{(\f*x)/(2*x-\f)};
    }
	
    % COB-RGB-HHA
    \addplot+[black,solid,mark=none, line width=\deflinewidth,forget plot] table[x=Recall,y=Precision] {data/pr/NYUD-v2_test_fb_COB-rgbHHA.txt};
    \addplot+[black,solid,mark=o, mark size=1.3, line width=\deflinewidth] table[x=Recall,y=Precision] {data/pr/NYUD-v2_test_fb_COB-rgbHHA_ods.txt};
    \addlegendentry{[\showodsf{NYUD-v2_test}{fb}{COB-rgbHHA}] COB-RGB-HHA}
    
    % COB
    \addplot+[orange,solid,mark=none, line width=\deflinewidth,forget plot] table[x=Recall,y=Precision] {data/pr/NYUD-v2_test_fb_COB.txt};
    \addplot+[orange,solid,mark=o, mark size=1.3, line width=\deflinewidth] table[x=Recall,y=Precision] {data/pr/NYUD-v2_test_fb_COB_ods.txt};
    \addlegendentry{[\showodsf{NYUD-v2_test}{fb}{COB}] COB-RGB}
    
    % COB-PC
    \addplot+[gray,solid,mark=none, line width=\deflinewidth,forget plot] table[x=Recall,y=Precision] {data/pr/NYUD-v2_test_fb_COB-PC.txt};
    \addplot+[gray,solid,mark=o, mark size=1.3, line width=\deflinewidth] table[x=Recall,y=Precision] {data/pr/NYUD-v2_test_fb_COB-PC_ods.txt};
    \addlegendentry{[\showodsf{NYUD-v2_test}{fb}{COB-PC}] COB-PC}

	% HED
    \addplot+[cyan,dashed,mark=none, line width=\deflinewidth,forget plot] table[x=Recall,y=Precision] {data/pr/NYUD-v2_test_fb_HED.txt};
    \addplot+[cyan,dashed,mark=square,mark options={solid}, mark size=1.25, line width=\deflinewidth] table[x=Recall,y=Precision] {data/pr/NYUD-v2_test_fb_HED_ods.txt};
    \addlegendentry{[\showodsf{NYUD-v2_test}{fb}{HED}] HED~\cite{XiTu15}}
    
	% SE
    \addplot+[green,dashed,mark=none, line width=\deflinewidth,forget plot] table[x=Recall,y=Precision] {data/pr/NYUD-v2_test_fb_SE.txt};
    \addplot+[green,dashed,mark=square,mark options={solid}, mark size=1.25, line width=\deflinewidth] table[x=Recall,y=Precision] {data/pr/NYUD-v2_test_fb_SE_ods.txt};
    \addlegendentry{[\showodsf{NYUD-v2_test}{fb}{SE}] SE~\cite{DoZi15}}
    
    % SE-all
    \addplot+[blue,solid,mark=none, line width=\deflinewidth,forget plot] table[x=Recall,y=Precision] {data/pr/NYUD-v2_test_fb_SE-all.txt};
    \addplot+[blue,solid,mark=square,mark options={solid}, mark size=1.25, line width=\deflinewidth] table[x=Recall,y=Precision] {data/pr/NYUD-v2_test_fb_SE-all_ods.txt};
    \addlegendentry{[\showodsf{NYUD-v2_test}{fb}{SE-all}] SE-all~\cite{Gup+14}}
    
	% MCG-BSDS500
    \addplot+[red,solid,mark=none, line width=\deflinewidth,forget plot] table[x=Recall,y=Precision] {data/pr/NYUD-v2_test_fb_MCG.txt};
    \addplot+[red,solid,mark=o, mark size=1.3, line width=\deflinewidth] table[x=Recall,y=Precision] {data/pr/NYUD-v2_test_fb_MCG_ods.txt};
    \addlegendentry{[\showodsf{NYUD-v2_test}{fb}{MCG}] MCG-BSDS500~\cite{Pont-Tuset2016}}

    \end{axis}
\end{tikzpicture}}
\hspace{-1.5mm}
\resizebox{0.5\linewidth}{!}{%
\begin{tikzpicture}[/pgfplots/width=0.75\linewidth, /pgfplots/height=0.75\linewidth]
    \begin{axis}[% Axis labels
                 ymin=0,ymax=1,xmin=0,xmax=1,
    			 % Axis labels
        		 xlabel=Recall,
        		 ylabel=Precision,
         		 xlabel shift={-2pt},
        		 ylabel shift={-3pt},
         		 % General appearance
		         font=\small,
		         axis equal image=true,
		         enlargelimits=false,
		         clip=true,
		         % Grids 
        	     grid style=dotted, grid=both,
                 major grid style={white!65!black},
        		 minor grid style={white!85!black},
		 		 xtick={0,0.1,...,1.1},
        		 ytick={0,0.1,...,1.1},
         		 minor xtick={0,0.02,...,1},
		         minor ytick={0,0.02,...,1},
		         xticklabels={0,.1,.2,.3,.4,.5,.6,.7,.8,.9,1},
		         yticklabels={0,.1,.2,.3,.4,.5,.6,.7,.8,.9,1},
        		 % Legend
				 legend cell align=left,
        		 legend style={at={(0.98,0.98)},anchor=north east},
		         % Title
		         title style={yshift=-1ex,},
		         title={NYUD-v2 - Region Measure ($F_{op}$)}]
        
    % Iso-f curves
    \foreach \f in {0.1,0.2,...,0.9}{%
       \addplot[white!50!green,line width=0.2pt,domain=(\f/(2-\f)):1,samples=200,forget plot]{(\f*x)/(2*x-\f)};
    }
	
    % COB-RGB-HHA
    \addplot+[black,solid,mark=none, line width=\deflinewidth,forget plot] table[x=Recall,y=Precision] {data/pr/NYUD-v2_test_fop_COB-rgbHHA.txt};
    \addplot+[black,solid,mark=o, mark size=1.3, line width=\deflinewidth] table[x=Recall,y=Precision] {data/pr/NYUD-v2_test_fop_COB-rgbHHA_ods.txt};
    \addlegendentry{[\showodsf{NYUD-v2_test}{fop}{COB-rgbHHA}] COB-RGB-HHA}
    
    % COB
    \addplot+[orange,solid,mark=none, line width=\deflinewidth,forget plot] table[x=Recall,y=Precision] {data/pr/NYUD-v2_test_fop_COB.txt};
    \addplot+[orange,solid,mark=o, mark size=1.3, line width=\deflinewidth] table[x=Recall,y=Precision] {data/pr/NYUD-v2_test_fop_COB_ods.txt};
    \addlegendentry{[\showodsf{NYUD-v2_test}{fop}{COB}] COB-RGB}
    
    % COB-PC
    \addplot+[gray,solid,mark=none, line width=\deflinewidth,forget plot] table[x=Recall,y=Precision] {data/pr/NYUD-v2_test_fop_COB-PC.txt};
    \addplot+[gray,solid,mark=o, mark size=1.3, line width=\deflinewidth] table[x=Recall,y=Precision] {data/pr/NYUD-v2_test_fop_COB-PC_ods.txt};
    \addlegendentry{[\showodsf{NYUD-v2_test}{fop}{COB-PC}] COB-PC}

	% SE-all
    \addplot+[blue,solid,mark=none, line width=\deflinewidth,forget plot] table[x=Recall,y=Precision] {data/pr/NYUD-v2_test_fop_HHA.txt};
    \addplot+[blue,solid,mark=square,mark options={solid}, mark size=1.25, line width=\deflinewidth] table[x=Recall,y=Precision] {data/pr/NYUD-v2_test_fop_HHA_ods.txt};
    \addlegendentry{[\showodsf{NYUD-v2_test}{fop}{HHA}] SE-all~\cite{Gup+14}}
    
	% MCG-BSDS500
    \addplot+[red,solid,mark=none, line width=\deflinewidth,forget plot] table[x=Recall,y=Precision] {data/pr/NYUD-v2_test_fop_MCG.txt};
    \addplot+[red,solid,mark=o, mark size=1.3, line width=\deflinewidth] table[x=Recall,y=Precision] {data/pr/NYUD-v2_test_fop_MCG_ods.txt};
    \addlegendentry{[\showodsf{NYUD-v2_test}{fop}{MCG}] MCG-BSDS500~\cite{Pont-Tuset2016}}

    \end{axis}
\end{tikzpicture}}
\end{minipage}
\hspace{1mm}
\begin{minipage}{0.20\linewidth}
\setlength{\tabcolsep}{4pt} % General space between cols (6pt standard)
\center
\footnotesize
%\rowcolors{3}{rowblue}{white}
\resizebox{\textwidth}{!}{%
\begin{tabular}{lccc}
\toprule
       & \multicolumn{3}{c}{Boundaries - $F_b$} \\
Method &  ODS & OIS & AP\\
\midrule
COB-RGB-HHA  & \bf%
\mbox{\input{data/pr/NYUD-v2_test_fb_COB-rgbHHA_ods_f.txt}\hspace{-2.5pt}}%
 & \bf%
\mbox{\input{data/pr/NYUD-v2_test_fb_COB-rgbHHA_ois_f.txt}\hspace{-2.5pt}}%
 & \bf%
\mbox{\input{data/pr/NYUD-v2_test_fb_COB-rgbHHA_ap.txt}\hspace{-2.5pt}}%
 \\
COB-RGB  & %
\mbox{\input{data/pr/NYUD-v2_test_fb_COB_ods_f.txt}\hspace{-2.5pt}}%
 & %
\mbox{\input{data/pr/NYUD-v2_test_fb_COB_ois_f.txt}\hspace{-2.5pt}}%
 & %
\mbox{\input{data/pr/NYUD-v2_test_fb_COB_ap.txt}\hspace{-2.5pt}}%
 \\
COB-PC  & %
\mbox{\input{data/pr/NYUD-v2_test_fb_COB-PC_ods_f.txt}\hspace{-2.5pt}}%
 & %
\mbox{\input{data/pr/NYUD-v2_test_fb_COB-PC_ois_f.txt}\hspace{-2.5pt}}%
 & %
\mbox{\input{data/pr/NYUD-v2_test_fb_COB-PC_ap.txt}\hspace{-2.5pt}}%
 \\
HED~\cite{XiTu15}   & %
\mbox{\input{data/pr/NYUD-v2_test_fb_HED_ods_f.txt}\hspace{-2.5pt}}%
 & %
\mbox{\input{data/pr/NYUD-v2_test_fb_HED_ois_f.txt}\hspace{-2.5pt}}%
 & %
\mbox{\input{data/pr/NYUD-v2_test_fb_HED_ap.txt}\hspace{-2.5pt}}%
 \\
SE~\cite{DoZi15}   & %
\mbox{\input{data/pr/NYUD-v2_test_fb_SE_ods_f.txt}\hspace{-2.5pt}}%
 & %
\mbox{\input{data/pr/NYUD-v2_test_fb_SE_ois_f.txt}\hspace{-2.5pt}}%
 & %
\mbox{\input{data/pr/NYUD-v2_test_fb_SE_ap.txt}\hspace{-2.5pt}}%
 \\
SE-all~\cite{Gup+14}   & %
\mbox{\input{data/pr/NYUD-v2_test_fb_SE-all_ods_f.txt}\hspace{-2.5pt}}%
 & %
\mbox{\input{data/pr/NYUD-v2_test_fb_SE-all_ois_f.txt}\hspace{-2.5pt}}%
 & %
\mbox{\input{data/pr/NYUD-v2_test_fb_SE-all_ap.txt}\hspace{-2.5pt}}%
 \\
MCG-B~\cite{Pont-Tuset2016}   & %
\mbox{\input{data/pr/NYUD-v2_test_fb_MCG_ods_f.txt}\hspace{-2.5pt}}%
 &  %
\mbox{\input{data/pr/NYUD-v2_test_fb_MCG_ois_f.txt}\hspace{-2.5pt}}%
 & %
\mbox{\input{data/pr/NYUD-v2_test_fb_MCG_ap.txt}\hspace{-2.5pt}}%
 \\
\bottomrule
\end{tabular}}
\rule{0mm}{1mm}
\resizebox{\textwidth}{!}{%
\begin{tabular}{lccc}
\toprule
       & \multicolumn{3}{c}{Regions - $F_\mathit{op}$} \\
Method &  ODS & OIS & AP\\
\midrule
COB-RGB-HHA  & \bf%
\mbox{\input{data/pr/NYUD-v2_test_fop_COB-rgbHHA_ods_f.txt}\hspace{-2.5pt}}%
 & \bf%
\mbox{\input{data/pr/NYUD-v2_test_fop_COB-rgbHHA_ois_f.txt}\hspace{-2.5pt}}%
 & \bf%
\mbox{\input{data/pr/NYUD-v2_test_fop_COB-rgbHHA_ap.txt}\hspace{-2.5pt}}%
 \\
COB-RGB  & %
\mbox{\input{data/pr/NYUD-v2_test_fop_COB_ods_f.txt}\hspace{-2.5pt}}%
 & %
\mbox{\input{data/pr/NYUD-v2_test_fop_COB_ois_f.txt}\hspace{-2.5pt}}%
 & %
\mbox{\input{data/pr/NYUD-v2_test_fop_COB_ap.txt}\hspace{-2.5pt}}%
 \\
COB-PC  & %
\mbox{\input{data/pr/NYUD-v2_test_fop_COB-PC_ods_f.txt}\hspace{-2.5pt}}%
 & %
\mbox{\input{data/pr/NYUD-v2_test_fop_COB-PC_ois_f.txt}\hspace{-2.5pt}}%
 & %
\mbox{\input{data/pr/NYUD-v2_test_fop_COB-PC_ap.txt}\hspace{-2.5pt}}%
 \\
SE-all~\cite{Gup+14}  & %
\mbox{\input{data/pr/NYUD-v2_test_fop_HHA_ods_f.txt}\hspace{-2.5pt}}%
 & %
\mbox{\input{data/pr/NYUD-v2_test_fop_HHA_ois_f.txt}\hspace{-2.5pt}}%
 & %
\mbox{\input{data/pr/NYUD-v2_test_fop_HHA_ap.txt}\hspace{-2.5pt}}%
 \\
MCG-BSDS500~\cite{Pont-Tuset2016}   & %
\mbox{\input{data/pr/NYUD-v2_test_fop_MCG_ods_f.txt}\hspace{-2.5pt}}%
 & %
\mbox{\input{data/pr/NYUD-v2_test_fop_MCG_ois_f.txt}\hspace{-2.5pt}}%
 & %
\mbox{\input{data/pr/NYUD-v2_test_fop_MCG_ap.txt}\hspace{-2.5pt}}%
 \\
\bottomrule
\end{tabular}}
\rule{0mm}{3mm}

\resizebox{\textwidth}{!}{%
\begin{tabular}{lccc}
\toprule
Ablation:  & \multicolumn{3}{c}{Boundaries - $F_\mathit{b}$} \\
Method &  ODS & OIS & AP\\
\midrule
ResNet50-RGB-HHA  & %
\mbox{\input{data/pr/NYUD-v2_test_fb_ResNet50_nyud_rgbHHA_ods_f.txt}\hspace{-2.5pt}}%
 & \bf%
\mbox{\input{data/pr/NYUD-v2_test_fb_ResNet50_nyud_rgbHHA_ois_f.txt}\hspace{-2.5pt}}%
 & \bf%
\mbox{\input{data/pr/NYUD-v2_test_fb_ResNet50_nyud_rgbHHA_ap.txt}\hspace{-2.5pt}}%
 \\
ResNet50-RGB  & \bf%
\mbox{\input{data/pr/NYUD-v2_test_fb_ResNet50_nyud_rgb_ods_f.txt}\hspace{-2.5pt}}%
 & %
\mbox{\input{data/pr/NYUD-v2_test_fb_ResNet50_nyud_rgb_ois_f.txt}\hspace{-2.5pt}}%
 & %
\mbox{\input{data/pr/NYUD-v2_test_fb_ResNet50_nyud_rgb_ap.txt}\hspace{-2.5pt}}%
 \\
ResNet50-RGBD  & %
\mbox{\input{data/pr/NYUD-v2_test_fb_ResNet50_nyud_rgbd_ods_f.txt}\hspace{-2.5pt}}%
 & %
\mbox{\input{data/pr/NYUD-v2_test_fb_ResNet50_nyud_rgbd_ois_f.txt}\hspace{-2.5pt}}%
 & %
\mbox{\input{data/pr/NYUD-v2_test_fb_ResNet50_nyud_rgbd_ap.txt}\hspace{-2.5pt}}%
 \\
\bottomrule\end{tabular}}

\end{minipage}
\vspace{-1mm}
\caption{\textbf{NYUD-v2 test}: Precision-recall curves for evaluation of boundaries ($F_b$~\cite{Martin2004}), and regions ($F_{op}$~\cite{Pont-Tuset2016a}). ODS, OIS, and AP summary measures.}
\label{fig:pr_nyud}
\end{figure*}

We retrain COB on VOC'12 train set and report the results on the validation set. We use the instance level annotation of the database, and extract contours from the semantic segmentation annotations of the database. The uncertain areas (annotated with value of 255) are treated as background. We compare to several baselines, together with recent state-of-the-art results. Specifically, Khoreva et al.~\cite{Kho+16} retrained HED~\cite{XiTu15} on object contours, and Yang et al.~\cite{Yan+16} proposed a novel encoder-decoder architecture to tackle the same task. We evaluate the best pre-computed results provided by the authors in both cases. The results are quantified in Figure~\ref{fig:pr_pascal_obj}. We observe that COB obtains state-of-the-art results in all metrics. CEDN~\cite{Yan+16} performs better in the high precision regime. However, the authors used extra images from the SBD dataset~\cite{Har+11} for training their detector. Also, CEDN is trained on an improved version of the ground truth, aligning the uncertain areas of VOC'12 with the the true image boundaries by using a CRF. We report results of COB trained only on VOC'12 train set, to be consistent with the results of Khoreva et al.~\cite{Kho+16}. In this experiment, we use \texttt{maxDist} of 0.01, as is adopted by the literature~\cite{UiFe15,Kho+16}.

Figure~\ref{fig:qual_obj_cont} illustrates some qualitative results, as well as the differences of generic segmentation and object boundary detection. We show our results on images of the VOC'12 val set using the model trained on PASCAL Context for generic image segmentation, and we compare qualitatively to the model retrained on the 20 classes of VOC'12 for object boundary detection. In the latter case, the detections are focused on the 20 object classes, disregarding strong contour cues of the background that are detected by the generic segmentation model.

\subsection{RGB-D boundary detection on NYUD-v2 dataset}
\label{sec:exp:rgbd_boundary}
\begin{figure*}
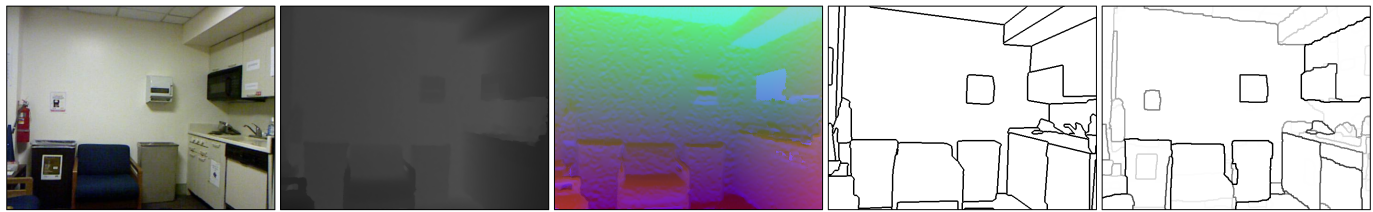

\centering
\resizebox{\linewidth}{!}{\setlength{\fboxsep}{0pt}\noindent%
\showonerow{img_5001}{img}{depth}{HHA}{gt}{result}{rgbd}{0.21}}\\[0.8mm]
\caption{\textbf{Data and results on NYUD-v2}. From left to right: RGB image, depth, HHA features~\cite{Gup+14}, ground truth, and COB detections}
\label{fig:qual_rgbd}
\end{figure*}
The NYUD (v2) dataset~\cite{Sil+12} consists of 1449 RGB-D indoor images, divided into splits of 795 training and 654 testing images, with the corresponding semantic and instance level segmentations. Gupta et al.~\cite{GAM13} adopted this dataset for contour detection. In their experiments, they obtained the respective boundary annotations from the instance-level segmentations of the dataset. We evaluated the performance by using the standard benchmarks of BSDS. Following~\cite{DoZi15, XiTu15, Gup+14}, we increased the tolerance for incorrect localizations from 0.0075 of the image diagonal to 0.11, to compensate for inaccurate annotations of boundaries. 

We use the extra information of depth to train different variants of COB on the NYUD dataset. Gupta et al.~\cite{Gup+14} used the camera parameters of the images to encode the depth information in three channels: horizontal disparity, height above ground, and the angle of the local surface normal with the inferred gravity direction at each pixel (HHA). We retrain three different variants of the CNN: (a) Only using RGB data (ResNet50-RGB), (b) Incorporating depth information into a fourth channel (ResNet50-RGBD), and (c) Concatenating RGB and HHA channels and operate on 6 channels directly (ResNet50-RGB-HHA). Figure~\ref{fig:qual_rgbd} illustrates an overview of the data, along with depth and HHA features that we used, as well as the results obtained by COB. In Figure~\ref{fig:pr_nyud} we show the ablation analysis by directly evaluating the CNN output, without any post-processing. We observe that the CNNs retrained on RGB and RGB-HHA channels obtain significantly better results than the one trained on RGB-D data, showing that HHA features provide an appropriate encoding for depth information.  We retrain the full pipeline of COB (including orientations) on NYUD and we report the precision-recall curves. We compare with various state-of-the-art methods, showing significant improvements. Specifically, we compare with the SE~\cite{DoZi15} detector retrained on the RGB-D data of NYUD, the detector proposed by~\cite{Gup+14} trained on RGB and depth normal gradients, and the best result reported on NYUD by HED~\cite{XiTu15}, where the authors trained two different variants of the detector on RGB and HHA modalities respectively and averaged the obtained results. For completeness, we report results obtained by the original MCG~\cite{Pont-Tuset2016} without any retraining on NYUD. The best result is obtained by the variant of COB trained on both RGB and HHA modalities. Compared to its RGB-only counterpart, the particular model achieves higher accuracy, suggesting that the depth embeddings are useful cues to discern contours when RGB modality alone is unable to do so. It is noteworthy that the post-processing step (orientations and UCMs) further boosts the performance of COB. For example, performance increases from 0.745 (ResNet50-RGB-HHA) to 0.784 (COB-RGB-HHA) by plugging in the orientations and the UCM pipeline to the trained ResNet50 architecture. We also report the results of the model trained on PASCAL Context (COB-PC) and operating only on RGB data, showing that it performs fairly well without any retraining on the NYUD dataset.

\subsection{Efficiency Analysis}
\label{sec:exp:speed}
Contour detection and image segmentation, as a preprocessing step towards high-level applications, need to 
be computationally efficient. 
The previous state-of-the-art in hierarchical image segmentation~\cite{Pont-Tuset2016,Arb+11} was of limited use in
practice due to its computational load.

As a core in our system, the forward pass of our network to compute the contour strength and 8 orientations
takes 0.28 seconds on a NVidia Titan X GPU.
Table~\ref{table:timing} shows the timing comparison between the full system COB (Ours) and some related baselines on PASCAL Context. We divide
the timing into different relevant parts, namely, the contour detection step, the Oriented Watershed Transform (OWT) and Ultrametric Contour Map (UCM) computation, and the globalization (normalized cuts) step.

\begin{table}[h]
\setlength{\tabcolsep}{4pt} % General space between cols (6pt standard)
\center
\footnotesize
\resizebox{0.96\linewidth}{!}{%
\begin{tabular}{lcccc}
\toprule
Steps          		& (1) MCG~\cite{Pont-Tuset2016} 	& (2) MCG-HED 	& (3) Fast UCMs  & (4) COB (Ours) \\
\midrule
Contours 			&  \ \,3.08 				&\ \ \ 0.39*	& \ \ \ 0.39*    & \,\ 0.28*      \\
OWT, UCM   			&     11.33 				&     11.58 	& \ \,  1.63     &     0.51       \\
Globalize 			&  \ \,9.96 				&  \ \,9.97 	& \ \,  9.92     &     0.00       \\
\midrule
Total Time    		&     24.37				&     21.94	&      11.94     &  \textbf{0.79} \\
\bottomrule
\end{tabular}
}
\vspace{2mm}
\caption{\textbf{Timing experiments}: Comparing our approach to different baselines. Times computed using a GPU are marked with an asterisk.}
\label{table:timing}
\end{table}

Column (1) shows the timing for the original MCG~\cite{Pont-Tuset2016}, which uses Structured Edges (SE)~\cite{DoZi15}. As a first baseline, Column (2) displays the timing of MCG if we naively substitute SE by HED~\cite{XiTu15} at the three scales (running on a GPU).
By applying the sparse boundaries representation we reduce the UCM and OWT time from 11.58 to 1.63 seconds (Column (3)). 
Our final technique COB, in which we remove the globalization step, computes the three scales in one pass and add
contour orientations, takes 0.79 seconds in mean.
Overall, comparing to previous state-of-the-art, we get a significant improvement at a fraction of the computation time (24.37 to 0.79 seconds).

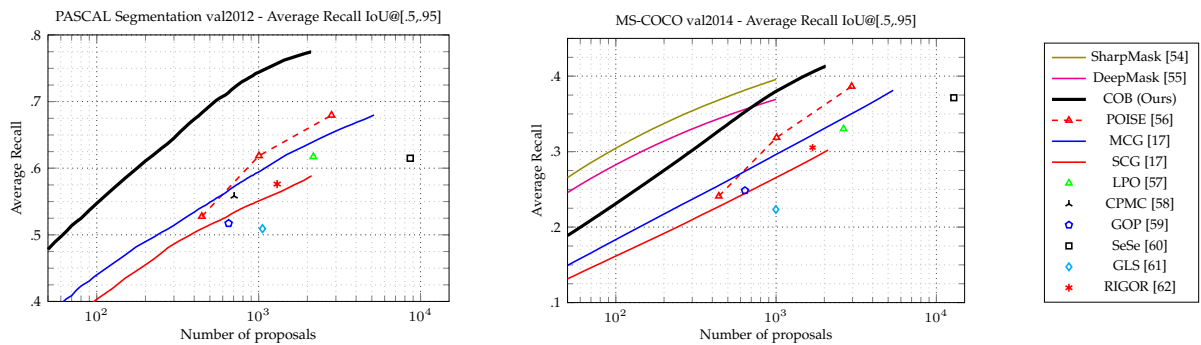
\begin{figure*}
\centering
\hspace{-20mm}
\begin{minipage}[b]{0.425\linewidth}
\centering
\resizebox{0.765\linewidth}{!}{%
\begin{tikzpicture}[/pgfplots/width=1.1\linewidth, /pgfplots/height=0.8\linewidth]
    \begin{axis}[ymin=0.4,ymax=0.8,xmin=50,xmax=15000,enlargelimits=false,
        xlabel=Number of proposals,
        ylabel=Average Recall,
        font=\scriptsize, grid=both,
        grid style=dotted,
        axis equal image=false,
        ytick={0,0.1,...,1},
        yticklabels={0,.1,.2,.3,.4,.5,.6,.7,.8,.9,1},
        minor ytick={0,0.025,...,1},
        major grid style={white!20!black},
        minor grid style={white!70!black},
        xlabel shift={-3pt},
        ylabel shift={-4pt},
        xmode=log,
     	% Title
		title style={yshift=-1ex,},
		title={PASCAL Segmentation val2012 - Average Recall IoU@[.5,.95]}]]

		\addplot+[black,mark=none, ultra thick]                          table[x=ncands,y=average_recall] {data/obj_cands/Pascal_Segmentation_val_2012_COB.txt};
        \addplot+[red,dashed,mark=triangle, mark options={solid}, mark size=1.8, thick]                          table[x=ncands,y=average_recall] {data/obj_cands/Pascal_Segmentation_val_2012_POISE.txt};
        \addplot+[blue,solid,mark=none, thick]                          table[x=ncands,y=average_recall] {data/obj_cands/Pascal_Segmentation_val_2012_MCG.txt};
        \addplot+[red,solid,mark=none, thick]                                  table[x=ncands,y=average_recall] {data/obj_cands/Pascal_Segmentation_val_2012_SCG.txt};
        \addplot+[only marks,blue,solid,mark=pentagon,mark size=1.8, thick]    table[x=ncands,y=average_recall] {data/obj_cands/Pascal_Segmentation_val_2012_GOP.txt};
	  	\addplot+[only marks,cyan,solid,mark=diamond,mark size=1.9, thick]      table[x=ncands,y=average_recall] {data/obj_cands/Pascal_Segmentation_val_2012_GLS.txt};
	    \addplot+[only marks,red,solid,mark=asterisk, mark size=2, thick]   table[x=ncands,y=average_recall] {data/obj_cands/Pascal_Segmentation_val_2012_RIGOR.txt};
	    \addplot+[only marks,black,solid,mark=square, mark size=1.45, thick]    table[x=ncands,y=average_recall] {data/obj_cands/Pascal_Segmentation_val_2012_SeSe.txt};
	    \addplot+[only marks,green,solid,mark=triangle, mark size=1.8, thick]    table[x=ncands,y=average_recall] {data/obj_cands/Pascal_Segmentation_val_2012_LPO.txt};
	    \addplot+[only marks,black,solid,mark=Mercedes star,mark size=2, thick]    table[x=ncands,y=average_recall] {data/obj_cands/Pascal_Segmentation_val_2012_CPMC.txt};

	\end{axis}
   \end{tikzpicture}}
\end{minipage}
\begin{minipage}[b]{0.43\linewidth}
\centering
\resizebox{1.145\linewidth}{!}{%
\begin{tikzpicture}[/pgfplots/width=1.1\linewidth, /pgfplots/height=0.8\linewidth]
    \begin{axis}[ymin=0.1,ymax=0.45,xmin=50,xmax=15000,enlargelimits=false,
        xlabel=Number of proposals,
        ylabel=Average Recall,
        font=\scriptsize, grid=both,
        grid style=dotted,
        axis equal image=false,
        legend style={at={(1.2,0)},anchor=south west},
        ytick={0,0.1,...,1},
        yticklabels={0,.1,.2,.3,.4,.5,.6,.7,.8,.9,1},
        minor ytick={0,0.025,...,1},
        major grid style={white!20!black},
        minor grid style={white!70!black},
        xlabel shift={-3pt},
        ylabel shift={-4pt},
        xmode=log,
     	% Title
		title style={yshift=-1ex,},
		title={MS-COCO val2014 - Average Recall IoU@[.5,.95]}]]
		
		\addplot+[olive,solid,mark=none,thick]                          coordinates {( 1, 0.7)( 1, 0.8)};
	    \addplot+[magenta,solid,mark=none,thick]                          coordinates {( 1, 0.7)( 1, 0.8)};
	    
		\addplot[black,solid,mark=none, ultra thick]                          coordinates {( 1, 0.7)( 1, 0.8)};
        \label{fig:recall:ours}
        \addplot[red,dashed,mark=triangle, mark options={solid}, mark size=1.8, thick]                          coordinates {( 1, 0.7)( 1, 0.8)};
        \label{fig:recall:poise}
        \addplot[blue,solid,mark=none, thick]                          coordinates {( 1, 0.7)( 1, 0.8)};
        \label{fig:recall:mcg}
        \addplot[red,solid,mark=none, thick]                                  coordinates {( 1, 0.7)( 1, 0.8)};
        \label{fig:recall:scg}
	    \addplot[only marks,green,solid,mark=triangle, mark size=1.8, thick]                                coordinates {( 1, 0.7)( 1, 0.8)};
	    \addplot[only marks,black,solid,mark=Mercedes star, mark size=2, thick]                                coordinates {( 1, 0.7)( 1, 0.8)};
        \addplot[only marks,blue,solid,mark=pentagon,mark size=1.8, thick]    coordinates {( 1, 0.7)( 1, 0.8)};
        \label{fig:recall:gop}
        \addplot[only marks,black,solid,mark=square, mark size=1.45, thick]    coordinates {( 1, 0.7)( 1, 0.8)};
	    \label{fig:recall:sese}
	  	\addplot[only marks,cyan,solid,mark=diamond,mark size=1.9, thick]      coordinates {( 1, 0.7)( 1, 0.8)};
		\label{fig:recall:gls}
	    \addplot[only marks,red,solid,mark=asterisk, mark size=2, thick]   coordinates {( 1, 0.7)( 1, 0.8)};
	    \label{fig:recall:rigor}
	    
        \addlegendentry{SharpMask~\cite{Pinheiro2016}}
		\addlegendentry{DeepMask~\cite{Pinheiro2015}}
        \addlegendentry{COB (Ours)}
        \addlegendentry{POISE~\cite{Humayun2015}}
		\addlegendentry{MCG~\cite{Pont-Tuset2016}}
        \addlegendentry{SCG~\cite{Pont-Tuset2016}}
	    \addlegendentry{LPO~\cite{Kraehenbuehl2015}}
	    \addlegendentry{CPMC~\cite{Carreira2012b}}
        \addlegendentry{GOP~\cite{Kraehenbuehl2014}}
		\addlegendentry{SeSe~\cite{Uijlings2013}}
		\addlegendentry{GLS~\cite{Rantalankila2014}}
	    \addlegendentry{RIGOR~\cite{Humayun2014}}
	    
	    \addplot+[olive,solid,mark=none,thick]                          table[x=ncands,y=average_recall] {data/obj_cands/COCO_val2014_SharpMaskRegions.txt};
	    \addplot+[magenta,solid,mark=none,thick]                          table[x=ncands,y=average_recall] {data/obj_cands/COCO_val2014_DeepMaskRegions.txt};
		    
	    \addplot+[black,solid,mark=none, ultra thick]                          table[x=ncands,y=average_recall] {data/obj_cands/COCO_val2014_COB.txt};
        \addplot+[red,dashed,mark=triangle, mark options={solid}, mark size=1.8, thick]                          table[x=ncands,y=average_recall] {data/obj_cands/COCO_val2014_POISE.txt};
        \addplot+[blue,solid,mark=none, thick]                          table[x=ncands,y=average_recall] {data/obj_cands/COCO_val2014_MCG.txt};
        \addplot+[red,solid,mark=none, thick]                                  table[x=ncands,y=average_recall] {data/obj_cands/COCO_val2014_SCG.txt};
        \addplot+[only marks,blue,solid,mark=pentagon,mark size=1.8, thick]    table[x=ncands,y=average_recall] {data/obj_cands/COCO_val2014_GOP.txt};
	  	\addplot+[only marks,cyan,solid,mark=diamond,mark size=1.9, thick]      table[x=ncands,y=average_recall] {data/obj_cands/COCO_val2014_GLS.txt};
	    \addplot+[only marks,red,solid,mark=asterisk, mark size=1.8, thick]   table[x=ncands,y=average_recall] {data/obj_cands/COCO_val2014_RIGOR.txt};
	    \addplot+[only marks,black,solid,mark=square, mark size=1.45, thick]    table[x=ncands,y=average_recall] {data/obj_cands/COCO_val2014_SeSe.txt};
	    \addplot+[only marks,green,solid,mark=triangle, mark size=1.8, thick]    table[x=ncands,y=average_recall] {data/obj_cands/COCO_val2014_LPO.txt};
	\end{axis}
   \end{tikzpicture}}
\end{minipage}
\vspace{-4mm}
\caption{\textbf{Segmented object proposals evaluation on PASCAL Segmentation val and MS-COCO val}: Dashed lines refer to methods that do not provide a ranked set of proposals, but they need to be re-parameterized.}
\label{fig:proposals_eval}
\end{figure*}

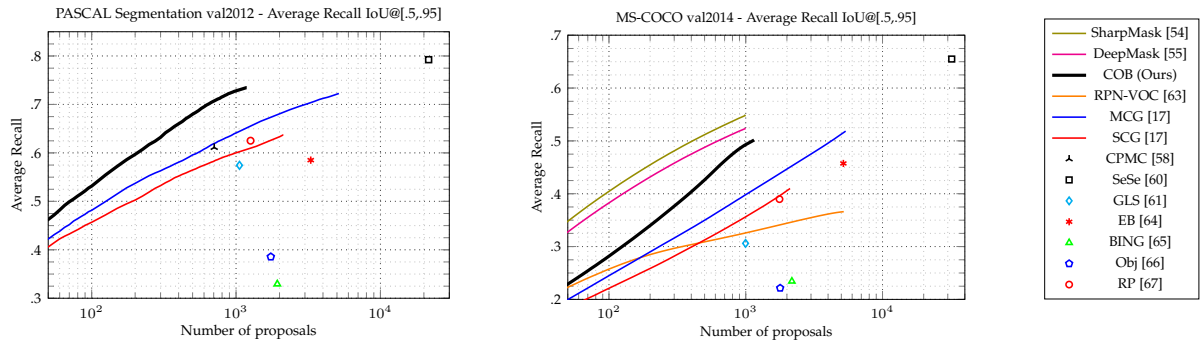
\begin{figure*}
\hspace{-3mm}
\centering
\hspace{-20mm}
\begin{minipage}[b]{0.425\linewidth}
\centering
\resizebox{0.765\linewidth}{!}{%
\begin{tikzpicture}[/pgfplots/width=1.1\linewidth, /pgfplots/height=0.8\linewidth]
   \begin{axis}[ymin=0.3,ymax=0.85,xmin=50,xmax=30000,enlargelimits=false,
        xlabel=Number of proposals,
        ylabel=Average Recall,
        font=\scriptsize, grid=both,
        grid style=dotted,
        axis equal image=false,
        ytick={0,0.1,...,1},
        yticklabels={0,.1,.2,.3,.4,.5,.6,.7,.8,.9,1},
        minor ytick={0,0.025,...,1},
        major grid style={white!20!black},
        minor grid style={white!70!black},
        xlabel shift={-3pt},
        ylabel shift={-4pt},
        xmode=log,
     	% Title
		title style={yshift=-1ex,},
		title={PASCAL Segmentation val2012 - Average Recall IoU@[.5,.95]}]]

		\addplot+[black,mark=none, ultra thick]                          			table[x=ncands,y=average_recall] {data/obj_cands_box/Pascal_Segmentation_val_2012_COB.txt};
        \addplot+[blue,solid,mark=none, thick]                         			table[x=ncands,y=average_recall] {data/obj_cands_box/Pascal_Segmentation_val_2012_MCG.txt};
        \addplot+[red,solid,mark=none, thick]                                  	table[x=ncands,y=average_recall] {data/obj_cands_box/Pascal_Segmentation_val_2012_SCG.txt};
	  	\addplot+[only marks,red,solid,mark=asterisk, mark size=1.8, thick]     	table[x=ncands,y=average_recall] {data/obj_cands_box/Pascal_Segmentation_val_2012_EB.txt};
	  	\addplot+[only marks,cyan,solid,mark=diamond,mark size=1.9, thick]      	table[x=ncands,y=average_recall] {data/obj_cands_box/Pascal_Segmentation_val_2012_GLS.txt};
	    \addplot+[only marks,green,solid,mark=triangle, mark size=1.8, thick]     table[x=ncands,y=average_recall] {data/obj_cands_box/Pascal_Segmentation_val_2012_BING.txt};
	   	\addplot+[only marks,black,solid,mark=Mercedes star, mark size=2, thick]  table[x=ncands,y=average_recall] {data/obj_cands_box/Pascal_Segmentation_val_2012_CPMC.txt};
	    \addplot+[only marks,black,solid,mark=square, mark size=1.45, thick]    	table[x=ncands,y=average_recall] {data/obj_cands_box/Pascal_Segmentation_val_2012_SeSe.txt};
	    \addplot+[only marks,red,solid,mark=o,mark size=1.6, thick]    		    table[x=ncands,y=average_recall] {data/obj_cands_box/Pascal_Segmentation_val_2012_RP.txt};
	    \addplot+[only marks,blue,solid,mark=pentagon,mark size=1.8, thick]     	table[x=ncands,y=average_recall] {data/obj_cands_box/Pascal_Segmentation_val_2012_Obj.txt};
	\end{axis}
   \end{tikzpicture}}
\end{minipage}
%\hspace{-3mm}
\begin{minipage}[b]{0.43\linewidth}
\centering
\resizebox{1.145\linewidth}{!}{%
\begin{tikzpicture}[/pgfplots/width=1.1\linewidth, /pgfplots/height=0.8\linewidth]
    \begin{axis}[ymin=0.2,ymax=0.7,xmin=50,xmax=40000,enlargelimits=false,
        xlabel=Number of proposals,
        ylabel=Average Recall,
        font=\scriptsize, grid=both,
        grid style=dotted,
        axis equal image=false,
        legend style={at={(1.2,0)},anchor=south west},
        ytick={0,0.1,...,1},
        yticklabels={0,.1,.2,.3,.4,.5,.6,.7,.8,.9,1},
        minor ytick={0,0.025,...,1},
        major grid style={white!20!black},
        minor grid style={white!70!black},
        xlabel shift={-3pt},
        ylabel shift={-4pt},
        xmode=log,
     	% Title
		title style={yshift=-1ex,},
		title={MS-COCO val2014 - Average Recall IoU@[.5,.95]}]]
		
		\addplot+[olive,solid,mark=none,thick]                          coordinates {( 1, 0.7)( 1, 0.8)};
	    \addplot+[magenta,solid,mark=none,thick]                          coordinates {( 1, 0.7)( 1, 0.8)};
	    
		\addplot[black,solid,mark=none, ultra thick]                          coordinates {( 1, 0.7)( 1, 0.8)};
        \label{fig:recall:ours_coco}
        \addplot[orange,solid,mark=none,thick]                          coordinates {( 1, 0.7)( 1, 0.8)};

        \addplot[blue,solid,mark=none, thick]                          coordinates {( 1, 0.7)( 1, 0.8)};
        \label{fig:recall:mcg_coco}
        \addplot[red,solid,mark=none, thick]                                  coordinates {( 1, 0.7)( 1, 0.8)};
        \label{fig:recall:scg_coco}
   	    \addplot[only marks,black,solid,mark=Mercedes star, mark size=2, thick]                                coordinates {( 1, 0.7)( 1, 0.8)};
        \addplot[only marks,black,solid,mark=square, mark size=1.45, thick]    coordinates {( 1, 0.7)( 1, 0.8)};
	    \label{fig:recall:sese_coco}
	  	\addplot[only marks,cyan,solid,mark=diamond,mark size=1.9, thick]      coordinates {( 1, 0.7)( 1, 0.8)};
		\label{fig:recall:gls_coco}
        \addplot[only marks,red,solid,mark=asterisk, mark size=1.8, thick]   coordinates {( 1, 0.7)( 1, 0.8)};
	    \addplot[only marks,green,solid,mark=triangle, mark size=1.8, thick]                                coordinates {( 1, 0.7)( 1, 0.8)};
        \addplot[only marks,blue,solid,mark=pentagon,mark size=1.8, thick]    coordinates {( 1, 0.7)( 1, 0.8)};
                \addplot[only marks,red,solid,mark=o,mark size=1.6, thick]    coordinates {( 1, 0.7)( 1, 0.8)};
        \addlegendentry{SharpMask~\cite{Pinheiro2016}}
		\addlegendentry{DeepMask~\cite{Pinheiro2015}}
        \addlegendentry{COB (Ours)}
		\addlegendentry{RPN-VOC~\cite{Ren2015}}
		\addlegendentry{MCG~\cite{Pont-Tuset2016}}
        \addlegendentry{SCG~\cite{Pont-Tuset2016}}
	    \addlegendentry{CPMC~\cite{Carreira2012b}}
		\addlegendentry{SeSe~\cite{Uijlings2013}}
		\addlegendentry{GLS~\cite{Rantalankila2014}}
	    \addlegendentry{EB~\cite{Zitnick2014}}
        \addlegendentry{BING~\cite{Cheng2014}}
	    \addlegendentry{Obj~\cite{Alexe2012}}
	    \addlegendentry{RP~\cite{Manen2013}}
	    
	    	    \addplot+[olive,solid,mark=none,thick]                          table[x=ncands,y=average_recall] {data/obj_cands_box/COCO_val2014_SharpMaskBoxes.txt};
	    \addplot+[magenta,solid,mark=none,thick]                          table[x=ncands,y=average_recall] {data/obj_cands_box/COCO_val2014_DeepMaskBoxes.txt};
	    
	    \addplot+[black,solid,mark=none, ultra thick]                          table[x=ncands,y=average_recall] {data/obj_cands_box/COCO_val2014_COB.txt};
	    \addplot+[orange,solid,mark=none,thick]                          table[x=ncands,y=average_recall] {data/obj_cands_box/COCO_val2014_RPN_VOC07.txt};
        \addplot+[blue,solid,mark=none, thick]                          table[x=ncands,y=average_recall] {data/obj_cands_box/COCO_val2014_MCG.txt};
        \addplot+[red,solid,mark=none, thick]                                  table[x=ncands,y=average_recall] {data/obj_cands_box/COCO_val2014_SCG.txt};
	  	\addplot+[only marks,cyan,solid,mark=diamond,mark size=1.9, thick]      table[x=ncands,y=average_recall] {data/obj_cands_box/COCO_val2014_GLS.txt};
	    \addplot+[only marks,black,solid,mark=square, mark size=1.45, thick]    table[x=ncands,y=average_recall] {data/obj_cands_box/COCO_val2014_SeSe.txt};
	    \addplot+[only marks,blue,solid,mark=pentagon,mark size=1.8, thick]     table[x=ncands,y=average_recall] {data/obj_cands_box/COCO_val2014_Obj.txt};
	  	\addplot+[only marks,red,solid,mark=asterisk, mark size=1.8, thick]     table[x=ncands,y=average_recall] {data/obj_cands_box/COCO_val2014_EB.txt};
	   \addplot+[only marks,green,solid,mark=triangle, mark size=1.8, thick]      table[x=ncands,y=average_recall] {data/obj_cands_box/COCO_val2014_BING.txt};
	    \addplot+[only marks,red,solid,mark=o,mark size=1.6, thick]     table[x=ncands,y=average_recall] {data/obj_cands_box/COCO_val2014_RP.txt};

	\end{axis}
   \end{tikzpicture}}
\end{minipage}
\vspace{-4mm}
\caption{\textbf{Bounding-box object proposals evaluation on PASCAL Segmentation val and MS-COCO val}: Note that COB is designed to detect segmented object proposals and not bounding-box proposals.}
\label{fig:proposals_eval_boxes}
\end{figure*}

\section{Experiments on High-Level Applications}
\label{sec:exp_high}
This section is dedicated to present the interaction of COB boundaries and segments with higher vision tasks. In Section~\ref{sec:exp_high:prop} we evaluate COB as object proposals by plugging in the detected UCMs into the combinatorial goruping pipeline of MCG~\cite{Pont-Tuset2016}.
In Section~\ref{sec:exp_high:semantic_contours} we study the interplay of our boundary detector with  semantic contours and semantic segmentation by combining COB with Dilated Network~\cite{YuKo15} and PSPNet~\cite{Zha+16}, and in Section~\ref{sec:exp_high:obj_detection} we couple the COB proposals with the Fast-RCNN~\cite{Girshick2015} pipeline for object detection. In all cases, we show that COB co-operates well with existing approaches by improving their performance.

\subsection{Object Proposals}
\label{sec:exp_high:prop}
Object proposals are an integral part of current object detection and semantic segmentation pipelines~\cite{Girshick2014,Girshick2015,Ren2015}, as they provide a reduced search space on locations, scales, and shapes over the image.
This section evaluates COB as a segmented and bounding box proposal technique, when using our
high-quality region hierarchies in conjunction with the combinatorial grouping framework of MCG~\cite{Pont-Tuset2016}.
In terms of segmented object proposals, we compare against the most recent techniques SharpMask~\cite{Pinheiro2016}, DeepMask~\cite{Pinheiro2015}, POISE~\cite{Humayun2015}, MCG and SCG~\cite{Pont-Tuset2016}, LPO~\cite{Kraehenbuehl2015}, GOP~\cite{Kraehenbuehl2014}, SeSe~\cite{Uijlings2013}, GLS~\cite{Rantalankila2014}, and RIGOR~\cite{Humayun2014}.
In terms of bounding box proposals, we compare also against  SharpMask~\cite{Pinheiro2016}, DeepMask~\cite{Pinheiro2015}, EB~\cite{Zitnick2014}, RPN~\cite{Ren2015} MCG and SCG~\cite{Pont-Tuset2016}, LPO~\cite{Kraehenbuehl2015}, BING~\cite{Cheng2014}, SeSe~\cite{Uijlings2013}, GLS~\cite{Rantalankila2014}, RIGOR~\cite{Humayun2014}, Obj~\cite{Alexe2012}, and RP~\cite{Manen2013}.
Recent thorough comparisons of object proposal generation methods can be found in~\cite{Pont-Tuset2015b,Hosang2015}.

We perform experiments on the PASCAL 2012 Segmentation dataset~\cite{Eve+12} and on the bigger and more challenging MS-COCO~\cite{Lin2014a} (val2014 set). The hierarchies and combinatorial grouping are trained on PASCAL Context. To assess the generalization capability, we evaluate on MS-COCO, which contains a large number of previously unseen categories, without further retraining.

Figure~\ref{fig:proposals_eval} shows the average recall~\cite{Hosang2015} with respect to the number of object proposals.
In PASCAL VOC'12 Segmentation, the absolute gap of improvement of COB is at least of +13\% with the second-best technique, and consistent in all the range of number of proposals.
In MS-COCO, even though we did not train on any MS-COCO image, COB reaches competitive results for the task, with only very recent techniques~\cite{Pinheiro2016,Pinheiro2015} reaching higher Average Recall when evaluating a low number of proposals.
This shows that our contours, regions, and proposals are properly learning a generic concept of object rather than some specific categories.

Figure~\ref{fig:proposals_eval_boxes} shows the evaluation in terms of bounding box object proposals. COB is less competitive in terms of box proposals, however the algorithm was not specifically designed for detecting bounding boxes. We also show the comparison to RPN~\cite{Ren2015}, which is trained on VOC'07, and thus does not generalize well in the classes of COCO.

\subsection{Semantic Boundaries and Semantic Segmentation}
\begin{table*}
\setlength{\tabcolsep}{0.2em}
\resizebox{.95\textwidth}{!}{%
\begin{tabular}{l|cccccccccccccccccccc|c}
\toprule
% ------ Matlab-generated LaTeX code ------
Technique & Plane & Bicycle & Bird & Boat & Bottle & Bus & Car & Cat & Chair & Cow & Table & Dog & Horse & MBike & Person & Plant & Sheep & Sofa & Train & TV & Mean maxF  \\ \midrule
COB-dil        & \bf\ 84.2 & \bf\ 72.3 & \bf\ 81.0 &    \ 64.2 & \bf\ 68.8 &    \ 81.7 &    \ 71.5 & \bf\ 79.4 & \bf\ 55.2 & \bf\ 79.1 &    \ 40.8 & \bf\ 79.9 & \bf\ 80.4 & \bf\ 75.6 & \bf\ 77.3 & \bf\ 54.4   & \bf\ 82.8 &    \ 51.7 &    \ 72.1 &    \ 62.4 & \bf\ 70.7 \\
DilatedConv~\cite{YuKo15}    &    \ 83.7 &    \ 71.8 &    \ 78.8 & \bf\ 65.5 &    \ 66.3 & \bf\ 82.6 & \bf\ 73.0 &    \ 77.3 &    \ 47.3 &    \ 76.8 &    \ 37.2 &    \ 78.4 &    \ 79.4 &    \ 75.2 &    \ 73.8 &    \ 46.2   &    \ 79.5 &    \ 46.6 & \bf\ 76.4 & \bf\ 63.8 &    \ 69.0 \\
BNF~\cite{BST15c}           &    \ 76.7 &    \ 60.5 &    \ 75.9 &    \ 60.7 &    \ 63.1 &    \ 68.4 &    \ 62.0 &    \ 74.3 &    \ 54.1 &    \ 76.0 & \bf\ 42.9 &    \ 71.9 &    \ 76.1 &    \ 68.3 &    \ 70.5 &    \ 53.7   &    \ 79.6 & \bf\ 51.9 &    \ 60.7 &    \ 60.9 &    \ 65.4 \\
HFL~\cite{BST15b}  &    \ 73.6 &    \ 61.1 &    \ 74.2 &    \ 57.0 &    \ 58.7 &    \ 70.2 &    \ 60.8 &    \ 71.8 &    \ 46.3 &    \ 72.1 &    \ 36.0 &    \ 70.9 &    \ 72.9 &    \ 67.5 &    \ 69.9 &    \ 44.1   &    \ 73.1 &    \ 42.2 &    \ 62.2 &    \ 60.4 &    \ 62.2 \\
\cite{Kho+16}        &    \ 65.9 &    \ 54.1 &    \ 63.6 &    \ 47.9 &    \ 47.0 &    \ 60.4 &    \ 50.9 &    \ 56.5 &    \ 40.4 &    \ 56.0 &    \ 30.0 &    \ 57.5 &    \ 58.0 &    \ 57.4 &    \ 59.5 &    \ 39.0   &    \ 64.2 &    \ 35.4 &    \ 51.0 &    \ 42.4 &    \ 51.9 \\
\cite{Har+11}        &    \ 41.5 &    \ 46.7 &    \ 15.6 &    \ 17.1 &    \ 36.5 &    \ 42.7 &    \ 40.3 &    \ 22.6 &    \ 18.8 &    \ 27.0 &    \ 12.5 &    \ 18.2 &    \ 35.4 &    \ 29.4 &    \ 48.1 &    \ 13.8   &    \ 26.9 &    \ 11.0 &    \ 22.0 &    \ 31.3 &    \ 27.9 \\
\bottomrule
% ------ End of Matlab-generated LaTeX code ------
\end{tabular}}
\vspace{1mm}
\caption{SBD val evaluation: Semantic contours results: maximal $F_{b}$ per class and mean maximal $F_{b}$ is reported for all methods.}
%\vspace{-2mm}
\label{tab:sem_cont_mf}
\setlength{\tabcolsep}{0.2em}
\resizebox{.95\textwidth}{!}{%
\begin{tabular}{l|cccccccccccccccccccc|c}
\toprule
% ------ Matlab-generated LaTeX code ------
Technique  & Plane & Bicycle & Bird & Boat & Bottle & Bus & Car & Cat & Chair & Cow & Table & Dog & Horse & MBike & Person & Plant & Sheep & Sofa & Train & TV & Mean AP      \\ \midrule
COB-dil        & \bf\ 85.7 & \bf\ 69.3 & \bf\ 77.6 & \bf\ 59.7 & \bf\ 64.1 & \bf\ 82.9 & \bf\ 69.7 & \bf\ 80.5 & \bf\ 41.8 & \bf\ 79.4 & \bf\ 26.0 & \bf\ 78.9 & \bf\ 81.5 & \bf\ 74.7 & \bf\ 77.3 & \bf\ 43.8   & \bf\ 82.8 & \bf\ 39.3 & \bf\ 73.3 & \bf\ 56.4 & \bf\ 67.2 \\
BNF~\cite{BST15c}            &    \ 75.9 &    \ 46.0 &    \ 70.5 &    \ 48.9 &    \ 48.6 &    \ 65.3 &    \ 53.5 &    \ 65.2 &    \ 38.2 &    \ 69.7 &    \ 20.9 &    \ 62.3 &    \ 72.2 &    \ 56.6 &    \ 63.3 &    \ 38.5   &    \ 75.7 &    \ 31.4 &    \ 45.6 &    \ 48.1 &    \ 54.8 \\
HFL~\cite{BST15b}            &    \ 71.3 &    \ 54.9 &    \ 68.8 &    \ 45.6 &    \ 48.3 &    \ 70.9 &    \ 56.5 &    \ 65.6 &    \ 29.0 &    \ 65.8 &    \ 17.6 &    \ 64.3 &    \ 68.3 &    \ 64.0 &    \ 65.6 &    \ 28.8   &    \ 66.5 &    \ 25.8 &    \ 59.5 &    \ 49.8 &    \ 54.3 \\
\cite{Kho+16}        &    \ 67.1 &    \ 50.5 &    \ 62.2 &    \ 42.1 &    \ 38.9 &    \ 57.8 &    \ 47.7 &    \ 53.7 &    \ 32.1 &    \ 52.3 &    \ 17.5 &    \ 53.1 &    \ 56.0 &    \ 53.2 &    \ 57.7 &    \ 29.4   &    \ 62.2 &    \ 24.0 &    \ 46.2 &    \ 32.8 &    \ 46.8 \\
\cite{Har+11}        &    \ 38.4 &    \ 38.9 &    \ 8.6 &    \ 9.3 &    \ 23.0 &    \ 37.1 &    \ 33.6 &    \ 18.4 &    \ 11.5 &    \ 16.0 &    \ 5.1 &    \ 12.2 &    \ 29.0 &    \ 21.3 &    \ 46.9 &    \ 7.2   &    \ 15.8 &    \ 5.6 &    \ 14.4 &    \ 21.4 &    \ 20.7 \\

\bottomrule
% ------ End of Matlab-generated LaTeX code ------

\end{tabular}}
\vspace{1mm}
\caption{SBD val evaluation: Semantic contours results: Average Precision (AP) per class and mean AP (mAP) is reported for all methods.}
\vspace{-2mm}
\label{tab:sem_cont_ap}
\end{table*}

\begin{table*}
\setlength{\tabcolsep}{0.2em}
\resizebox{.95\textwidth}{!}{%
\begin{tabular}{l|ccccccccccccccccccccc|c}
\toprule
% ------ Matlab-generated LaTeX code ------
Technique & BG & Plane & Bicycle & Bird & Boat & Bottle & Bus & Car & Cat & Chair & Cow &  Table & Dog & Horse & MBike & Person & Plant & Sheep & Sofa & Train & TV & Mean  \\ \midrule
COB-dil &    \bf\ 93.5  &    \bf\ 90.3 &    \bf\ 39.7 &    \bf\ 83.2 &    \bf\ 66.2 &    \bf\ 68.9 &    \bf\ 92.6 &    \bf\ 84.6 &    \bf\ 89.2 &    \bf\ 36.9 &    \bf\ 84.7 &    \bf\ 53.1   &    \bf\ 82.9 &   \bf \ 87.0 &    \bf\ 83.1 &    \bf\ 86.3 &    \bf\ 54.7   &    \bf\ 84.8 &    \bf\ 45.7 &    \bf\ 84.6 &    \bf\ 68.9 &    \bf\ 74.3 \\
DilatedConv~\cite{YuKo15}         &    \ 92.8  &    \ 87.1 &    \ 39.2 &    \ 79.6 &    \ 65.9 &    \ 66.3 &    \ 90.0 &    \ 82.5 &    \ 85.3 &    \ 36.2 &    \ 81.7 &    \ 51.7   &    \ 78.1 &    \ 83.8 &    \ 80.2 &    \ 83.4 &    \ 50.5   &    \ 82.6 &    \ 43.1 &    \ 83.8 &    \ 65.3 &    \ 71.9 \\ \midrule
COB-PSP      & \bf\ 95.4  & \bf\ 90.9 & \bf\ 44.8 &    \ 90.2 & \bf\ 76.1 & \bf\ 84.1 &    \ 96.1 & \bf\ 92.1 & \bf\ 95.3 &    \ 45.6 & \bf\ 95.4 & \bf\ 59.9   & \bf\ 92.0 & \bf\ 93.2 &    \ 90.8 & \bf\ 90.1 & \bf\ 68.0   & \bf\ 93.4 & \bf\ 50.2 & \bf\ 93.3 &    \ 79.8 & \bf\ 81.7 \\
PSPNet~\cite{Zha+16}               &    \ 95.3  &    \ 90.7 &    \ 44.4 & \ 90.2 &    \ 74.8 &    \ 83.4 & \bf\ 96.3 &    \ 92.0 &    \ 95.0 & \bf\ 46.4 &    \ 94.6 &    \ 59.1   &    \ 91.9 &    \ 92.5 & \bf\ 91.0 &    \ 89.9 &    \ 66.0   &    \ 91.6 &    \ 50.2 &    \ 93.0 & \bf\ 80.0 &    \ 81.3 \\
\bottomrule
% ------ End of Matlab-generated LaTeX code ------

\end{tabular}}
\vspace{1mm}
\caption{PASCAL VOC Segmentation val evaluation: Effect of COB on Semantic Segmentation. Per-class IoU and mean IoU are reported.}
\vspace{-2mm}
\label{tab:sem_seg}
\end{table*}

\begin{table*}
\setlength{\tabcolsep}{0.2em}
\resizebox{.95\textwidth}{!}{%
\begin{tabular}{l|cccccccccccccccccccc|c}
\toprule
% ------ Matlab-generated LaTeX code ------
Technique  & Plane & Bicycle & Bird & Boat & Bottle & Bus & Car & Cat & Chair & Cow & Table & Dog & Horse & MBike & Person & Plant & Sheep & Sofa & Train & TV & Mean      \\ \midrule
COB &    \ 69.5 &    \ 76.8 & \bf\ 69.7 &    \ 53.3 &    \bf\ 44.6 &    \bf\ 80.5 & \bf\ 81.3 & \bf\ 83.1 & \bf\ 45.3 &    \bf\ 74.2 & \bf\ 69.4 &    \bf\ 80.1 & \bf\ 84.2 &    \bf\ 76.7 & \bf\ 72.8 & \bf\ 35.9 & \bf\ 67.1 &    \bf\ 68.4 &    \bf\ 75.1 &    \ 65.4 & \bf\ 68.7 \\
SeSe  & \bf\ 76.0 & \ 76.8 &    \ 65.3 &    \bf\ 54.6 &    \ 38.0 &    \ 76.5 &    \ 78.2 &    \ 81.6 &    \ 40.1 &    \ 74.1 &    \ 66.5 &    \ 78.9 &    \ 81.8 &    \ 74.5 &    \ 66.2 &    \ 32.9 &    \ 65.6 &    \ 67.7 &    \ 73.4 &    \bf\ 66.8 &    \ 66.8 \\

\bottomrule
% ------ End of Matlab-generated LaTeX code ------
\end{tabular}}
\vspace{1mm}
\caption{VOC 2007 test evaluation: Object Detection performance (mAP) of Fast-RCNN~\cite{Girshick2015}, using object proposals from~\cite{Uijlings2013} (original) or COB.}
\vspace{-6mm}
\label{tab:voc07_obj_detection}
\end{table*}

\label{sec:exp_high:semantic_contours}
The task of Semantic Boundaries, introduced by~\cite{Har+11}, requires not only detecting the boundaries, but also associating a semantic class to them. It can be thought as a combination of Boundary Detection and Semantic Segmentation, where except for the binary information of boundaries, one needs to label each of the detected pixels with the corresponding semantic class. The common approach to this task is to separately approach semantic segmentation and contour detection, and fuse the results of the two tasks~\cite{Har+11, BST15b, BST15c}. Hariharan et al~\cite{Har+11} tackled the task with generic object detectors and bottom up contours. Bertasius et al.~\cite{BST15b, BST15c} show that results can be significantly improved when using deep-learning based semantic segmenters and contour detectors. Kokkinos~\cite{Kok16} approaches the task with fully-convolutional networks trained end to end, although the results do not reach the current state of the art. 

We also follow the most common approach of mixing the two tasks. We couple the COB boundaries with Semantic Segmentation results by dilated convolutions~\cite{YuKo15}. Specifically, we mask the boundaries with Semantic Segmentation results, with a tolerance of 0.02 of the image diagonal. 

We report results on the SBD~\cite{Har+11} database, for semantic boundary detection, by using the standard benchmark. Tables~\ref{tab:sem_cont_mf} and \ref{tab:sem_cont_ap} compare the results among various methods, in both metrics used in the benchmark (mean maximal F-measure and Average Precision) for all classes. The combination of COB with~\cite{YuKo15}, denoted with COB-dil, achieves state-of-the-art results in both metrics. For fair comparison, we also include the results obtained by evaluating the semantic segmentation results obtained by~\cite{YuKo15} directly as contours. We show that COB fairly improves the result.

Having explored the performance of COB combined with the Dilated Convolution network on Semantic Boundaries, it is interesting to investigate the dual task: the effects of COB in semantic segmentation. We treat the COB UCMs as superpixels, by applying a low value threshold (0.1) to the hierarchy, which results in high recall. We then snap the semantic segmentation results to the superpixels by majority voting of the regions, i.e superpixels that overlap more than 50\% with the semantic class, are assigned the corresponding label. Table~\ref{tab:sem_seg} reports the effects of such snapping on Semantic Segmentation, on the validation split of PASCAL VOC Segmentation dataset. In addition to the Dilated network, we also explored the most recent PSPNet~\cite{Zha+16} as the base semantic segmenter. Results improve consistently almost for all the classes in both cases, indicating that COB superpixels are further refining the semantic segmentation results on boundary locations. We observe a more moderate improvement in the PSPNet results, mainly because of the reduced false detections. We have excluded all images of VOC Segmentation val set for training the COB model.

\begin{figure}
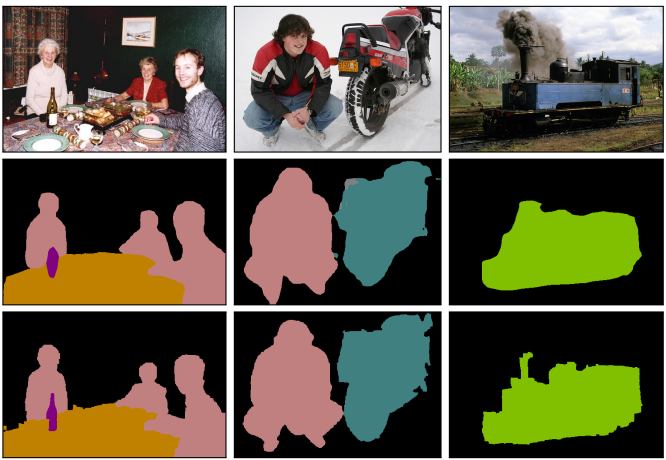

\centering
\resizebox{\linewidth}{!}{\setlength{\fboxsep}{0pt}\noindent%
\showinonecol{2008_000885}{2008_002115}{2008_003105}{img}{sem_seg}{0.21}}\\[0.8mm]
\resizebox{\linewidth}{!}{\setlength{\fboxsep}{0pt}\noindent%
\showinonecol{2008_000885}{2008_002115}{2008_003105}{Dil}{sem_seg}{0.21}}\\[0.8mm]
\resizebox{\linewidth}{!}{\setlength{\fboxsep}{0pt}\noindent%
\showinonecol{2008_000885}{2008_002115}{2008_003105}{snapped}{sem_seg}{0.21}}\\
\caption{\textbf{Qualitative results for Semantic Segmentation}. Row 1: original images, Row 2: Dilated Convolution Network, Row 3: Dilated Network with COB superpixels.}
\label{fig:qual_sem_seg}
\vspace{-3mm}
\end{figure}

In Figure~\ref{fig:qual_sem_seg} we present some qualitative results. Snapping to COB superpixels improves mainly on boundary locations, as well as on noisy semantic segmentation detections in places where COB superpixels are not present.

\subsection{COB Object Proposals for Object Detection}
\label{sec:exp_high:obj_detection}

Object Proposals have been extensively used to facilitate object detection~\cite{Girshick2014,Girshick2015,Ren2015}. Most common pipelines use object proposals in the form of a bounding box to regress a class score and a refined prediction of the bounding box locations. Even though our approach provides segmented object proposals from a hierarchy of regions, it is possible to study their effect on common object detection pipelines by simply extracting the bounding box around them. 

We evaluate the bounding box proposals generated by COB by feeding them into the Fast-RCNN~\cite{Girshick2015} pipeline for Object Detection. The original approach uses the VGG network~\cite{SiZi15} together with the box proposals generated by the Selective Search~\cite{Uijlings2013} algorithm to predict class probability and refine the localization for each of them. The final detection performance is evaluated by performing non-maximum suppression on the detections.

Experiments are performed on the VOC'07 detection database. The database consists of 5011 training, and we report the performance on its 4952 testing images. In our experiments, we change the box proposals of Selective Search, to the ones generated by COB. We keep all the hyper-parameters of the original approach unchanged, both at training and test times. Table~\ref{tab:voc07_obj_detection} quantitatively evaluates the effects of COB proposals in performance. We observe improvements in object detection performance (mean Average Precision - mAP), which further proves the high quality of the proposals generated by COB. We would like to emphasize that the latest developments on Object Detection use joint training of bounding box proposals and object class scores~\cite{Ren2015, Liu+16, Red+16, Dai+16}, which together with training on external data achieves much higher results. Instead, we focus on proving the high quality of COB proposals compared to other object proposal techniques.

\section{Conclusions}
In this work, we have developed an approach to detect contours at multiple scales, together with their orientations, in a single forward pass of a convolutional neural network. We provide a fast framework for generating region hierarchies by efficiently combining multiscale oriented contour detections, thanks to a new sparse boundary representation.
We shift from the BSDS to PASCAL to unwind all the potential of 
data-hungry methods such as CNNs and by observing that BSDS is close to saturation.

Our technique achieves state-of-the-art performance by a significant margin for contour detection, the estimation of their orientation, and generic (RGB and RGB-D) image segmentation. We show that our architecture is modular by using two different CNN base architectures, which suggests that it will be able to transfer further improvements in CNN base architectures to perceptual grouping. We also show that our method does not require globalization, which was a speed bottleneck in previous approaches. The generalization of COB was further demonstrated when applied to high-level vision tasks (object proposals, object detection, and semantic contours and segmentation) in combination with recent pipelines, where the results are improved in all cases.

All our code, CNN models, pre-computed results, dataset splits, and benchmarks are publicly available at \url{www.vision.ee.ethz.ch/\textasciitilde cvlsegmentation/}.

% or


%\appendices
%% you can choose not to have a title for an appendix
%% if you want by leaving the argument blank
%\section{}

\ifCLASSOPTIONcompsoc
  \section*{Acknowledgments}
\else
  \section*{Acknowledgment}
\fi

Research funded by the EU Framework Programme for Research and Innovation Horizon 2020 (Grant No. 645331, EurEyeCase), and by the Swiss Commission for Technology and Innovation (CTI, Grant No. 19015.1 PFES-ES, NeGeVA). The authors gratefully acknowledge support by armasuisse and thank NVidia for donating the GPUs used in this work.

\ifCLASSOPTIONcaptionsoff
  \newpage
\fi

\bibliographystyle{IEEEtran}
\bibliography{pami2017}

\begin{thebibliography}{10}
\providecommand{\url}[1]{#1}
\csname url@samestyle\endcsname
\providecommand{\newblock}{\relax}
\providecommand{\bibinfo}[2]{#2}
\providecommand{\BIBentrySTDinterwordspacing}{\spaceskip=0pt\relax}
\providecommand{\BIBentryALTinterwordstretchfactor}{4}
\providecommand{\BIBentryALTinterwordspacing}{\spaceskip=\fontdimen2\font plus
\BIBentryALTinterwordstretchfactor\fontdimen3\font minus
  \fontdimen4\font\relax}
\providecommand{\BIBforeignlanguage}[2]{{%
\expandafter\ifx\csname l@#1\endcsname\relax
\typeout{** WARNING: IEEEtran.bst: No hyphenation pattern has been}%
\typeout{** loaded for the language `#1'. Using the pattern for}%
\typeout{** the default language instead.}%
\else
\language=\csname l@#1\endcsname
\fi
#2}}
\providecommand{\BIBdecl}{\relax}
\BIBdecl

\bibitem{Kokkinos2016}
I.~Kokkinos, ``Pushing the boundaries of boundary detection using deep
  learning,'' in \emph{ICLR}, 2016.

\bibitem{XiTu15}
S.~Xie and Z.~Tu, ``Holistically-nested edge detection,'' in \emph{ICCV}, 2015.

\bibitem{BST15a}
G.~Bertasius, J.~Shi, and L.~Torresani, ``Deepedge: A multi-scale bifurcated
  deep network for top-down contour detection,'' in \emph{CVPR}, 2015.

\bibitem{BST15b}
------, ``High-for-low and low-for-high: Efficient boundary detection from deep
  object features and its applications to high-level vision,'' in \emph{ICCV},
  2015.

\bibitem{She+15}
W.~Shen, X.~Wang, Y.~Wang, X.~Bai, and Z.~Zhang, ``Deep{C}ontour: A deep
  convolutional feature learned by positive-sharing loss for contour
  detection,'' in \emph{CVPR}, 2015.

\bibitem{GaLe14}
Y.~Ganin and V.~Lempitsky, ``N$^{4}$-fields: Neural network nearest neighbor
  fields for image transforms,'' in \emph{ACCV}, 2014.

\bibitem{Russakovsky2015}
O.~Russakovsky, J.~Deng, H.~Su, J.~Krause, S.~Satheesh, S.~Ma, Z.~Huang,
  A.~Karpathy, A.~Khosla, M.~Bernstein, A.~C. Berg, and L.~Fei-Fei, ``{ImageNet
  Large Scale Visual Recognition Challenge},'' \emph{IJCV}, 2015.

\bibitem{Krizhevsky2012}
A.~Krizhevsky, I.~Sutskever, and G.~E. Hinton, ``Imagenet classification with
  deep convolutional neural networks,'' in \emph{NIPS}, 2012.

\bibitem{Sze+15}
C.~Szegedy, W.~Liu, Y.~Jia, P.~Sermanet, S.~Reed, D.~Anguelov, D.~Erhan,
  V.~Vanhoucke, and A.~Rabinovich, ``Going deeper with convolutions,'' in
  \emph{CVPR}, 2015.

\bibitem{SiZi15}
K.~Simonyan and A.~Zisserman, ``Very deep convolutional networks for
  large-scale image recognition,'' in \emph{ICLR}, 2015.

\bibitem{He+16}
K.~He, X.~Zhang, S.~Ren, and J.~Sun, ``Deep residual learning for image
  recognition,'' in \emph{CVPR}, 2016.

\bibitem{Martin2001}
D.~Martin, C.~Fowlkes, D.~Tal, and J.~Malik, ``A database of human segmented
  natural images and its application to evaluating segmentation algorithms and
  measuring ecological statistics,'' in \emph{ICCV}, 2001.

\bibitem{Eve+12}
M.~Everingham, L.~Van~Gool, C.~K.~I. Williams, J.~Winn, and A.~Zisserman, ``The
  {PASCAL} {V}isual {O}bject {C}lasses {C}hallenge 2012 {(VOC2012)}
  {R}esults,''
  http://www.pascal-network.org/challenges/VOC/voc2012/workshop/index.html.

\bibitem{Har+11}
B.~Hariharan, P.~Arbel{\'a}ez, L.~Bourdev, S.~Maji, and J.~Malik, ``Semantic
  contours from inverse detectors,'' in \emph{ICCV}, 2011.

\bibitem{Mot+14}
R.~Mottaghi, X.~Chen, X.~Liu, N.-G. Cho, S.-W. Lee, S.~Fidler, R.~Urtasun, and
  A.~Yuille, ``The role of context for object detection and semantic
  segmentation in the wild,'' in \emph{CVPR}, 2014.

\bibitem{Arb+11}
P.~Arbel{\'a}ez, M.~Maire, C.~Fowlkes, and J.~Malik, ``Contour detection and
  hierarchical image segmentation,'' \emph{TPAMI}, vol.~33, no.~5, pp.
  898--916, 2011.

\bibitem{Pont-Tuset2016}
J.~Pont-Tuset, P.~Arbel\'{a}ez, J.~Barron, F.Marques, and J.~Malik,
  ``Multiscale combinatorial grouping for image segmentation and object
  proposal generation,'' \emph{TPAMI}, vol.~39, no.~1, pp. 128 -- 140, 2017.

\bibitem{Lin2014a}
T.~Lin, M.~Maire, S.~Belongie, L.~D. Bourdev, R.~B. Girshick, J.~Hays,
  P.~Perona, D.~Ramanan, P.~Doll{\'{a}}r, and C.~L. Zitnick, ``Microsoft
  {COCO:} common objects in context,'' \emph{{arXiv}:1405.0312}, 2014.

\bibitem{Man+16}
K.~Maninis, J.~Pont-Tuset, P.~Arbel\'{a}ez, and L.~V. Gool, ``Deep retinal
  image understanding,'' in \emph{MICCAI}, 2016.

\bibitem{Rob63}
L.~G. Roberts, ``Machine perception of three-dimensional solids,'' Ph.D.
  dissertation, MIT, 1963.

\bibitem{Kit83}
J.~Kittler, ``On the accuracy of the sobel edge detector,'' \emph{Image and
  Vision Computing}, vol.~1, no.~1, pp. 37--42, 1983.

\bibitem{Pre70}
J.~M. Prewitt, ``Object enhancement and extraction,'' \emph{Picture processing
  and Psychopictorics}, vol.~10, no.~1, pp. 15--19, 1970.

\bibitem{MaHi80}
D.~Marr and E.~Hildreth, ``Theory of edge detection,'' \emph{Proc. Royal Soc.
  of London}, vol. 207, no. 1167, pp. 187--217, 1980.

\bibitem{Cann86}
J.~Canny, ``A computational approach to edge detection,'' \emph{TPAMI}, no.~6,
  pp. 679--698, 1986.

\bibitem{Martin2004}
D.~Martin, C.~Fowlkes, and J.~Malik, ``Learning to detect natural image
  boundaries using local brightness, color, and texture cues,'' \emph{TPAMI},
  vol.~26, no.~5, pp. 530--549, 2004.

\bibitem{Kon+03}
S.~Konishi, A.~L. Yuille, J.~M. Coughlan, and S.~C. Zhu, ``Statistical edge
  detection: Learning and evaluating edge cues,'' \emph{TPAMI}, vol.~25, no.~1,
  pp. 57--74, 2003.

\bibitem{DTB06}
P.~Dollar, Z.~Tu, and S.~Belongie, ``Supervised learning of edges and object
  boundaries,'' in \emph{CVPR}, 2006.

\bibitem{Kok10a}
I.~Kokkinos, ``Boundary detection using {F}-measure-, filter-and feature-(f3)
  boost,'' in \emph{ECCV}, 2010.

\bibitem{ReBo12}
X.~Ren and L.~Bo, ``Discriminatively trained sparse code gradients for contour
  detection,'' in \emph{NIPS}, 2012.

\bibitem{LZD13}
J.~J. Lim, C.~L. Zitnick, and P.~Doll{\'a}r, ``Sketch tokens: A learned
  mid-level representation for contour and object detection,'' in \emph{CVPR},
  2013.

\bibitem{DoZi15}
P.~Doll{\'a}r and C.~L. Zitnick, ``Fast edge detection using structured
  forests,'' \emph{TPAMI}, vol.~37, no.~8, pp. 1558--1570, 2015.

\bibitem{BST15c}
G.~Bertasius, J.~Shi, and L.~Torresani, ``Semantic segmentation with boundary
  neural fields,'' in \emph{CVPR}, 2016.

\bibitem{XiTu17}
S.~Xie and Z.~Tu, ``Holistically-nested edge detection,'' \emph{International
  Journal of Computer Vision}, pp. 1--16, 2017.

\bibitem{Ren08}
X.~Ren, ``Multi-scale improves boundary detection in natural images,''
  \emph{ECCV}, 2008.

\bibitem{HaFo15}
S.~Hallman and C.~C. Fowlkes, ``Oriented edge forests for boundary detection,''
  in \emph{CVPR}, 2015.

\bibitem{Kho+16}
A.~Khoreva, R.~Benenson, M.~Omran, M.~Hein, and B.~Schiele, ``Weakly supervised
  object boundaries,'' in \emph{CVPR}, 2016.

\bibitem{Li+16}
Y.~Li, M.~Paluri, J.~M. Rehg, and P.~Doll{\'a}r, ``Unsupervised learning of
  edges,'' in \emph{CVPR}, 2016.

\bibitem{Yan+16}
J.~Yang, B.~Price, S.~Cohen, H.~Lee, and M.-H. Yang, ``Object contour detection
  with a fully convolutional encoder-decoder network,'' in \emph{CVPR}, 2016.

\bibitem{GAM13}
S.~Gupta, P.~Arbel\'{a}ez, and J.~Malik, ``Perceptual organization and
  recognition of indoor scenes from {RGB-D} images,'' in \emph{CVPR}, 2013.

\bibitem{Gup+14}
S.~Gupta, R.~Girshick, P.~Arbel{\'a}ez, and J.~Malik, ``Learning rich features
  from {RGB-D} images for object detection and segmentation,'' in \emph{ECCV},
  2014.

\bibitem{Shi2000}
J.~Shi and J.~Malik, ``Normalized cuts and image segmentation,'' \emph{TPAMI},
  vol.~22, no.~8, 2000.

\bibitem{Iso+14}
P.~Isola, D.~Zoran, D.~Krishnan, and E.~H. Adelson, ``Crisp boundary detection
  using pointwise mutual information,'' in \emph{ECCV}, 2014.

\bibitem{Najman1996}
L.~Najman and M.~Schmitt, ``Geodesic saliency of watershed contours and
  hierarchical segmentation,'' \emph{TPAMI}, vol.~18, no.~12, pp. 1163--1173,
  1996.

\bibitem{Lee+14}
C.-Y. Lee, S.~Xie, P.~Gallagher, Z.~Zhang, and Z.~Tu, ``Deeply-supervised
  nets,'' \emph{arXiv preprint arXiv:1409.5185}, 2014.

\bibitem{Pont-Tuset2016a}
J.~Pont-Tuset and F.~Marques, ``Supervised evaluation of image segmentation and
  object proposal techniques,'' \emph{TPAMI}, vol.~38, no.~7, pp. 1465--1478,
  2016.

\bibitem{Jia+14}
Y.~Jia, E.~Shelhamer, J.~Donahue, S.~Karayev, J.~Long, R.~Girshick,
  S.~Guadarrama, and T.~Darrell, ``Caffe: Convolutional architecture for fast
  feature embedding,'' \emph{arXiv preprint arXiv:1408.5093}, 2014.

\bibitem{Zhao2015}
Q.~Zhao, ``Segmenting natural images with the least effort as humans,'' in
  \emph{BMVC}, 2015.

\bibitem{Ren2013}
Z.~Ren and G.~Shakhnarovich, ``Image segmentation by cascaded region
  agglomeration,'' in \emph{CVPR}, 2013.

\bibitem{Felzenszwalb2004}
P.~F. Felzenszwalb and D.~P. Huttenlocher, ``Efficient graph-based image
  segmentation,'' \emph{IJCV}, vol.~59, p. 2004, 2004.

\bibitem{Comaniciu2002}
D.~Comaniciu and P.~Meer, ``Mean shift: a robust approach toward feature space
  analysis,'' \emph{TPAMI}, vol.~24, no.~5, pp. 603 --619, 2002.

\bibitem{Maninis2016a}
K.~Maninis, J.~Pont-Tuset, P.~Arbel\'{a}ez, and L.~V. Gool, ``Convolutional
  oriented boundaries,'' in \emph{ECCV}, 2016.

\bibitem{UiFe15}
J.~Uijlings and V.~Ferrari, ``Situational object boundary detection,'' in
  \emph{CVPR}, 2015.

\bibitem{Sil+12}
P.~K. Nathan~Silberman, Derek~Hoiem and R.~Fergus, ``Indoor segmentation and
  support inference from rgbd images,'' in \emph{ECCV}, 2012.

\bibitem{Pinheiro2016}
P.~O. Pinheiro, T.-Y. Lin, R.~Collobert, and P.~Doll{\'a}r, ``Learning to
  refine object segments,'' in \emph{ECCV}, 2016.

\bibitem{Pinheiro2015}
P.~O. Pinheiro, R.~Collobert, and P.~Dollar, ``Learning to segment object
  candidates,'' in \emph{NIPS}, 2015.

\bibitem{Humayun2015}
A.~Humayun, F.~Li, and J.~M. Rehg, ``The middle child problem: Revisiting
  parametric min-cut and seeds for object proposals,'' in \emph{ICCV}, 2015.

\bibitem{Kraehenbuehl2015}
P.~Kr{\"a}henb{\"u}hl and V.~Koltun, ``Learning to propose objects,'' in
  \emph{CVPR}, 2015.

\bibitem{Carreira2012b}
J.~Carreira and C.~Sminchisescu, ``{CPMC}: Automatic object segmentation using
  constrained parametric min-cuts,'' \emph{TPAMI}, vol.~34, no.~7, pp.
  1312--1328, 2012.

\bibitem{Kraehenbuehl2014}
P.~Kr{\"a}henb{\"u}hl and V.~Koltun, ``Geodesic object proposals,'' in
  \emph{ECCV}, 2014.

\bibitem{Uijlings2013}
J.~R.~R. Uijlings, K.~E.~A. van~de Sande, T.~Gevers, and A.~W.~M. Smeulders,
  ``Selective search for object recognition,'' \emph{IJCV}, vol. 104, no.~2,
  pp. 154--171, 2013.

\bibitem{Rantalankila2014}
P.~Rantalankila, J.~Kannala, and E.~Rahtu, ``Generating object segmentation
  proposals using global and local search,'' in \emph{CVPR}, 2014.

\bibitem{Humayun2014}
A.~Humayun, F.~Li, and J.~M. Rehg, ``{RIGOR}: {R}ecycling {I}nference in
  {G}raph {C}uts for generating {O}bject {R}egions,'' in \emph{CVPR}, 2014.

\bibitem{Ren2015}
S.~Ren, K.~He, R.~Girshick, and J.~Sun, ``Faster {R-CNN}: Towards real-time
  object detection with region proposal networks,'' in \emph{NIPS}, 2015.

\bibitem{Zitnick2014}
C.~L. Zitnick and P.~Doll{\'a}r, ``Edge boxes: Locating object proposals from
  edges,'' in \emph{ECCV}, 2014.

\bibitem{Cheng2014}
M.-M. Cheng, Z.~Zhang, W.-Y. Lin, and P.~H.~S. Torr, ``{BING}: Binarized normed
  gradients for objectness estimation at 300fps,'' in \emph{CVPR}, 2014.

\bibitem{Alexe2012}
B.~Alexe, T.~Deselaers, and V.~Ferrari, ``Measuring the objectness of image
  windows,'' \emph{TPAMI}, vol.~34, pp. 2189--2202, 2012.

\bibitem{Manen2013}
S.~Man\'en, M.~Guillaumin, and L.~Van~Gool, ``{Prime Object Proposals with
  Randomized Prim's Algorithm},'' in \emph{ICCV}, 2013.

\bibitem{YuKo15}
F.~Yu and V.~Koltun, ``Multi-scale context aggregation by dilated
  convolutions,'' in \emph{ICLR}, 2016.

\bibitem{Zha+16}
H.~Zhao, J.~Shi, X.~Qi, X.~Wang, and J.~Jia, ``Pyramid scene parsing network,''
  in \emph{CVPR}, 2017.

\bibitem{Girshick2015}
R.~Girshick, ``Fast {R-CNN},'' in \emph{ICCV}, 2015.

\bibitem{Girshick2014}
R.~Girshick, J.~Donahue, T.~Darrell, and J.~Malik, ``Rich feature hierarchies
  for accurate object detection and semantic segmentation,'' in \emph{CVPR},
  2014.

\bibitem{Pont-Tuset2015b}
J.~Pont-Tuset and L.~Van~Gool, ``Boosting object proposals: From {Pascal} to
  {COCO},'' in \emph{ICCV}, 2015.

\bibitem{Hosang2015}
J.~Hosang, R.~Benenson, P.~Doll\'ar, and B.~Schiele, ``What makes for effective
  detection proposals?'' \emph{TPAMI}, vol.~38, no.~4, pp. 814--–830, 2016.

\bibitem{Kok16}
I.~Kokkinos, ``Ubernet: Training a universal convolutional neural network for
  low-, mid-, and high-level vision using diverse datasets and limited
  memory,'' in \emph{CVPR}, 2017.

\bibitem{Liu+16}
W.~Liu, D.~Anguelov, D.~Erhan, C.~Szegedy, and S.~Reed, ``Ssd: Single shot
  multibox detector,'' in \emph{ECCV}, 2016.

\bibitem{Red+16}
J.~Redmon, S.~Divvala, R.~Girshick, and A.~Farhadi, ``You only look once:
  Unified, real-time object detection,'' in \emph{CVPR}, 2016.

\bibitem{Dai+16}
J.~Dai, Y.~Li, K.~He, and J.~Sun, ``R-fcn: Object detection via region-based
  fully convolutional networks,'' in \emph{ECCV}, 2016.

\end{thebibliography}

\begin{IEEEbiography}[{\vspace{-3mm}\includegraphics[width=1in,height=1.25in,clip,keepaspectratio]{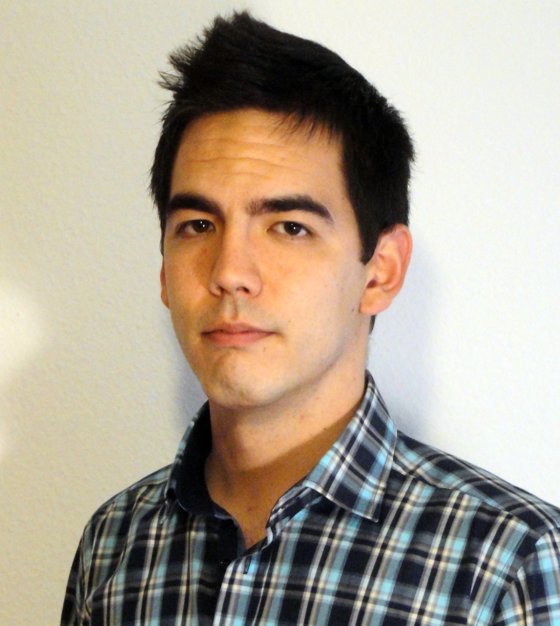}}]
{Kevis-Kokitsi Maninis} is a PhD candidate at ETHZ, Switzerland,
in Prof. Luc Van Gool's Computer Vision Lab (2015).
He received the Diploma degree in Electrical and Computer Engineering from National Technical University of Athens (NTUA) in 2014.
He worked as undergraduate research assistant in the Signal Processing and Computer Vision group of NTUA (2013-2014).
\end{IEEEbiography}

\begin{IEEEbiography}[{\includegraphics[width=1in,height=1.25in,clip,keepaspectratio]{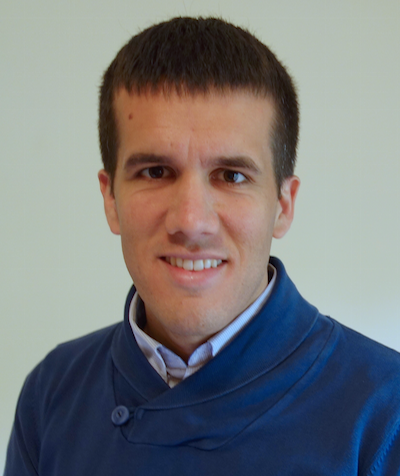}}]
{Jordi Pont-Tuset} is a post-doctoral researcher at ETHZ, Switzerland,
in Prof. Luc Van Gool's Computer Vision Lab (2015).
He received the degree in Mathematics in 2008, the degree in Electrical Engineering 
in 2008, the M.Sc. in Research on Information and Communication Technologies in 2010, and the Ph.D with honors in
2014; all from the Universitat Polit\`{e}cnica de Catalunya, BarcelonaTech (UPC).
He worked at Disney Research, Z\"urich (2014).
\end{IEEEbiography}

\begin{IEEEbiography}[{\includegraphics[width=1in,height=1.25in,clip,keepaspectratio]{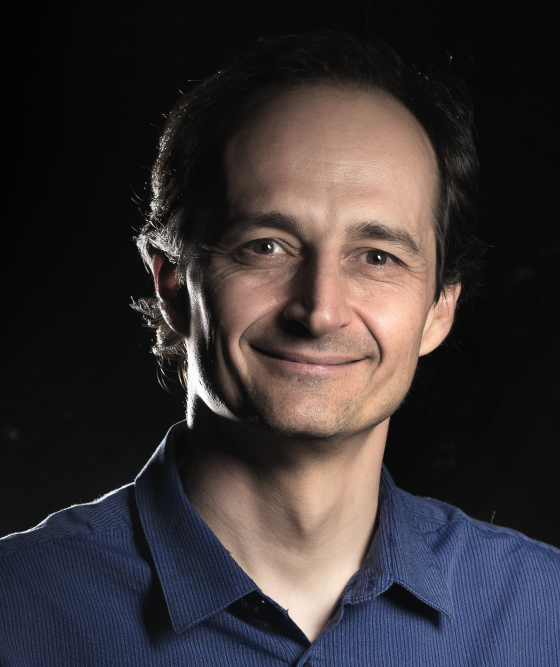}}]
{Pablo Arbel\'{a}ez} received a PhD with honors in Applied Mathematics from the Universit\'{e} Paris-Dauphine in 2005. 
He was a Research Scientist with the Computer Vision Group at UC Berkeley from 2007 to 2014. 
He currently holds a faculty position at Universidad de los Andes in Colombia.
His research interests are in computer vision, where he has worked on a number of problems, including perceptual grouping,
object recognition and the analysis of biomedical images.
\end{IEEEbiography}

\begin{IEEEbiography}[{\includegraphics[width=1in,height=1.25in,clip,keepaspectratio]{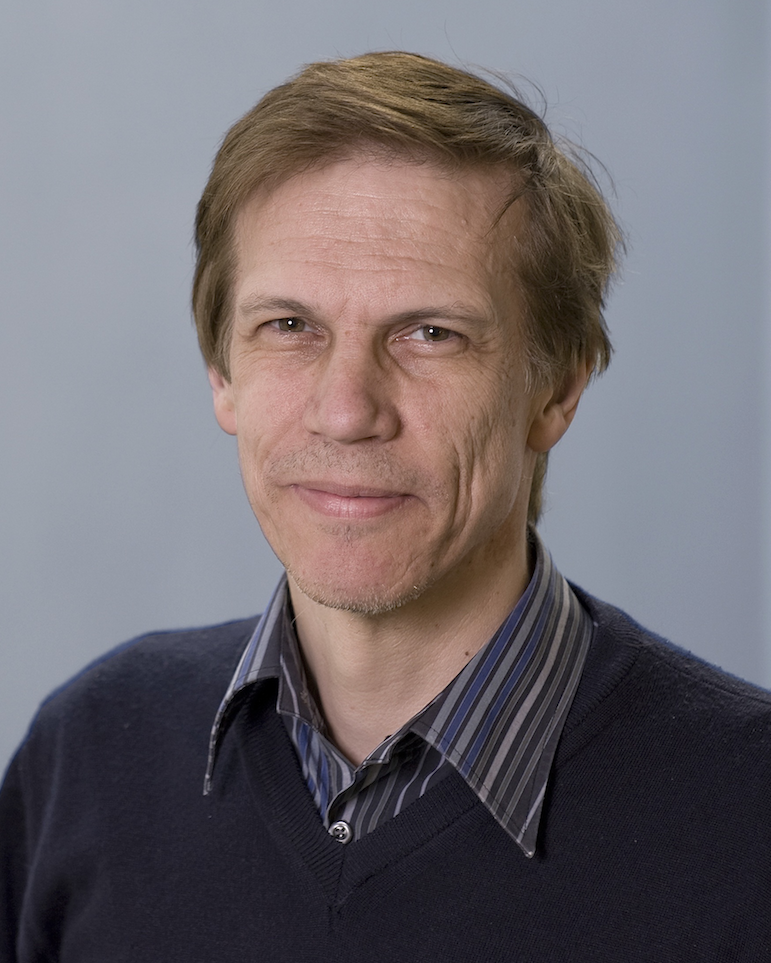}}]
{Luc Van Gool} got a degree in electromechanical engineering at the Katholieke Universiteit Leuven in 1981. Currently, he is professor at the Katholieke Universiteit Leuven, Belgium, and the ETHZ, Switzerland, Switzerland. He leads computer vision research at both places, where he also teaches computer vision. He has authored over 200 papers in this field. He has been a program committee member of several major computer vision conferences. His main interests include 3D reconstruction and modeling, object
recognition, tracking, and gesture analysis. He received several Best Paper awards. He is a co-founder of 5 spin-off companies.
\end{IEEEbiography}

\end{document}